\documentclass[acmtog]{acmart}
\AtBeginDocument{%
  }

\copyrightyear{2026}
\acmYear{2026}
\setcopyright{othergov}
\acmConference[SIGGRAPH Conference Papers '26]{Special Interest Group on Computer Graphics and Interactive Techniques Conference Conference Papers}{July 19--23, 2026}{Los Angeles, CA, USA}
\acmBooktitle{Special Interest Group on Computer Graphics and Interactive Techniques Conference Conference Papers (SIGGRAPH Conference Papers '26), July 19--23, 2026, Los Angeles, CA, USA}
\acmDOI{10.1145/3799902.3811167}
\acmISBN{979-8-4007-2554-8/2026/07}

\begin{teaserfigure}
	\centering
	\def\svgwidth{\linewidth}
	\begingroup%
  \ifx\svgwidth\undefined%
  \else%
  \fi%
  \global\let\svgwidth\undefined%
  \global\let\svgscale\undefined%
  \begin{picture}(1,0.24708077)%
    \put(0,0){\includegraphics[width=\unitlength,page=1]{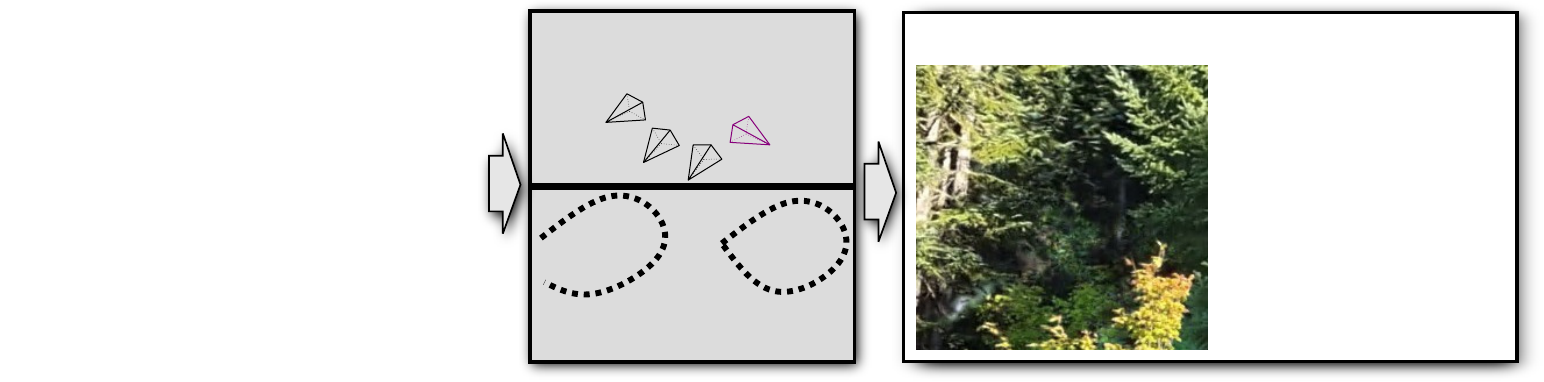}}%
    \put(0.44914391,0.21971625){\color[rgb]{0,0,0}\makebox(0,0)[t]{\smash{\begin{tabular}[t]{c}Pose Estimation\\with place recognition\end{tabular}}}}%
    \put(0.45087733,0.03875706){\color[rgb]{0,0,0}\makebox(0,0)[t]{\smash{\begin{tabular}[t]{c}Global Consistency\\with loop closure\end{tabular}}}}%
    \put(0.78429632,0.21561935){\color[rgb]{0,0,0}\makebox(0,0)[t]{\smash{\begin{tabular}[t]{c}Immediate Hierarchical Gaussian Reconstruction\end{tabular}}}}%
    \put(0.06814212,0.05899069){\color[rgb]{0,0,0}\makebox(0,0)[t]{\smash{\begin{tabular}[t]{c}Unordered Input\\Image Sequence\end{tabular}}}}%
    \put(0,0){\includegraphics[width=\unitlength,page=2]{teaser.pdf}}%
    \put(0.5939954,0.02603595){\color[rgb]{0,0,0}\makebox(0,0)[lt]{\smash{\begin{tabular}[t]{l}LoD fine\end{tabular}}}}%
    \put(0,0){\includegraphics[width=\unitlength,page=3]{teaser.pdf}}%
    \put(0.97111941,0.02686212){\color[rgb]{0,0,0}\makebox(0,0)[rt]{\smash{\begin{tabular}[t]{r}LoD coarse\end{tabular}}}}%
    \put(0,0){\includegraphics[width=\unitlength,page=4]{teaser.pdf}}%
  \end{picture}%
\endgroup%

	\vspace{-20pt}
	\caption{
	\label{fig:teaser}
	Our method takes an \emph{unordered} sequence of input RGB images and directly computes a radiance field from them. To achieve this, we introduce a fast online pose estimation approach building on fast visual place recognition, local bundle adjustment, loop detection and loop closure techniques. Furthermore, a hierarchical Gaussian model is jointly optimized, allowing immediate feedback of reconstruction results during arbitrarily large capture.
	}
\end{teaserfigure}

\usepackage{rotating}
\usepackage{colortbl}
\usepackage{stfloats}
\usepackage[ruled,lined,linesnumbered]{algorithm2e}

\renewcommand{\midrule}{\hline}
\renewcommand{\toprule}{\midrule\midrule}
\renewcommand{\bottomrule}{\midrule\midrule}

\definecolor{firstcolor}{RGB}{150,160,210} 
\definecolor{secondcolor}{RGB}{196,216,235} 
\newcommand{\first}[1]{\cellcolor{firstcolor}{#1}}
\newcommand{\second}[1]{\cellcolor{secondcolor}{#1}}

\makeatletter
\def\@mkteasers{%
  \ifx\@teaserfigures\@empty\else
    \long\def\@teaser##1{\par\bigskip\bgroup
      \captionsetup{type=figure}##1\egroup\par}
    \global\setbox\mktitle@bx=\vbox{\noindent\unvbox\mktitle@bx\par
      \noindent\@Description@presentfalse
      \@teaserfigures\par\if@Description@present\else
         \global\@undescribed@imagestrue
         \ClassWarning{\@classname}{A possible image without
           description}\fi
    \medskip}%
  \fi}
\makeatother

\begin{document}
\graphicspath{{figures/sources}}

\title{Immediate 3D Gaussian Splat Reconstruction of Unordered Input with Global Consistency}


\author{Andreas Meuleman}
\email{andreas.meuleman@gmail.com}
\affiliation{
	\institution{Inria, Université Côte d'Azur}
	\country{France}
}
\author{Linus Franke}
\authornote{Significant contribution.}
\email{linus.franke@fau.de}
\affiliation{
	\institution{Inria, Université Côte d'Azur}
	\country{France}
}
\author{Boris Zhestiankin}
\email{boris.zhestyankin@gmail.com}
\affiliation{
	\institution{Inria, Université Côte d'Azur}
	\country{France}
}
\affiliation{
	\institution{EPFL}
	\country{Switzerland}
}
\author{Camille Montemagni}
\email{contact@cammonte.com}
\affiliation{
	\institution{Inria, Université de Rennes}
	\country{France}
}
\author{George Drettakis}
\email{George.Drettakis@inria.fr}
\affiliation{
	\institution{Inria, Université Côte d'Azur}
	\country{France}
}


\begin{abstract}
3D Gaussian Splatting (3DGS) has become the method of choice for reconstructing and real-time rendering of captured scenes. 
To capture a scene with good visual quality, continuous image sequences are usually combined with out-of-order shots for better scene coverage. Structure from motion can reconstruct such captures, but only after they are all available and often with high computational cost. Incremental reconstruction methods -- often derived from SLAM solutions -- provide immediate feedback, but cannot handle the out-of-order capture we require. We provide the first immediate feedback solution for such radiance field capture that provides global consistency. We first introduce a method for fast matching in out-of-order sequences, by repurposing visual place recognition models and a covisibility graph, and provide an efficient way to find highly connected keyframes, improving quality even for ordered sequences. We show how these steps -- together with GPU optimization and careful Gaussian primitive placement -- provide fast local reconstruction, in our challenging radiance field reconstruction case. We then introduce a novel cluster-based method, again using the covisibility graph, to provide efficient loop closure that does not require sequential input. Finally, to handle large scenes in our context, we introduce a progressive hierarchy that allows our method to scale to large environments, without compromising efficiency.
Our results show we provide immediate feedback 3DGS reconstruction with good visual quality in several datasets, with up to thousands of input images.
\end{abstract}


\maketitle

\section{Introduction}

Radiance fields, and in particular 3D Gaussian Splatting (3DGS), have seen impressive adoption in the last few years~\cite{fei20243d}, with applications in many different fields such as VFX, architecture, e-commerce, virtual reality, etc. Creating a good quality 3DGS reconstruction typically involves unordered capture: users often take photos moving around abruptly, turning, or coming back to the same place to cover the entire scene.
Immediate feedback during capture is extremely important, allowing the user to determine if the capture is complete and of good quality.
Unfortunately, no solution exists that allows immediate feedback for typical 3DGS capture patterns.
We introduce a method that provides immediate feedback, and can handle the typical capture style required for 3DGS.

A vast majority of previous work on 3DGS assumes that camera poses are provided as input, typically using Structure from Motion (SfM)~\cite{schoenberger2016sfm}. Recently there has been interest in immediate feedback, most often building on visual Simultaneous Localization and Mapping (SLAM) solutions~\cite{homeyer2024droid, hhuang2024photoslam}. These, and methods that also create 3D Gaussians~\cite{meuleman2025onthefly} assume that images are provided in sequential order, so that an incoming frame can be matched with a set of immediately preceding frames. This assumption is also made for SLAM methods (and mixed 3DGS methods~\cite{zhu2024_loopsplat}) that perform loop closure~\cite{Liso_2024_CVPR}. Both poses and Gaussians need to be updated when loop closure occurs, without compromising interactivity. 
Even efficient GPU-friendly approaches~\cite{meuleman2025onthefly} cannot scale to the unordered case since more keyframes are needed for bundle adjustment (BA).
All of these issues are exacerbated when handling large scenes with thousands of images as input. 

Providing immediate feedback for unordered 3DGS captures raises several problems. First, instead of just trying to match an incoming frame to the last few preceding ones, matches potentially must be found in the entire set of previously registered frames. For this, we first introduce a fast, two-level matching algorithm building on visual place recognition~\cite{ali2023mixvpr} and learned features~\cite{potje2024xfeat}. Once matching pairs are validated, we use a weighted covisibility graph---originally developed for loop closure~\cite{mur2015ORBSLAM}---to allow fast extraction of subsets of well-connected frames (i.e., with many matches) that can be distant in time. Second, we need to perform fast local reconstruction in our context.
To allow bundle adjustment to scale while still exploiting the GPU effectively, we use the covisibility graph to restrict problem size and random sampling to maintain fixed size. This, together with several GPU optimizations, allows us to maintain immediate feedback in this more complex setting, as well as efficient updates of Gaussian primitives. Third, we need to handle loop closure, without the convenience of ordering where time can be used to detect loops. We introduce a cluster-based approach, using our fast subset extraction, allowing us to robustly identify when the user comes back to a part of the scene already captured. We propose a cluster-based ``welding" scheme and an optimization-free method to efficiently propagate the update due to loop closure to poses and Gaussian primitives. Finally, we handle large scenes, which would quickly saturate VRAM if we apply our solution naively, since we would need to keep all keyframes. For this, we introduce a \emph{progressive hierarchy}, determining which Gaussians do not contribute to the current frame, and consequently finding which keyframes to use for online optimization. The hierarchy also allows real-time rendering of such large scenes.

In summary, our contributions are:
\begin{itemize}
\item A fast matching approach for unordered input for efficient retrieval of well-connected subsets of keyframes, allowing GPU-friendly local reconstruction.
\item Cluster-based loop detection for unordered sequences, with efficient optimization-free updates of poses and Gaussian primitives.
\item A progressive hierarchy for large scenes that allows efficient optimization and rendering accommodating limited GPU resources.
\end{itemize}
Our experiments show we can handle the vast majority of datasets previously used for 3DGS, while providing immediate feedback during capture. Such datasets cause previous online methods to fail. Our method also improves ordered capture.

\section{Related Work}

We briefly cover closely related work on novel view synthesis, offline and incremental pose estimation/reconstruction including for 3D Gaussian Splatting and global consistency. Each is a vast field on its own, and we refer the reader to surveys available, respectively \cite{tewari2020state,chen2024survey3dgs}, \cite{ozyecsil2017survey,taketomi2017visual,macario2022comprehensive} and \cite{tsintotas2022revisiting}.

\paragraph{Novel View Synthesis}
Capturing real scenes from photographs or video ---allowing subsequent free-viewpoint navigation--- has seen impressive growth and adoption since the introduction of Neural Radiance Fields (NeRFs~\cite{mildenhall2020nerf,barron2021mipnerf}). 
NeRFs have a large computational overhead for volumetric ray tracing and neural network evaluation that was overcome by 3D Gaussian Splatting (3DGS)~\cite{kerbl3Dgaussians} and the numerous follow-up solutions~\cite{chen2024survey3dgs}. Recently, feed-forward radiance field methods have been proposed (e.g.,~\cite{chan2023generative,fan2024instantsplat,jiang2025anysplat,depthanything3}), but these solutions are based on direct image prediction, have few guarantees on reconstruction accuracy and are usually limited in the number of images they can process.

Initial radiance field solutions focused on small environments since handling larger scenes requires many hours of processing~\cite{barron2023zipnerf}. Several solutions have been proposed for 3DGS, typically introducing a hierarchy~\cite{octreeGS, hierarchicalgaussians24, Yang2025V3DG}; these data structures require a preprocessing step and are thus not suitable in our immediate feedback context.

A few recent approaches are compatible with progressive reconstruction~\cite{meuleman2023localrf, meuleman2025onthefly, guo2025ontheflylargescale3dreconstruction}, but expect ordered input and do not handle loop closure. In contrast, our approach allows global consistency with progressive reconstruction.

\paragraph{Offline pose estimation for Radiance Field Reconstruction} 
NeRF and 3DGS methods frequently assume that camera poses were already reconstructed.
This is typically achieved using Structure from Motion (SfM) pipelines like COLMAP~\cite{schoenberger2016sfm}. Such methods fundamentally expect all images to be available before processing begins. They typically build maps incrementally by reordering images based on connectivity and use expensive ---but accurate--- exhaustive matching processes.
These solutions (e.g., \cite{schoenberger2016sfm}, or the more recent \cite{pan2024glomap}) use global bundle adjustment, enabling high quality and globally consistent reconstructions, handling loop closure. Unfortunately, this comes at a high computational cost, because joint optimization of poses and landmarks scales poorly. Direct solvers for the Schur complement exhibit cubic complexity in the number of cameras in the worst case, while iterative methods like PCG~\cite{2021megba} reduce per-iteration cost but require more iterations as the problem conditioning degrades with size. 
Although hierarchical approaches such as COLMAP's Hierarchical Mapper improve scalability, they remain fundamentally offline processes, making them unsuitable for the strict latency constraints of interactive capture.
This precludes the use of such solutions to treat poses for large scenes and achieve immediate feedback.
Recent learning-based approaches have received attention, and in particular multiview-transformer solutions~\cite{maggio2025vggt-slam, murai2024_mast3rslam}; however, these still have significant computation overhead and limited accuracy, making them unsuitable in our context.

\paragraph{Incremental Reconstruction and 3DGS}
\label{sec:prev_incr}
Visual SLAM methods 
\linebreak
\cite{macario2022comprehensive} have been developed in the different context of robotics and fundamentally process frames \emph{sequentially}, exploiting temporal consistency for efficiency. Typically, only a small window of previous frames is used as a constrained feature search window to register a new image, and motion priors can be exploited. This limits robustness in the case of fast motion which is common for radiance field capture.
There is a vast body of research on ordering and relocalization such as ORB-SLAM~\cite{mur2015ORBSLAM}, using bag-of-words \cite{Lopez2012_BoBW}. These methods treat tracking and relocalization as distinct modes, pausing reconstruction when tracking is lost. Instead, we continuously search for the best possible matches. 
Most SLAM systems use CPU-based local bundle adjustment via Ceres~\cite{ceres} or g2o~\cite{kuemmerle2011g2o}. GPU-accelerated alternatives exist for dense systems like DROID-SLAM~\cite{teed2021droid} or large-scale sparse problems~\cite{2021megba, cudaba}. 
DROID-SLAM's system density hurts performance and the sparse solvers show large overhead and cannot optimize shared camera intrinsics due to their solver structure. 
Hierarchical methods for SLAM (e.g.,~\cite{hughes2022hydra}) are based on scene graphs; we propose a cluster-based approach specific to unordered input, and overall repurpose SLAM tools for radiance field capture with immediate feedback.

Several methods integrate Gaussian Splatting with pose estimation, but most rely on external SLAM backends~\cite{homeyer2024droid, hhuang2024photoslam} and inherit their limitations. End-to-end approaches exist but are either slow~\cite{lin2025longsplat,COGS2024, Fu_2024_CVPR} or restricted to limited camera motion, with no relocalization or loop closure \cite{gsslam2024,meuleman2025onthefly}. Others are demonstrated on ordered input~\cite{cheng2025outdoor,feng2024CaRtGS,MGSO,pan2025egg,wu2025monocular,deng2025gigaslam}. Some recent work focuses on specific input modalities (stereo~\cite{xin2025large}, RGB-D~\cite{deng2026vpgsslamvoxelbasedprogressive3d}, RGB-LIDAR~\cite{TLS-SLAM}).
We introduce several new solutions to improve the quality of the incremental joint optimization of poses and 3D Gaussians from unordered RGB input, while maintaining efficiency. 

\paragraph{Global Consistency}
Loop closure detection and pose graph optimization offer a more lightweight alternative to global bundle adjustment by optimizing poses only from loop closure constraints, but the methods typically remain iterative~\cite{Strasdat2010ScaleDL, tsintotas2022revisiting, konolige2008frameslam}.
Specific solutions have been explored, e.g., targeted to building semantics~\cite{sgraphs2} or using RGB-D inputs~\cite{whelan2015elasticfusion}; we target a more general algorithm. Neural SLAM methods show promise (e.g.,~\cite{Liso_2024_CVPR}) but current solutions do not yet achieve the performance/accuracy levels we require.
LoopSplat~\cite{zhu2024_loopsplat} corrects a Gaussian splat reconstruction by rigidly transforming primitives after loop closure, but targets RGB-D input and cannot handle scale drift inherent to monocular systems.
Our solution for global consistency builds on our covisibility graph, allowing an efficient optimization-free method to deform the trajectory via graph traversal, avoiding iterative optimization over all poses, and updating landmarks and Gaussians with scale taken into account.

\section{Overview}
{
For each incoming keyframe, we estimate its pose by matching against previously registered keyframes and running bundle adjustment, then back-project Gaussians and refine the representation. 
Unlike sequential captures where matching against a short temporal window is sufficient, unordered input requires efficiently searching the entire history for overlapping views, correcting drift on revisits, and bounding GPU memory as the scene grows.
}
We first perform a fast two-level matching approach and create a covisibility graph, together allowing us to quickly find well-connected subsets of keyframes from the entire registered sequence. Second, we introduce a fast, GPU-optimized local bundle adjustment used for pose estimation; a set of local Gaussians is jointly initialized and optimized per keyframe.
Third, we present a graph-based solution for loop closure that propagates changes using our covisibility graph and proposes a fast optimization-free update for poses and Gaussians. 
Finally, we present a progressive hierarchy for large scenes. We present each component next.
{A pipeline overview figure and pseudocode of the full algorithm are provided in the appendix, along with further algorithmic details for each step.}

\begin{figure}[!h]
\includegraphics[width=.49\linewidth]{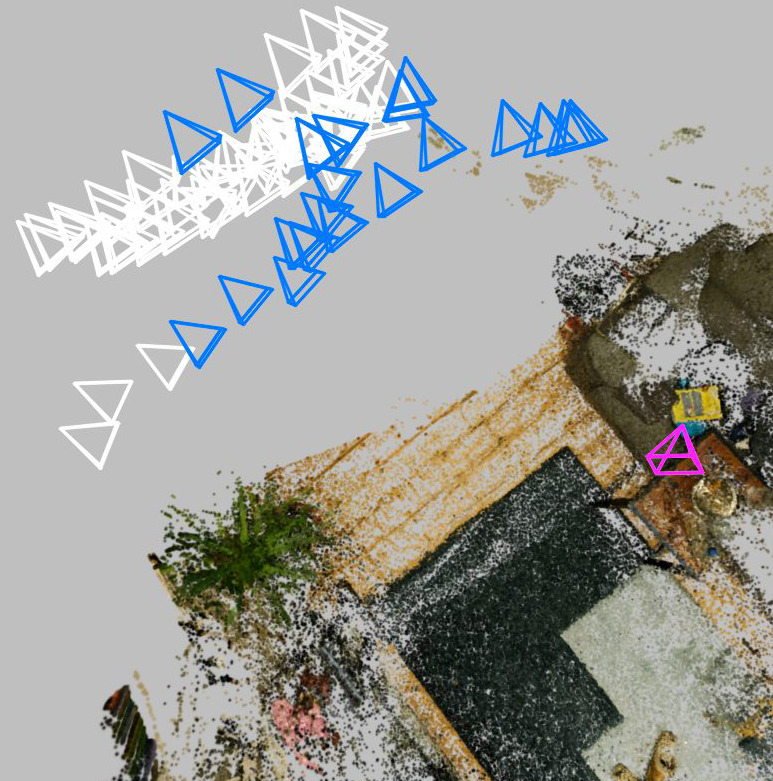}
\includegraphics[width=.49\linewidth]{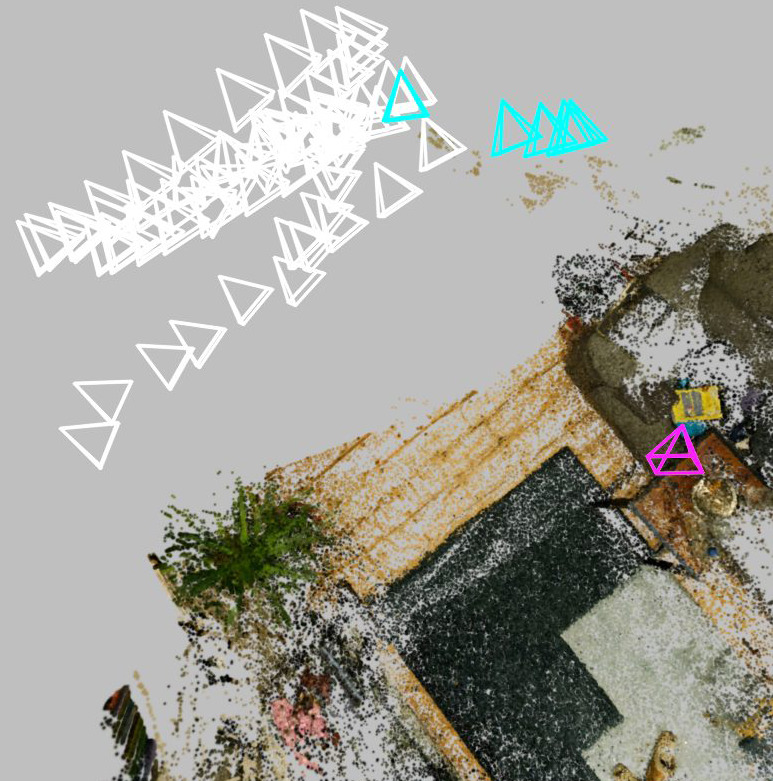}
\vspace{-5pt}
\caption{
\label{fig:mixvpr}
		We show an intermediate sequence of registered frames for the MipNeRF360 \textsc{room} scene. In white are all registered frames, and in magenta is the incoming frame $I_q$, which is out of order.
Left we show in blue the $\mathcal{K}=20$ images selected by MixVPR and right we show in cyan the $\mathcal{K}'=5$ images selected with brute-force matching and validated as pairs.
}
\end{figure}

\section{Fast Out-of-order Matching}
\label{sec:data_assoc}

Typical radiance field captures include abrupt motion around the scene, potentially requiring all previous keyframes to be examined to match the incoming keyframe. This is prohibitively expensive even for moderately large scenes. 
To find reliable keyframe pairs with potential matches, we introduce an efficient, two-level matching approach that builds on relocalization and loop closure tools for ordered sequences.
We then need to select good keyframes to use for local reconstruction (Sec.~\ref{sec:local_recons}): for this we use a weighted covisibility graph~\cite{mur2015ORBSLAM} with a greedy traversal. The graph is also central for global consistency (Sec.~\ref{sec:loop_closure}), allowing efficient pose update propagation.

\subsection{Efficient Two-level Matching}
\label{sec:twolevel}

For a new incoming (or query) keyframe $I_q$, we need to find the best candidates for matching from all previously registered keyframes.
Our two-level matching approach first uses a learned visual place recognition model, MixVPR~\cite{ali2023mixvpr},
to retrieve the $\mathcal{K}$ most similar keyframes for each incoming frame. 
Importantly, MixVPR extracts a 4096-dimensional global descriptor $\mathbf{d}$ for each image that can be compared to previously registered frames via cosine similarity, making it efficient for very large-scale retrieval.
{
Formally, the similarity between query frame $I_q$ and a registered keyframe $I_i$ is $s(I_q, I_i) = \mathbf{d}_q \cdot \mathbf{d}_i$, where $\mathbf{d}_q$ and $\mathbf{d}_i$ are the respective normalized global descriptors.
We retrieve the top-$\mathcal{K}$ candidates as $\mathcal{K} = \text{top-K}\{s(I_q, I_i) \mid I_i \in \mathcal{M}\}$, where $\mathcal{M}$ is the set of all registered keyframes, using $\mathcal{K}=20$.

Given this efficient preliminary step, we run a first round of geometric verification on the $\mathcal{K}$ candidates using brute-force feature matching with XFeat~\cite{potje2024xfeat}, a fast and robust local feature extractor, retaining the $\mathcal{K}'$ keyframes with the highest inlier count:
$\mathcal{K}' = \text{top-K'}\left\{\left|\mathcal{I}_i^{\text{BF}}\right| \mid I_i \in \mathcal{K}\right\}$,
where $\mathcal{I}_i^{\text{BF}}$ is the set of brute-force matches between $I_q$ and $I_i$, and $\mathcal{K}'=5$.
We then perform accurate matching on this reduced set using LightGlue~\cite{lindenberger2023lightglue}, ensuring efficiency through mixed precision and CUDA graphs.
A pair is validated if it yields more than $\tau_{m} = 500$ inliers after RANSAC with a fundamental matrix model:
$\mathcal{V} = \left\{I_i \in \mathcal{K}' \mid \left|\mathcal{I}_i^{\text{LG}}\right| > \tau_{m}\right\}$,
where $\mathcal{I}_i^{\text{LG}}$ is the set of LightGlue matches between $I_q$ and $I_i$.
A visualization can be seen in Fig.~\ref{fig:mixvpr}.
}

{Our matching is a speed/quality cascade: MixVPR matches at $\approx$1ms/1000 pairs, XFeat at $\approx$0.5ms/pair, and LightGlue+RANSAC at $\approx$4ms/pair, with corresponding quality gains. Each stage filters for the next, so we can only afford the slower and more accurate methods on progressively smaller candidate sets.}

This solution robustly tracks frames even when they are added out of order, or when revisiting places after a long time, without requiring a dedicated relocalization module. We maintain the speed required for immediate feedback and in addition we select the most relevant keyframes even for ordered sequences.

\subsection{Weighted Co-visibility Graph}
\label{sec:covisibility}

A key element of our solution is the ability to quickly identify and access most closely linked keyframes, both for local matching of an incoming frame but also to rapidly determine keyframes needed in loop closure (Sec.~\ref{sec:loop_closure}). 
We solve this by constructing 
a \emph{weighted covisibility graph}~\cite{mur2015ORBSLAM}
to represent the connectivity between keyframes based on their shared observations. 
Each node in the graph corresponds to a keyframe, and an edge is created between two nodes if the pair has been validated: we add the edges between $I_q$ and $\mathcal{V}$ into the graph ${G} = (\mathcal{M}, E, W)$ after each new frame is tracked. 
The edge weight is defined from the number of inlier matches between the two keyframes: $w_{ij} = \frac{1}{\#\text{inliers}(i,j)}$. 
A higher number of inliers indicates a stronger geometric relationship, i.e., a lower weight.
These weights allow us to prioritize stronger connections with efficient graph traversal algorithms, which is useful for local reconstruction, 
our solution for loop closure with clustering (Sec.~\ref{sec:loop_closure}) and for efficient drift correction (Sec.~\ref{sec:drift}).

\subsection{Finding Well-connected Subsets in the Graph}
\label{sec:keyframeselection}

Both choosing keyframes for local bundle adjustment (Sec.~\ref{sec:local_ba}), and finding keyframes on both sides of a loop closure (Sec.~\ref{sec:welding}), require the selection of sets of keyframes that are strongly connected.
We leverage the covisibility graph for this. We first describe connectivity-based keyframe selection for a new incoming frame $I_q$; we explain how it generalizes later. 

We consider an active set of nodes $\mathcal{S}_t$, initialized with the incoming frame $I_q$. In the graph, $I_q$ is connected to the validated pairs as explained above. We then start a greedy graph traversal, to find the nodes (keyframes) that are strongly connected to $I_q$. 
At each step, we add a node that has (locally) maximum connectivity to the active set $\mathcal{S}_t$. To find the node to add, we visit all candidate nodes, i.e., all nodes connected to a member of $\mathcal{S}_t$. We compute the total connectivity $c$ of each such node $I_i$ to $\mathcal{S}_t$, where connectivity is defined as: $c(I_i, \mathcal{S}_t) = \sum_{I_j \in \mathcal{S}_t} \#\text{inliers}(i,j)$.
We then choose the node $I_i$ with the maximum cumulative inliers, which can be written as:
\begin{equation}
\mathcal{S}_{t+1} = \mathcal{S}_t \cup \left\{\operatorname{argmax}_{I_i \in \mathcal{M} \setminus \mathcal{S}_t} c(I_i, \mathcal{S}_t)\right\}
\end{equation}

\begin{figure}[!h]
\def\svgwidth{\linewidth}
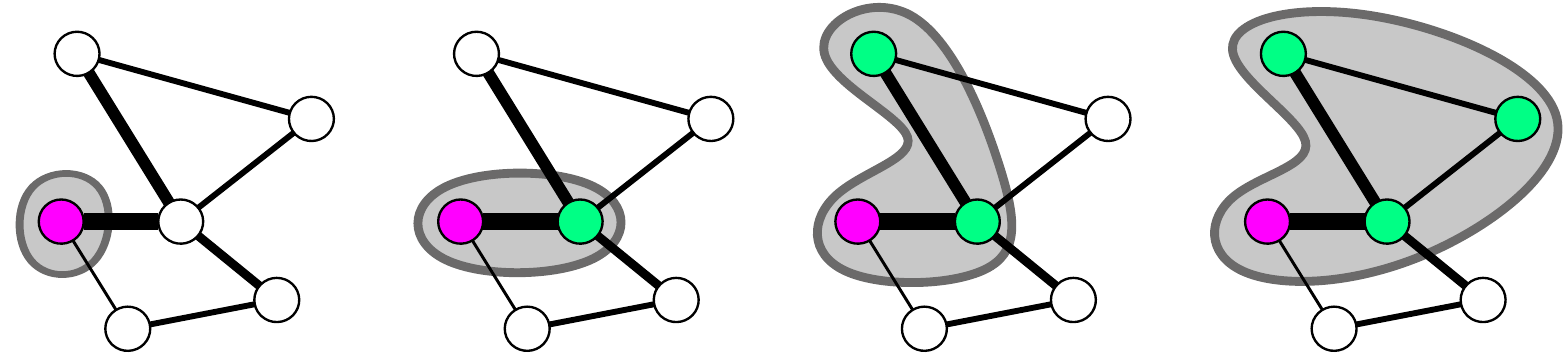
\vspace{-11pt}
\caption{\label{fig:greedy_graph}
Finding well-connected subsets with a greedy algorithm: From a starting frame $I_q$ (magenta), we expand the set $\mathcal{S}_t$ (gray) iteratively by considering the direct neighbors and including the node with the largest combined connectivity (indicated by strength of connection). 
}
\end{figure}

We continue with such steps until we have the desired number of keyframes $N$, which depends on the use case (see Fig.~\ref{fig:greedy_graph}).
Given that each graph node has a small number of connections (around 10 in our experiments), this greedy approach is very efficient.

\section{Efficient Local Pose and Gaussian Reconstruction}
\label{sec:local_recons}

In the following, we use the terms \textit{observations} for the 2D screen locations of matched features and \textit{landmarks} for the 3D points of matched features.
When reconstruction starts, we bootstrap our model by exhaustively matching the first 8 well-connected keyframes, rectifying outliers using Lambda Twist~\cite{Persson_2018_ECCVLambdaTwist} on the GPU as a minimal P3P solver for RANSAC.
We then use Depth Anything 3~\cite{depthanything3} to initialize poses, focal length and landmarks,
fixing inaccuracies in them with local bundle adjustment (BA).

\subsection{Local Bundle Adjustment}
\label{sec:local_ba}

At each incoming frame, we perform local BA. We first use the approach described above (Sec.~\ref{sec:keyframeselection}) to select $N=20$ keyframes, often far back in time in the set of registered keyframes. 
This is a higher number of keyframes than previous methods {that either consider direct matches only (ORB-SLAM) or a temporal window (5 frames for DROID-SLAM),} since the out-of-order problem is harder; the challenge is to be robust while maintaining efficiency.

To allow efficient GPU BA, we need to use a fixed set of landmarks. For this we perform multinomial sampling based on the number of observations per landmark to randomly select the $K$ landmarks from those that are observed by at least two of the selected keyframes ($K=10000$ in our tests).
We follow standard practice to fix $n_f$ keyframes that are used in the loss but not optimized (we use $n_f=4$).
We randomly choose these from the $N$ selected keyframes, which proves to be robust in practice.

\paragraph{Solver efficiency} 
Previous fast BA solutions~\cite{meuleman2025onthefly} are 
efficient and accurate for very small problems ($N=5$), but do not scale for the $N=20$ keyframe case we have.
We address this limitation while maintaining the GPU efficiency by allowing each landmark to be observed by a fixed subset of keyframes in the local window, selected based on covisibility. 
Furthermore, we implement GPU kernels for residual and Jacobian computation, and maintain a dense Schur complement formulation, which shows better performance than sparse solvers in our context.
We achieve 5$\times$ speedup against existing CPU and GPU optimizers \cite{kuemmerle2011g2o, cudaba, meuleman2025onthefly} (Sec.~\ref{sec:results}).

\subsection{Gaussian Placement and Optimization}

After a keyframe pose is estimated and refined with local BA, new Gaussian primitives are added to the reconstruction.
Recent methods place Gaussian primitives
using learned depth predictors with a scale and offset factor, typically computed once for the entire image~\cite{hierarchicalgaussians24}.
This, however, is often inaccurate~\cite{fink2025refinement}.
To overcome this limitation, we split the images into tiles, and compute a scale and offset for each tile. 
{We then upsample with linear interpolation to match the input resolution.}
This provides much finer correction to the predicted depth, improving the quality of the positions chosen for the Gaussians (see Sec.~\ref{sec:ablations}). 

Using this depth, we follow \citet{meuleman2025onthefly} in spawning new Gaussians. These Gaussians are directly linked to the keyframe, allowing later loop closure events (Sec.~\ref{sec:loop_closure}) to update the radiance field.
For fast optimization, the full Gaussian model is trained for 30 iterations using a custom differentiable renderer based on \citet{taming3dgs}, sampling nearby views (see Sec.~\ref{sec:frame_hierarchy}). 
{We fix 30 iterations per keyframe for reproducible evaluation; for live capture we instead run optimization asynchronously until the next keyframe is added.}

\subsection{Dealing with Unmatchable Keyframes}

When there is no visual overlap to the registered keyframes, pose estimation will fail.
In this case, we ``shelve" the keyframe to be retried later. 
When the cosine similarity of the shelved keyframe's visual descriptor compared to the registered keyframes becomes large, the keyframe is unshelved and retried.
This efficiently introduces a reordering scheme to avoid failure cases and allow disjoint sequences to be eventually matched. An exception to this is if the keyframe was unmatched because of low parallax, estimated with the triangulation angle. In this case, we disregard it as we also assume it to contribute little to the reconstruction quality.

\section{Out-of-order Global Consistency}

Reconstruction error accumulates over time, resulting in drift. Loop closure is required to correct the drift, and consolidate everything into the same space. Traditional ordered sequence loop closure uses time stamps to identify if the candidate 
has not been seen recently, simplifying loop closure detection; in the out-of-order setting, this is no longer possible.

To solve this problem we introduce \emph{clusters} of keyframes for robust detection.
When adding a new keyframe, we force the creation of two clusters of poses from the matched keyframes, the first one containing the new pose and poses connected to it.
If there is a second cluster of poses that is sufficiently large and disconnected from the first ---despite being matched--- this indicates a set of poses that ``see the same content'', thus requiring loop closure. 
We next use the clusters to robustly ``weld'' the respective poses, and propagate corrections efficiently with the graph to other poses and Gaussians. 

\subsection{Cluster-Based Loop Detection}
\label{sec:loop_closure}

We first take $2K'$ best matches 
(Sec.~\ref{sec:twolevel}), and create two separate clusters of keyframes $\mathcal{C}_1, \mathcal{C}_2$. 
{Selecting $2K'$ allows clusters of size $K'$ in both clusters if evenly distributed.}
We initialize the two clusters with the two keyframes that are the most distant in terms of connectivity. 
We run local reconstruction (Sec.~\ref{sec:local_recons}) on the largest of the two clusters. 
To determine if the clusters are not sufficiently connected, we check the number of hops in the covisibility graph that separates them. If the clusters are separated by more than $\tau_{lc}=5$ hops, and the smaller cluster has more than $\tau_{v}=2$ pairs, a loop closure occurs. 

Our approach does not require timestamps and is robust to small isolated match sets that are likely false positives. It makes local BA more robust by discarding small clusters, makes
the connectivity graph more accurate and prevents loop closure false positives.

\begin{figure}[!h]
\includegraphics[width=.49\linewidth]{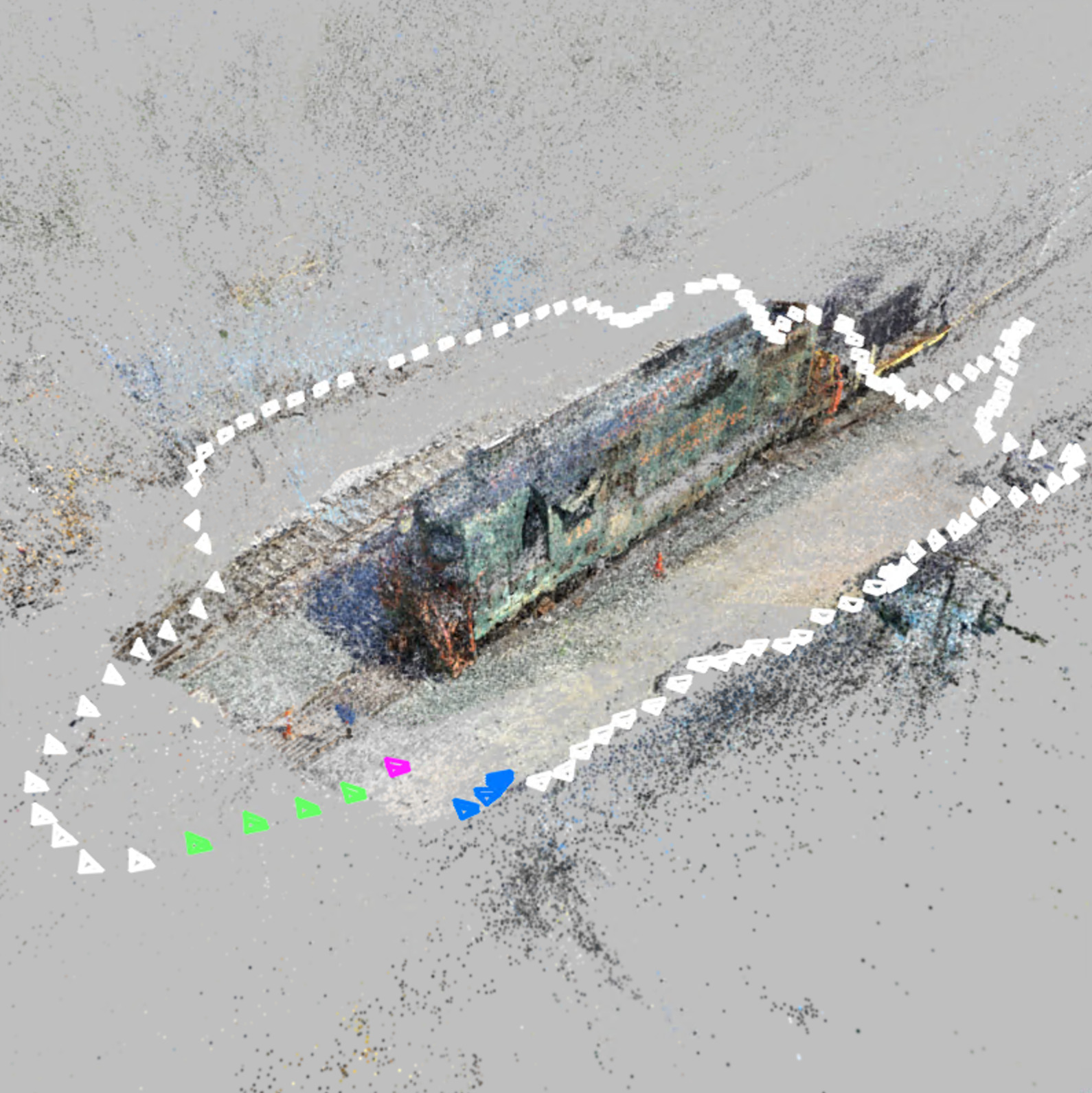}
\includegraphics[width=.49\linewidth]{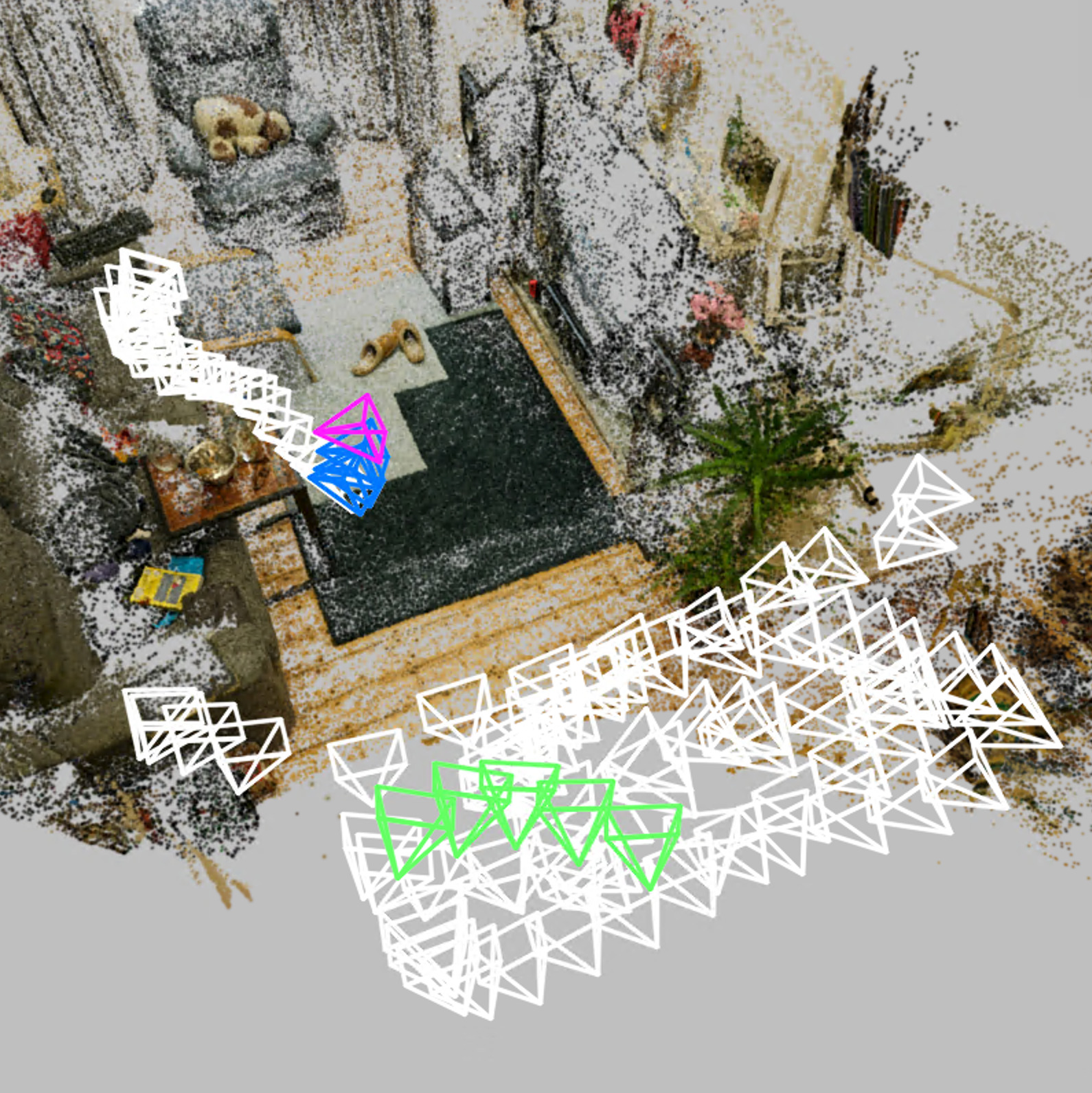}
\vspace{-5pt}
\caption{Left we show cluster-based loop closure in the train scene, in a (mostly) ordered scenario: blue poses are the new (free) cluster, while green are the fixed cluster. On the right, we use the same color coding, and we see that the two clusters come from disjoint, unordered sequences.
}
\label{fig:loop}
\end{figure}

\subsection{Welding Window Bundle Adjustment}
\label{sec:welding}

Once a loop closure is detected, we have two clusters of keyframes that are in separate spaces, and that require poses to be reoptimized together. We use the notion of a ``welding window''~\cite{mur2015ORBSLAM}, but in the new context of keyframe clusters. 

We perform BA on this welding window around the detected loop closure.
We use our keyframe selection strategy (Sec.~\ref{sec:keyframeselection}) to select the most relevant sets of keyframes on both sides of the loop closure, ensuring a more robust and accurate optimization. 
In this case, the initial set $\mathcal{S}_t$ contains the keyframes of each cluster.
We expand from there to select $N/2$ keyframes on each side {of the loop closure, allowing balanced optimization with our N-frames welding BA.}
We take the union of these expanded keyframe sets to define the welding window.
We then run our efficient local BA (Sec.~\ref{sec:local_recons}) over this window to optimize the poses and landmarks. To constrain landmark motion overall, we fix $n_f$ keyframes in the older set. 
We need to fix one set for the propagation step below, so we fix the old one as we expect it is likely already well optimized, more stable, and connected to more of the trajectory; the other is called the \emph{free cluster}.

\subsection{Drift Correction by Propagation Through the Graph}
\label{sec:drift}

Traditional solutions~\cite{Strasdat2010ScaleDL, whelan2015elasticfusion, whelan2016elasticfusion} use optimization to propagate corrections to the other poses after loop closure.
We avoid this by exploiting our covisibility graph to efficiently find ``most connected'' keyframes providing reliable poses. We estimate a rigid transformation that is propagated efficiently to other poses, by determining a \emph{backbone path} using Dijkstra's algorithm in the covisibility graph. 

In each cluster, we first select the set of most connected keyframes within the welding window $\mathcal{W}$: these \emph{anchor} keyframes have the most reliable pose. 
We then compute the rigid body transformation (SE(3)) between the initial and re-optimized pose for the anchor keyframe in the free cluster. 
To obtain the full similarity transformation $\mathbf{S}_{\Delta} \in Sim(3)$, we estimate the scale change $s_{\Delta}$ from landmark depth variations and compose it with the rigid transform.
We now have one anchor in each cluster and the similarity transform between their previous and ``welded'' pose. To propagate this transformation to all poses efficiently, we first compute a \emph{backbone path} composed of well-connected keyframes that are likely reliable,
using Dijkstra's algorithm to find the path with the smallest cumulative edge weights in the graph.
Loop edges are not yet in the graph, so the backbone goes through existing connections only. 
We then distribute the computed Sim(3) transformation along the backbone path, proportionally to the geodesic distance to the optimized keyframe.
We interpolate rotations with 6D representation \cite{Zhou_2019_CVPR} and scale using log interpolation.

Finally, we propagate the corrections to the rest of the trajectory by moving each keyframe following the transformation of the closest backbone keyframe in terms of geodesic distance, and applying the same correction. The most relevant backbone node for this keyframe is that with minimal geodesic distance.
After loop closure, we add the loop matches to the covisibility graph (Sec.~\ref{sec:covisibility}) so that they are used in subsequent local BA.

\begin{figure}[!h]
\def\svgwidth{\linewidth}
\small
\begingroup%
  \makeatletter%
  \providecommand\color[2][]{%
    \errmessage{(Inkscape) Color is used for the text in Inkscape, but the package 'color.sty' is not loaded}%
    \renewcommand\color[2][]{}%
  }%
  \providecommand\transparent[1]{%
    \errmessage{(Inkscape) Transparency is used (non-zero) for the text in Inkscape, but the package 'transparent.sty' is not loaded}%
    \renewcommand\transparent[1]{}%
  }%
  \providecommand\rotatebox[2]{#2}%
  \newcommand*\fsize{\dimexpr\f@size pt\relax}%
  \newcommand*\lineheight[1]{\fontsize{\fsize}{#1\fsize}\selectfont}%
  \ifx\svgwidth\undefined%
    \setlength{\unitlength}{934.38582293bp}%
    \ifx\svgscale\undefined%
      \relax%
    \else%
      \setlength{\unitlength}{\unitlength * \real{\svgscale}}%
    \fi%
  \else%
    \setlength{\unitlength}{\svgwidth}%
  \fi%
  \global\let\svgwidth\undefined%
  \global\let\svgscale\undefined%
  \makeatother%
  \begin{picture}(1,0.40634893)%
    \lineheight{1}%
    \setlength\tabcolsep{0pt}%
    \put(0,0){\includegraphics[width=\unitlength,page=1]{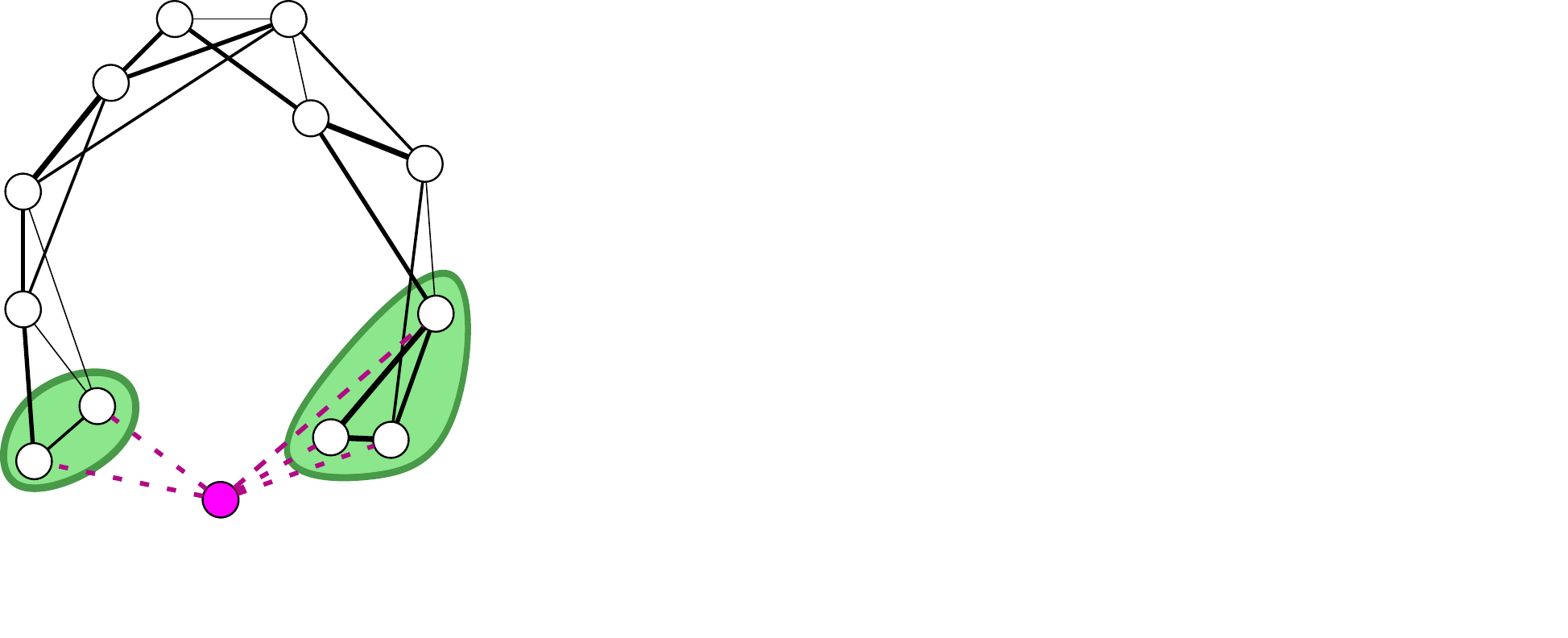}}%
    \put(0.15292842,0.02570666){\color[rgb]{0,0,0}\makebox(0,0)[t]{\smash{\begin{tabular}[t]{c}(a) Distinct Cluster \\Detection\end{tabular}}}}%
    \put(0,0){\includegraphics[width=\unitlength,page=2]{loopc.pdf}}%
    \put(0.50656499,0.02570666){\color[rgb]{0,0,0}\makebox(0,0)[t]{\lineheight{1.25}\smash{\begin{tabular}[t]{c}(b) Welding Window BA\end{tabular}}}}%
    \put(0.85953191,0.02570666){\color[rgb]{0,0,0}\makebox(0,0)[t]{\smash{\begin{tabular}[t]{c}(c) Backbone-based \\Propagation\end{tabular}}}}%
    \put(0,0){\includegraphics[width=\unitlength,page=3]{loopc.pdf}}%
  \end{picture}%
\endgroup%

\vspace{-11pt}
\caption{
    Loop Closure: (a) When a new keyframe (magenta) is matched (dotted lines) to distinct clusters of the graph, a loop closure is triggered. 
    (b) These clusters extended with connected frames then define a local welding window (gray), in which BA is done with older keyframes fixed (grayed out).
    (c) The changes are then propagated through the minimal backbone of the graph, with other poses connected through spatial locality. 
}
\end{figure}

Each Gaussian primitive added keeps track of the keyframe that added it. 
We apply the same Sim(3) transformation as that used for the keyframe that created each Gaussian, and we update scale and rotation accordingly.
\begin{table*}
\small
\caption{\label{tab:slam-eval} 
We show results for novel view synthesis on pose-free methods. We also include GLOMAP followed by Taming 3DGS (at 7k) (G+T3DGS) as an offline reference.
The {\colorbox{firstcolor}{best}} and {\colorbox{secondcolor}{second best}} are color coded for pose-free methods.}
\vspace{-7pt}
\begin{tabular}{l|cccc|cccc|cccc}
\toprule
 &
\multicolumn{4}{c|}{\textsc{TUM}} &
\multicolumn{4}{c|}{\textsc{MipNeRF360}} &
\multicolumn{4}{c}{\textsc{StaticHikes}}  \\
	{} & PSNR$^\uparrow$ & SSIM$^\uparrow$ & LPIPS$^\downarrow$ & Time$^\downarrow$ &
	PSNR$^\uparrow$ & SSIM$^\uparrow$ & LPIPS$^\downarrow$ & Time$^\downarrow$ &
	PSNR$^\uparrow$ & SSIM$^\uparrow$ & LPIPS$^\downarrow$ & Time$^\downarrow$ \\
\midrule
Photo-SLAM & 
{19.30} & {0.700} & 0.382 & {0:02:12} & {16.54} & {0.505} & {0.603} & {0:02:11} & 14.13 & {0.316} & 0.660 & \second{0:02:01} \\
MonoGS & 
16.60 & 0.682 & {0.381} & 0:16:18 & 14.46 & 0.436 & 0.663 & 0:04:05 & {15.46} & 0.301 & {0.659} & 0:09:19 \\
On-The-Fly NVS & 
{23.02} & {0.821} & {0.250} & \first{0:00:50} & \second{24.31} & \second{0.775} & \second{0.300} & \first{0:01:02} & \second{20.40} & \second{0.589} & \second{0.365} & \first{0:01:30} \\
S3PO-GS w\texttt{\textbackslash} refinement & 
{21.88} & {0.777} & {0.309} & {1:29:04} & {19.82} & {0.553} & {0.576} & {0:23:29} & {18.09} & {0.368} & {0.611} & {0:54:44} \\
LongSplat  & 
\first{26.51} & \first{0.875} & \first{0.176} & {1:41:08} & {23.63} & {0.657} & {0.415} & {2:25:26} & {20.05} & {0.461} & {0.481} & {22:40:56} \\
Ours & \second{24.77} & \second{0.853} & \second{0.205} & \second{0:01:22} & \first{26.29} & \first{0.839} & \first{0.241} & \second{0:01:17} & \first{22.12} & \first{0.696} & \first{0.289} & {0:02:14} \\
\hline
\hline
G+T3DGS (7k) & 
25.29 & 0.868 & 0.191 & 0:03:33 & 27.52 & 0.866 & 0.226 & 0:08:50 & 20.23 & 0.537 & 0.465 & 1:02:34 \\
\bottomrule
\end{tabular}
\end{table*}

\begin{table*}
	\caption{\label{tab:slam-eval-un} 
	We show results for novel view synthesis on unordered datasets: We include GLOMAP followed by Taming 3DGS (at 7k) (G+T3DGS) as an offline reference. GLOMAP requires respectively 
0:08:21, 0:03:53 and 0:04:12 for each dataset.
	}
\vspace{-7pt}
\small
\begin{tabular}{l|cccc|cccc|cccc}
\toprule
 &
\multicolumn{4}{c|}{\textsc{MipNeRF360}} &
	\multicolumn{4}{c|}{\textsc{Tanks and Temples}} &
\multicolumn{4}{c}{\textsc{Deep Blending}}  \\
	{} & PSNR$^\uparrow$ & SSIM$^\uparrow$ & LPIPS$^\downarrow$ & Time$^\downarrow$ &
	PSNR$^\uparrow$ & SSIM$^\uparrow$ & LPIPS$^\downarrow$ & Time$^\downarrow$ &
	PSNR$^\uparrow$ & SSIM$^\uparrow$ & LPIPS$^\downarrow$ & Time$^\downarrow$ \\
\midrule
	Ours & 25.60 & 0.772 & 0.270 & 0:01:28 & 21.42 & 0.774 & 0.252 & 0:01:37 & 23.83 & 0.813 & 0.290 & 0:01:47 \\
\hline
G+T3DGS (7k) & 27.08 & 0.815 & 0.262 & 0:09:57 & 21.90 & 0.779 & 0.271 & 0:05:13 & 25.80 & 0.855 & 0.268 & 0:05:36  \\
\bottomrule
\end{tabular}
\end{table*}

\begin{table*}
\caption{\label{tab:large-scale} Results for large scale scenes. For other methods time includes total time for COLMAP using the H3DGS and Octree-GS process and the actual 3DGS optimization of each method.}
\vspace{-7pt}
\small
\begin{tabular}{l|cccc|cccc|cccc}
\toprule
 &
\multicolumn{4}{c|}{\textsc{SmallCity*}} &
\multicolumn{4}{c|}{\textsc{Wayve*}} &
\multicolumn{4}{c}{\textsc{CityWalk}}  \\
  & PSNR$^\uparrow$ & SSIM$^\uparrow$ & LPIPS$^\downarrow$ & Time$^\downarrow$ &
	PSNR$^\uparrow$ & SSIM$^\uparrow$ & LPIPS$^\downarrow$ & Time$^\downarrow$ &
	PSNR$^\uparrow$ & SSIM$^\uparrow$ & LPIPS$^\downarrow$ & Time$^\downarrow$ \\
\midrule
{H3DGS} & 
21.17 & 0.679 & \second{}0.285 & 2:55:28 & 20.80 & 0.737 & \first{}0.227 & 7:29:45 & 11.78 & 0.557 & 0.560 & 22:09:20 \\
Octree-GS &  \first{}{24.18} & \first{}{0.831} & \first{}{0.273} & {1:20:59} &\second{} {22.58} & \second{}{0.765} & {0.313} & {2:21:48} & {17.33} & {0.638} & {0.520} & {3:11:40} \\
\hline
S3PO-GS w\texttt{\textbackslash} refinement & {19.71} & {0.652} & {0.462} & {0:57:31} & {17.81} & {0.644} & {0.423} & {1:25:26} & {10.72} & {0.439} & {0.579} & {6:00:28} \\
{On-The-Fly NVS} & \second{}23.59 & 0.789 & 0.323 &\first{}0:01:45 & 20.29 & 0.739 & 0.303 & \second{}0:04:29 & \second{}21.71 & \second{}0.712 & \second{}0.395 & \first{}0:25:03 \\
Ours& 23.56&\second{}0.800&0.302 & \second{}0:02:25 & \first{}23.04&\first{}0.822&\second{}	0.234&\first{}0:04:12 & \first{}22.03& \first{}0.772	& \first{} 0.336	& \second{}0:29:38 \\
\bottomrule
\end{tabular}
\end{table*}

\section{A Progressive Gaussian Hierarchy}
\label{sec:prog_hierarchy}

Processing large scenes for 3DGS requires some form of hierarchy~\cite{hierarchicalgaussians24,octreeGS}.
Such solutions require a substantial computational overhead to create or update the hierarchy that we cannot afford.
We propose a hierarchical structure that is fast to build and update, based on primitive screen size.
{
A node holds Gaussian parameters (position/opacity/rotation/scale/SH coefficients), and parent/children IDs. 
These enable going coarser or finer within the hierarchy, depending on required precision. 
Active nodes' Gaussian parameters are optimized with the rendering loss. 
}
At a given point in time, small footprint leaf Gaussian primitives will be offloaded to the CPU RAM, freeing up VRAM for the optimization. However, this poses challenges during progressive reconstruction.

\subsection{Adaptive Gaussian Merging and Retrieval}

We maintain an \emph{active set} of Gaussian primitives at any time, which contains leaf as well as intermediate nodes that result from merging.
We analyse the hierarchy every time a new keyframe is added, 
computing $S_i$, the screen size of each Gaussian $G_i$, from the current viewpoint.
We merge small, subpixel Gaussians with $S_i < \tau_l = 0.5$ pixels,
and only trigger merging when more than $20\%$ of the Gaussians fit this criterion to avoid too frequent merging.

Within the set of primitives selected for merging, we get the 4 closest Gaussians using a fast KNN search~\cite{Reducing3Dgs} and remove duplicates to prevent a Gaussian primitive from having multiple parents.
Merging is performed by groups of 4 Gaussians using KL-divergence based merging~\cite{hierarchicalgaussians24}.
We then update the hierarchy by
removing sub-pixel Gaussians from the active set, and adding new parent Gaussians, together with child-parent links. 
If a Gaussian already has a parent, we check if all children of this parent need to be merged; if so, we pull the parent from the hierarchy instead of creating a new node.
We can now optimize a smaller set of Gaussians, and can handle large scenes with limited GPU memory.
When loop closure occurs we move all Gaussians in the hierarchy following the transform of the keyframe most used to spawn its children; this is done asynchronously to avoid affecting latency. 

When navigating, if a merged node contributes more than $\tau_h = 1$, we reload the finer Gaussians and update their coarse parent nodes.
We then reload their children and remove these parent nodes from the active set, resulting in an
active set sufficiently fine to represent the scene for the keyframe used to optimize or the frame to render.
We use double buffering to avoid stutter.

\subsection{Frame Selection and Rendering}\label{sec:frame_hierarchy}

During online optimization, we need to avoid frames that mostly see offloaded primitives and use keyframes that see leaf or small Gaussians. For this we sample keyframes using a probability 
based on the number of leaf Gaussians still in the active set that were generated by this keyframe. We keep at least 50 keyframes to have context and optimize intermediate nodes. We then
remove keyframes that contribute $< 5\%$ of the max contribution by offloading them to the CPU.
{
Thanks to this scheme, intermediate hierarchy nodes are automatically optimized, not requiring explicit post-optimization.
}

We store the state hierarchy after optimization, and we can then reload it for rendering. 
We start with roots, i.e., nodes that do not have parents.
We then descend in the hierarchy until nodes are finer than $\tau = 1$, or we reach a leaf.
As we do this, we continuously check if there are parents whose node would have $S_i<\tau$ to potentially take a coarser node and offload Gaussians. This hierarchy traversal takes
5 to 30 ms per step depending on size, and is computed asynchronously 
with double buffering. 

\section{Results and Evaluation}
\label{sec:results}

\begin{figure*}[!ht]
\setlength{\tabcolsep}{1pt}
\renewcommand{\arraystretch}{0}
\begin{tabular}{ccccc}
    OTF-NVS & S3PO-GS & MonoGS & Ours & GT \\[2pt]
    {\includegraphics[width=0.195\textwidth]{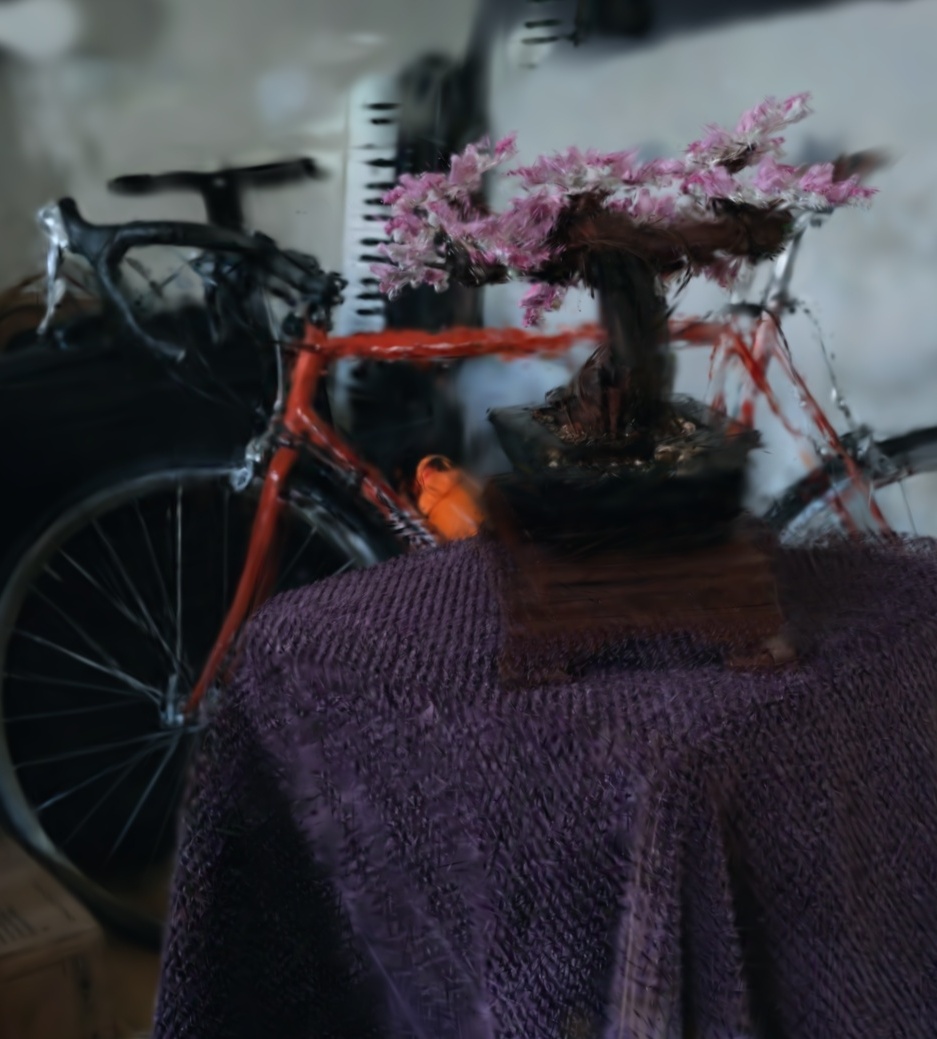}} &
    {\includegraphics[width=0.195\textwidth]{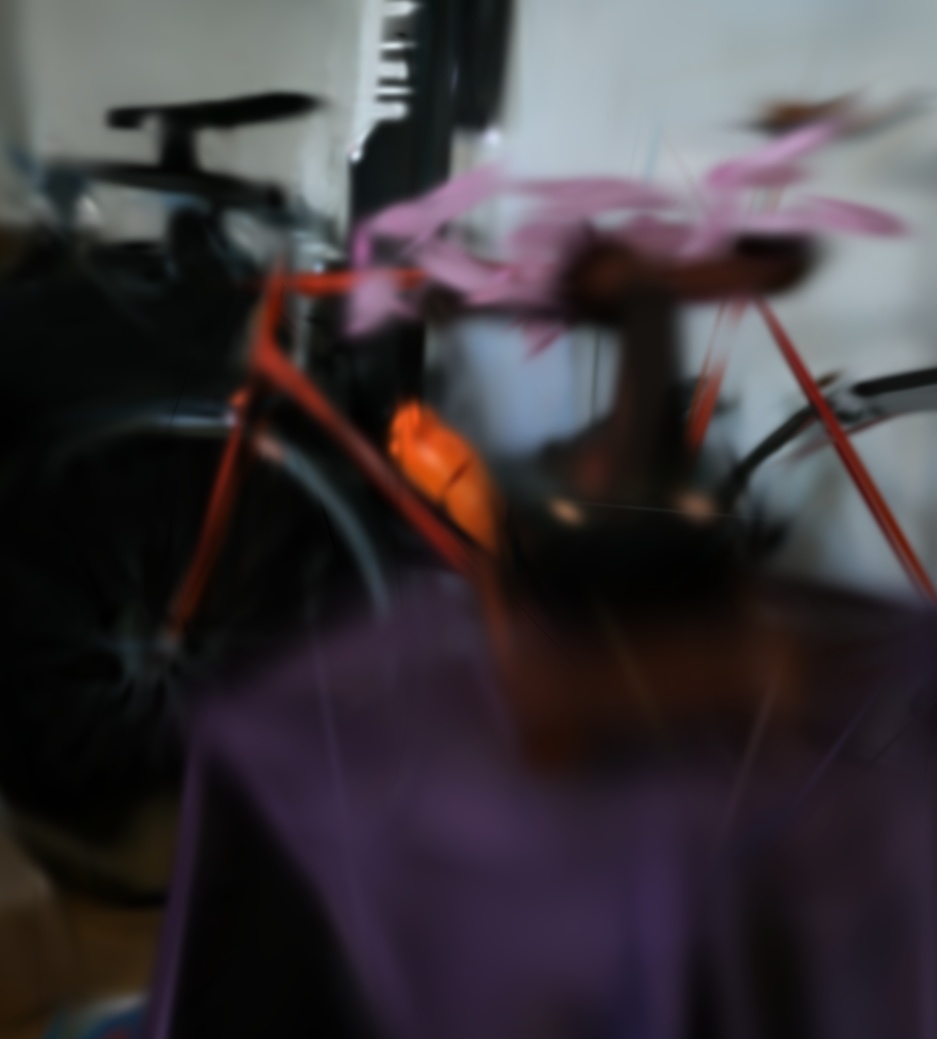}} &
    {\includegraphics[width=0.195\textwidth]{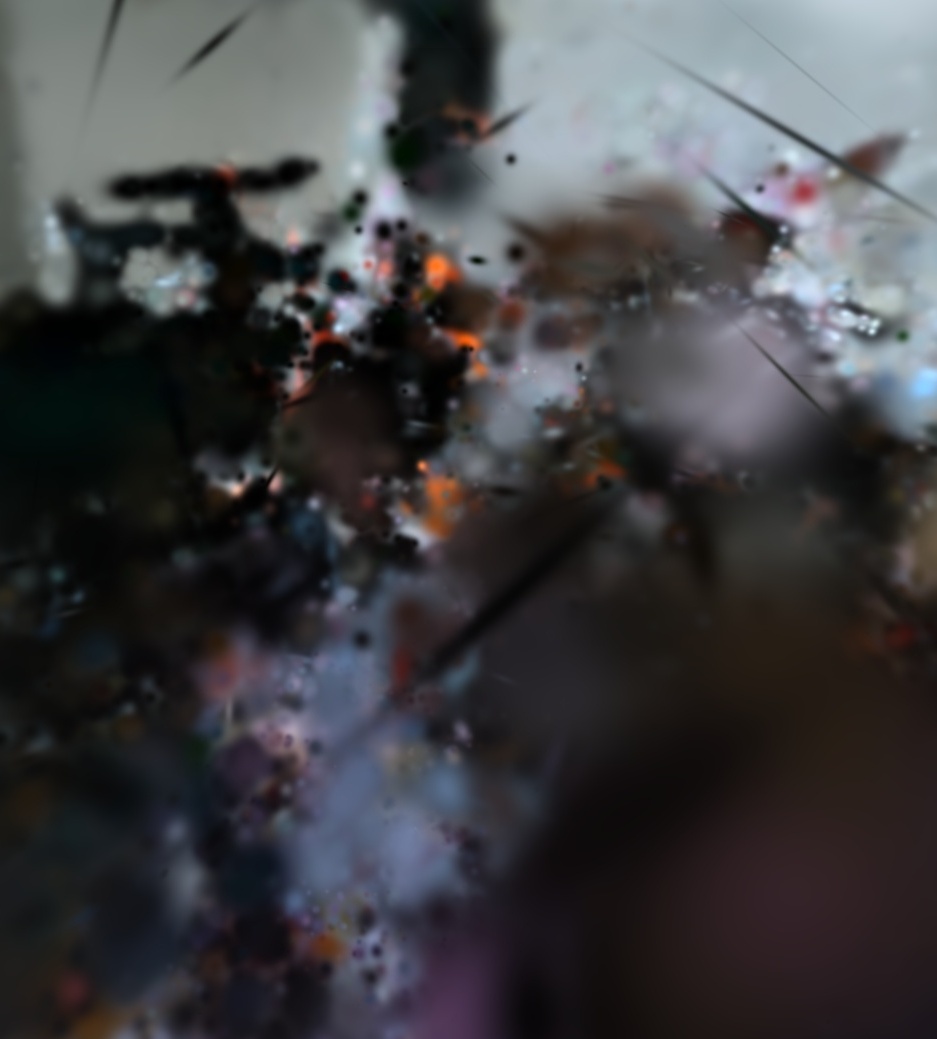}} &
    {\includegraphics[width=0.195\textwidth]{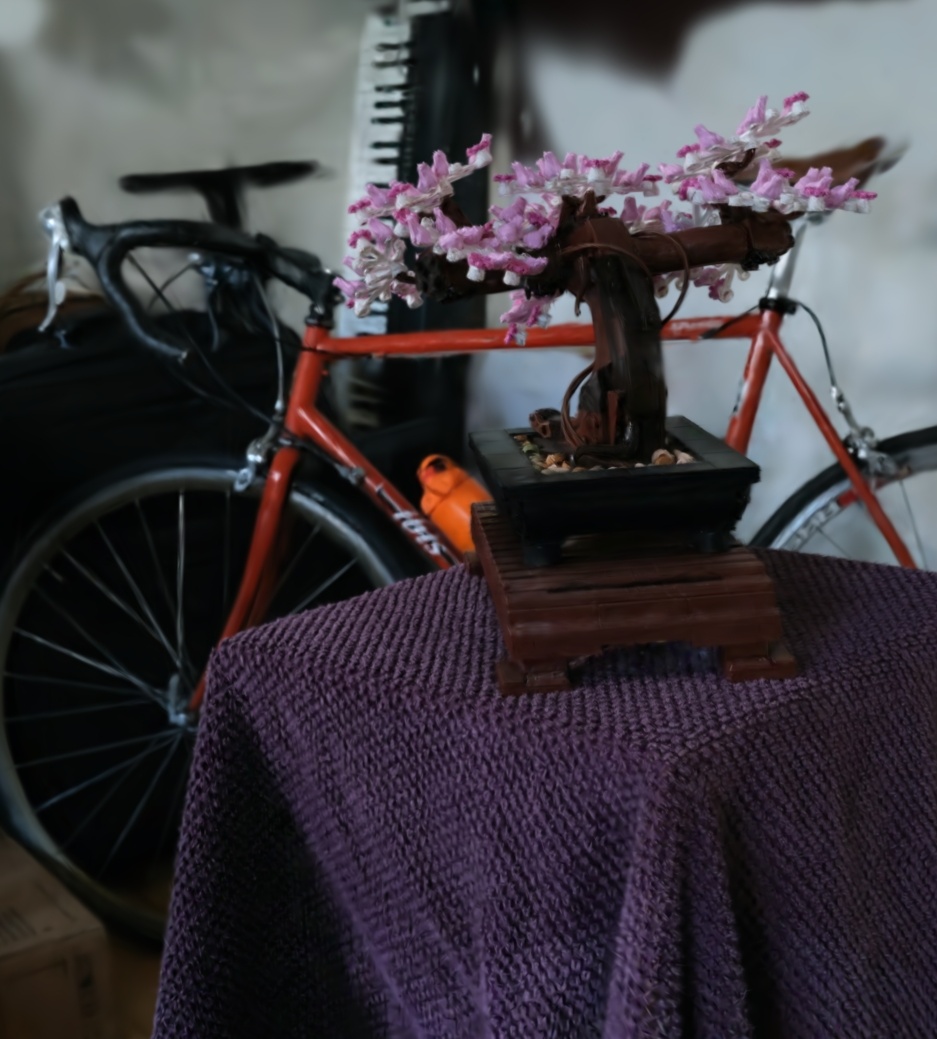}} &
    {\includegraphics[width=0.195\textwidth]{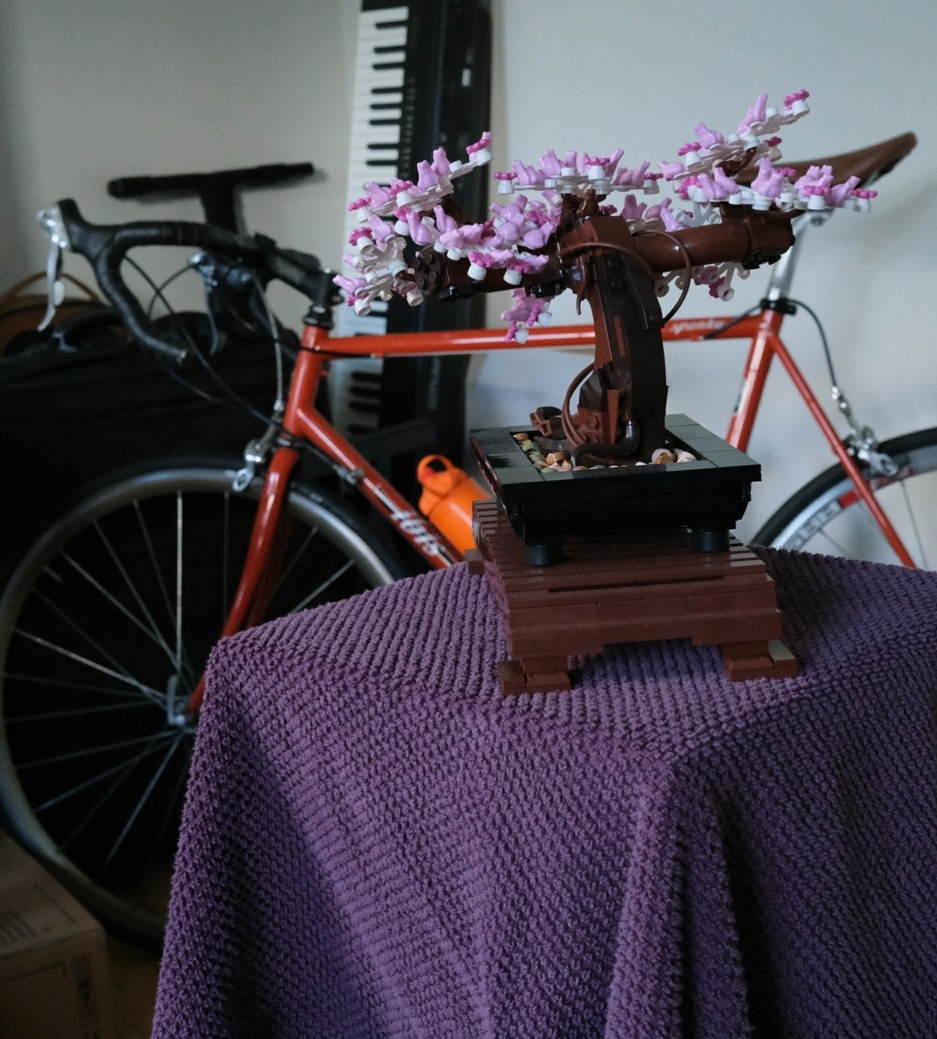}} \\[2pt]

    {\includegraphics[width=0.195\textwidth]{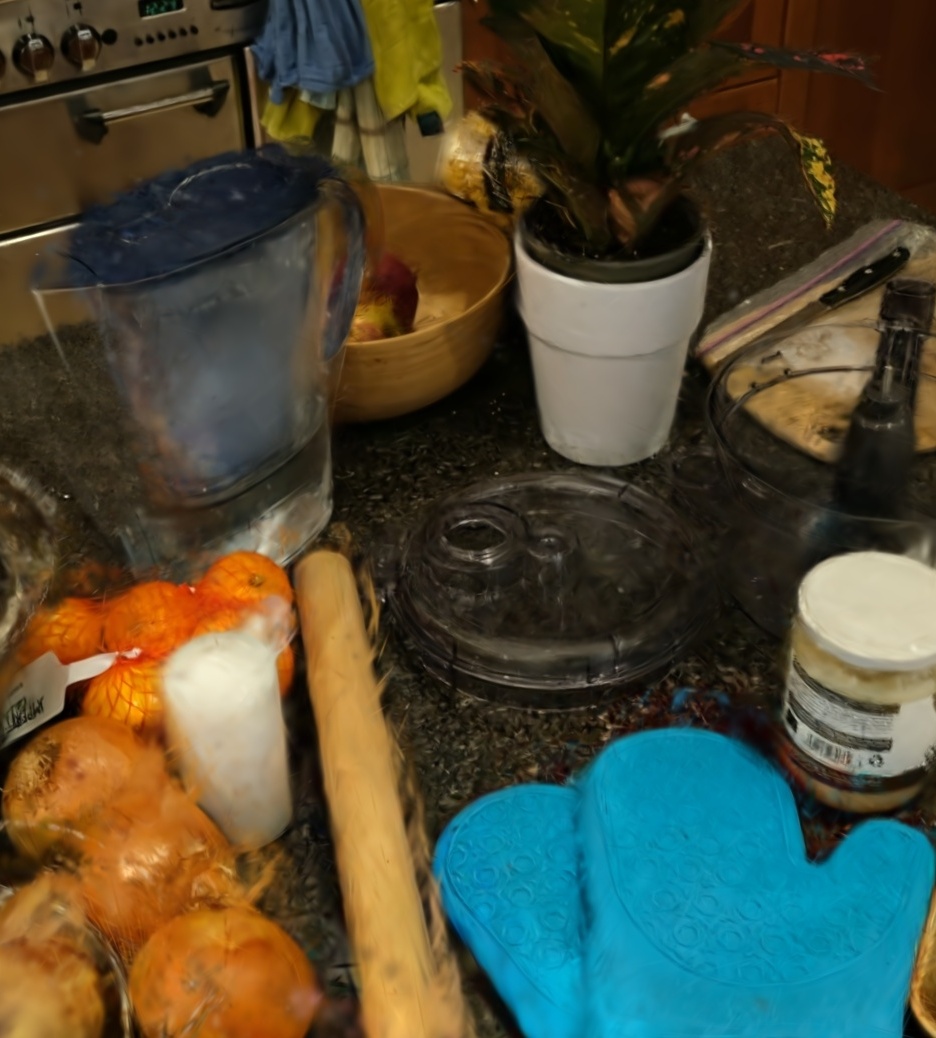}} &
    {\includegraphics[width=0.195\textwidth]{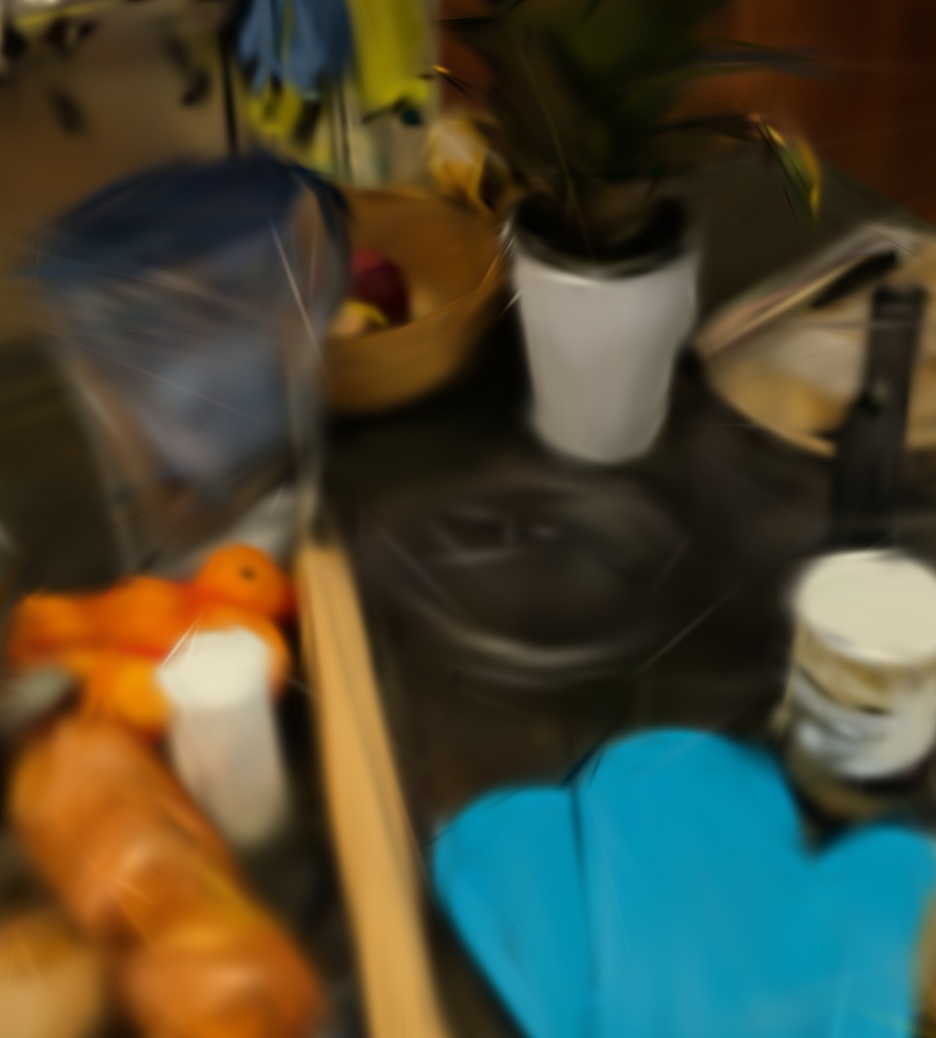}} &
    {\includegraphics[width=0.195\textwidth]{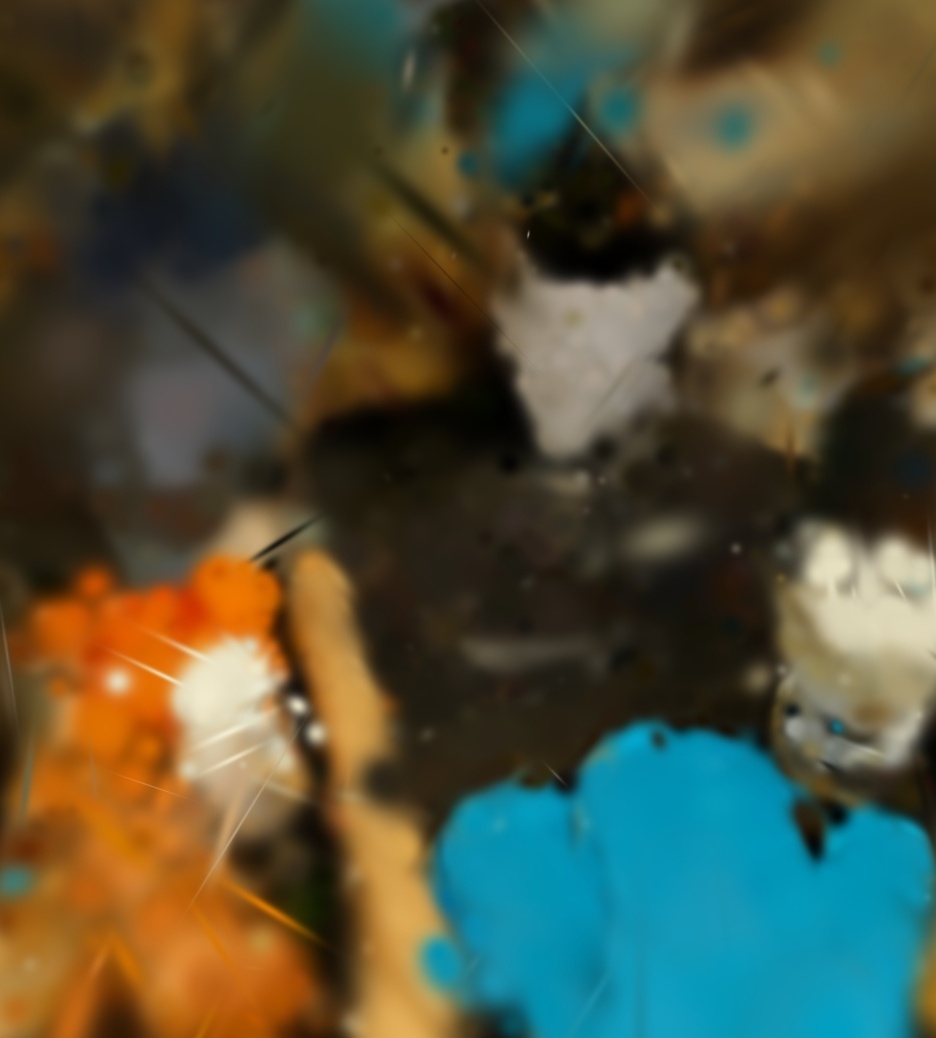}} &
    {\includegraphics[width=0.195\textwidth]{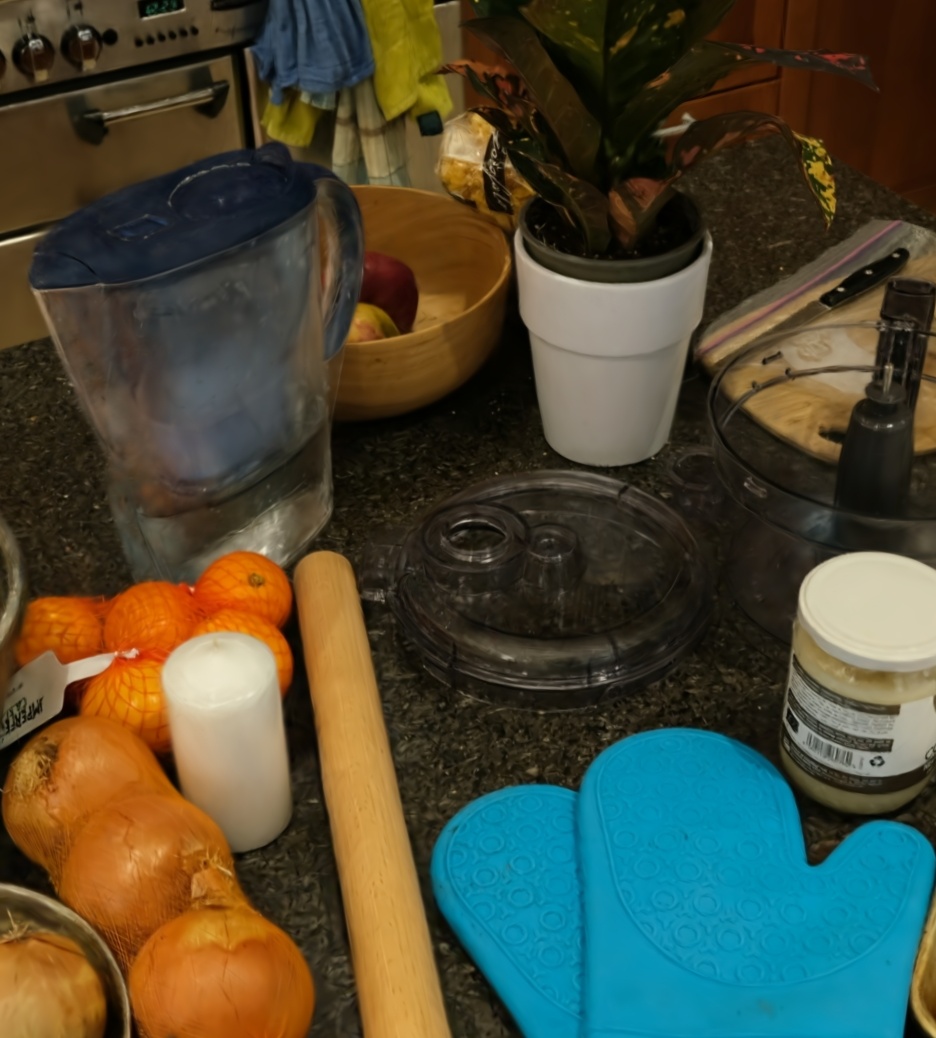}} &
    {\includegraphics[width=0.195\textwidth]{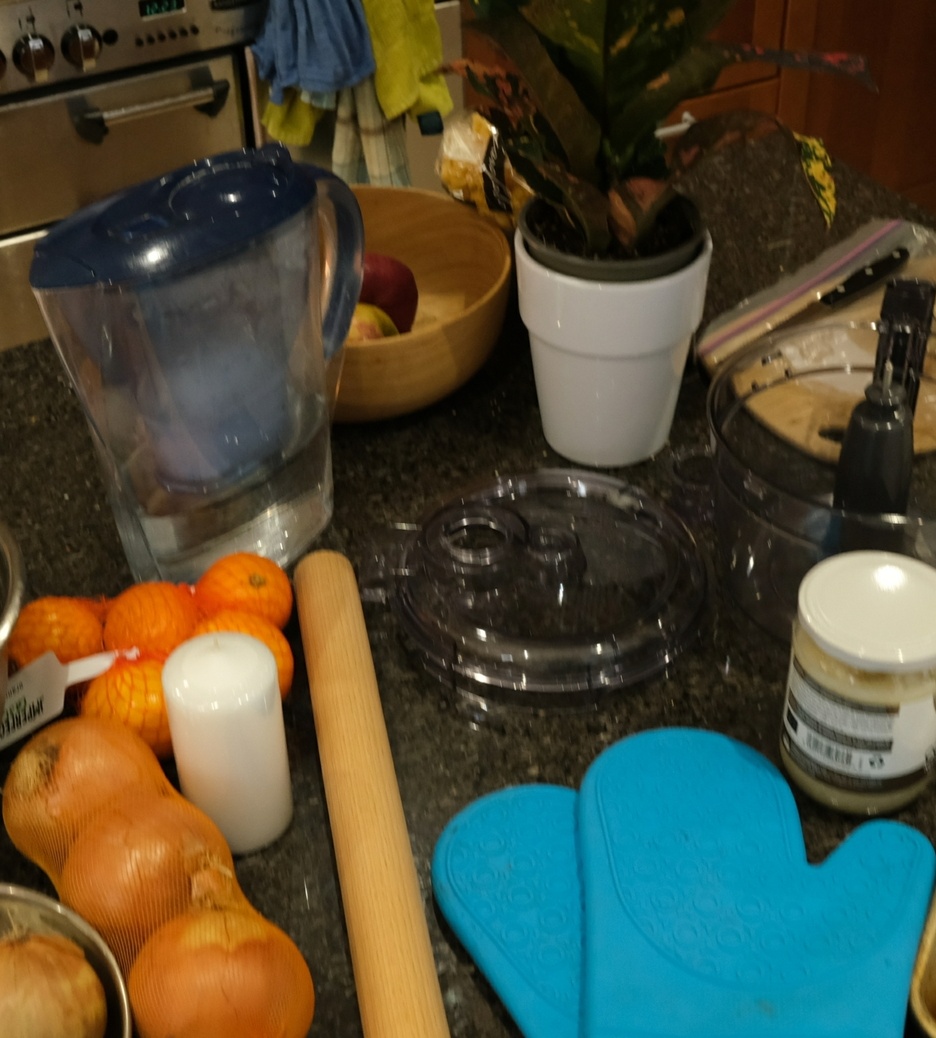}} \\[2pt]

    {\includegraphics[width=0.195\textwidth]{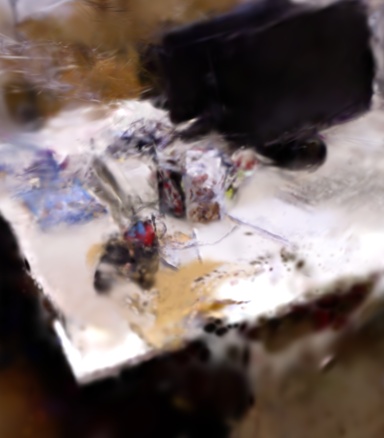}} &
    {\includegraphics[width=0.195\textwidth]{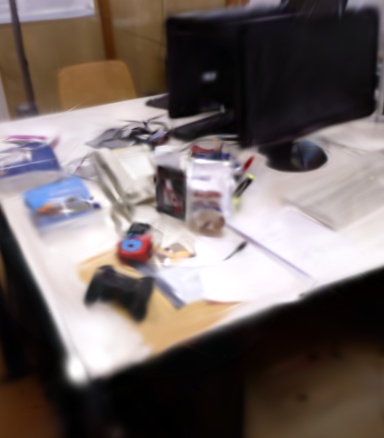}} &
    {\includegraphics[width=0.195\textwidth]{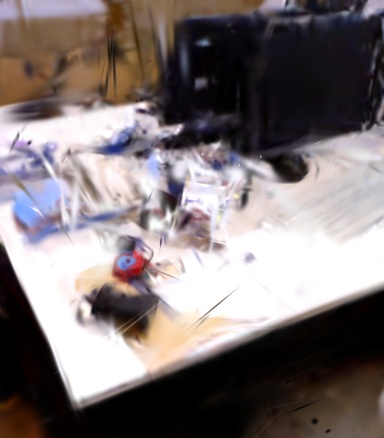}} &
    {\includegraphics[width=0.195\textwidth]{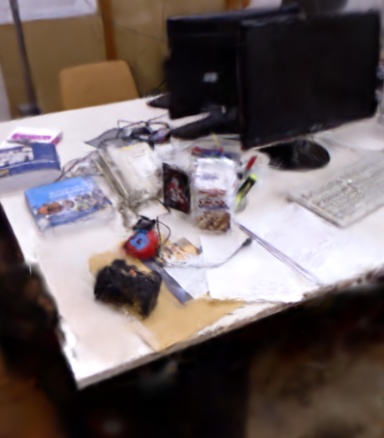}} &
    {\includegraphics[width=0.195\textwidth]{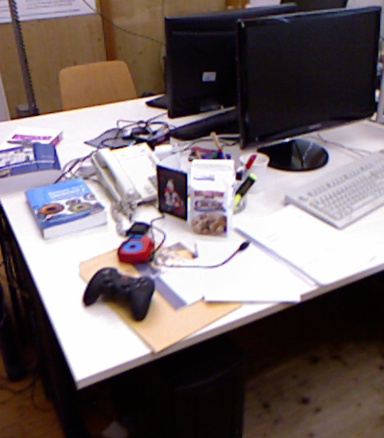}} \\[2pt]

    {\includegraphics[width=0.195\textwidth]{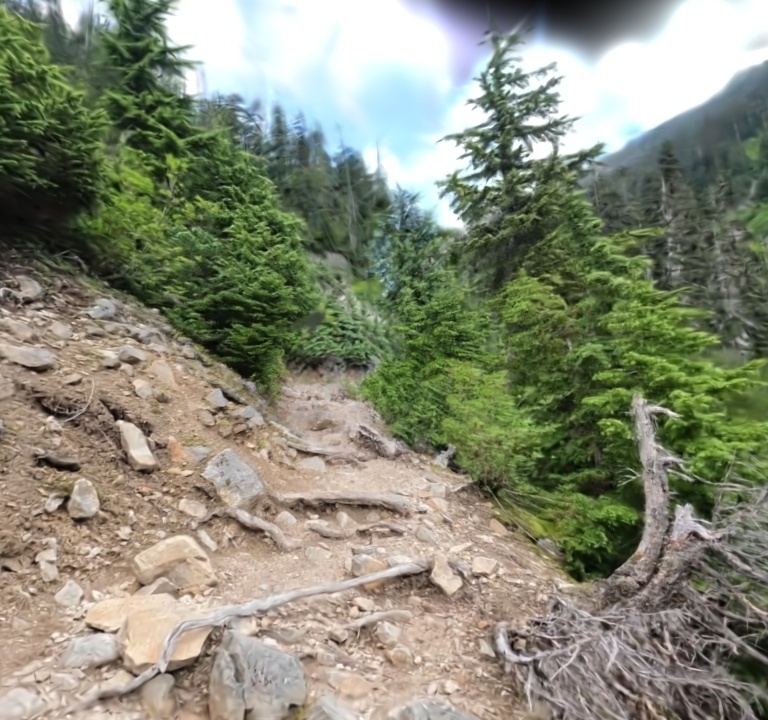}} &
    {\includegraphics[width=0.195\textwidth]{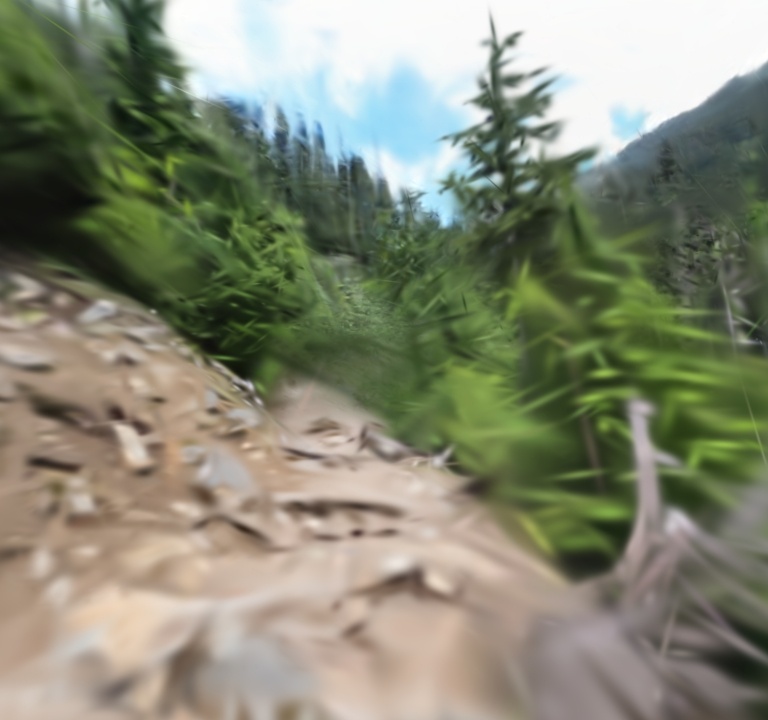}} &
    {\includegraphics[width=0.195\textwidth]{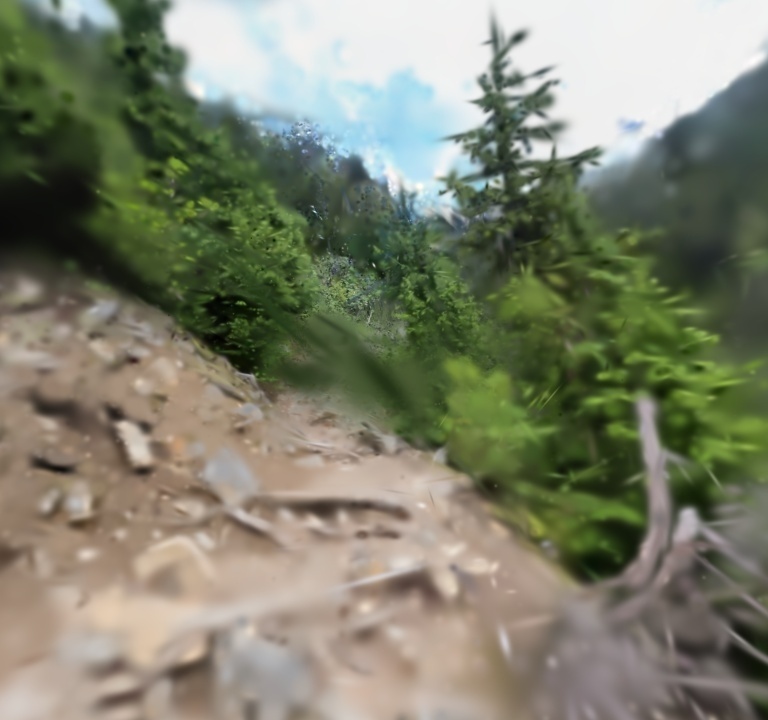}} &
    {\includegraphics[width=0.195\textwidth]{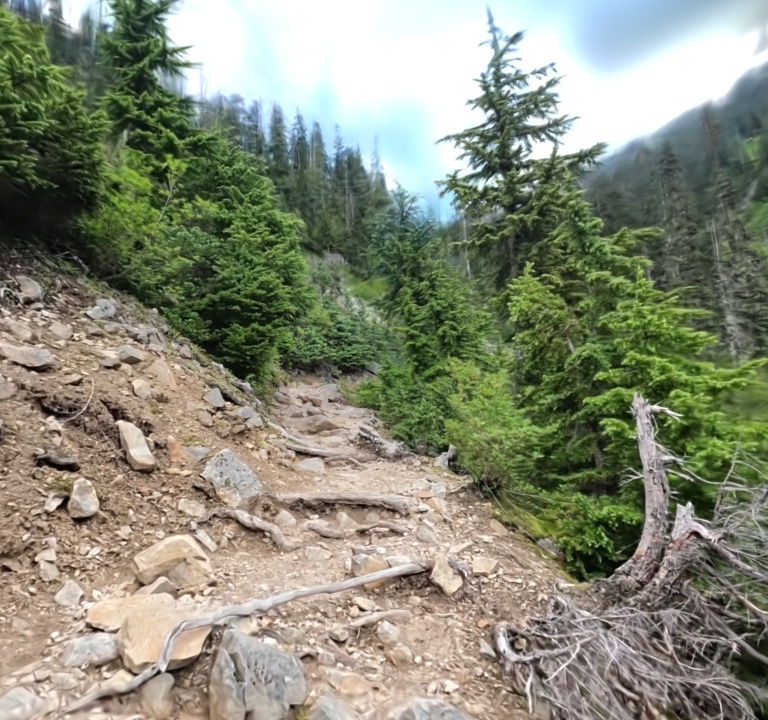}} &
    {\includegraphics[width=0.195\textwidth]{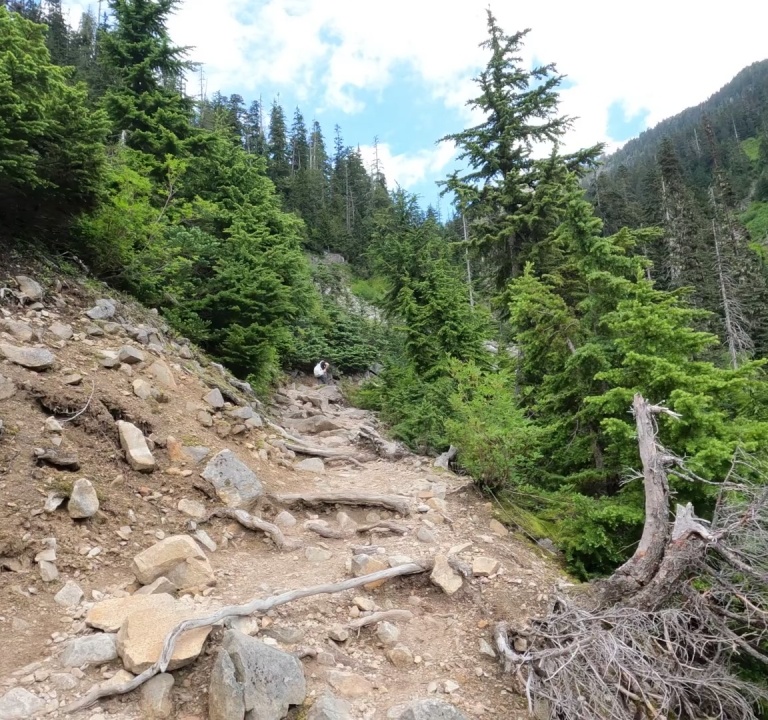}} \\[2pt]

    {\includegraphics[width=0.195\textwidth]{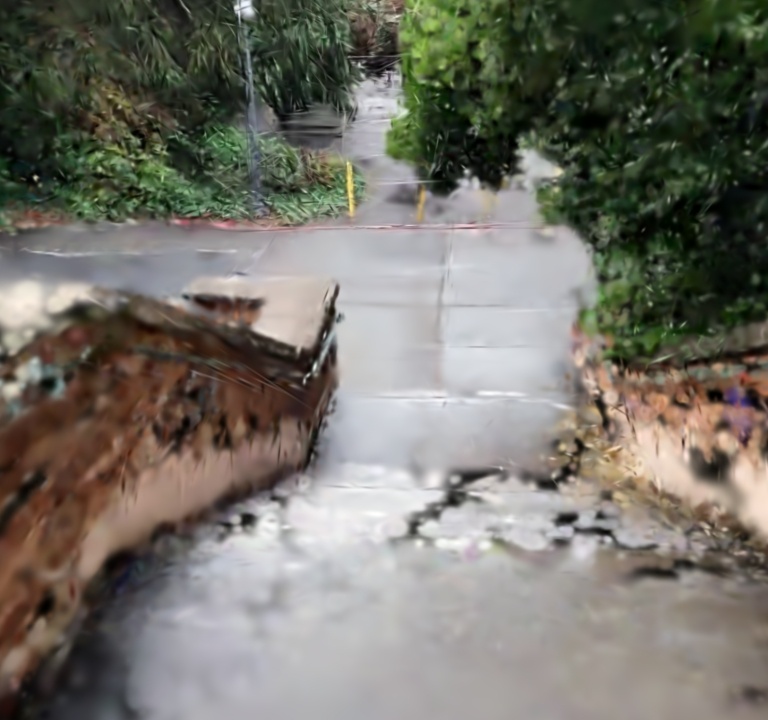}} &
    {\includegraphics[width=0.195\textwidth]{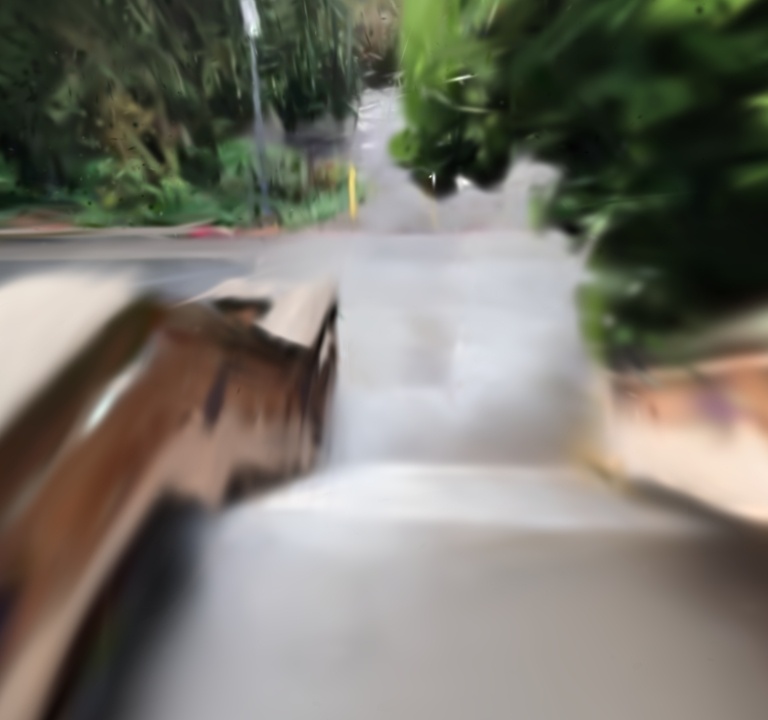}} &
    {\includegraphics[width=0.195\textwidth]{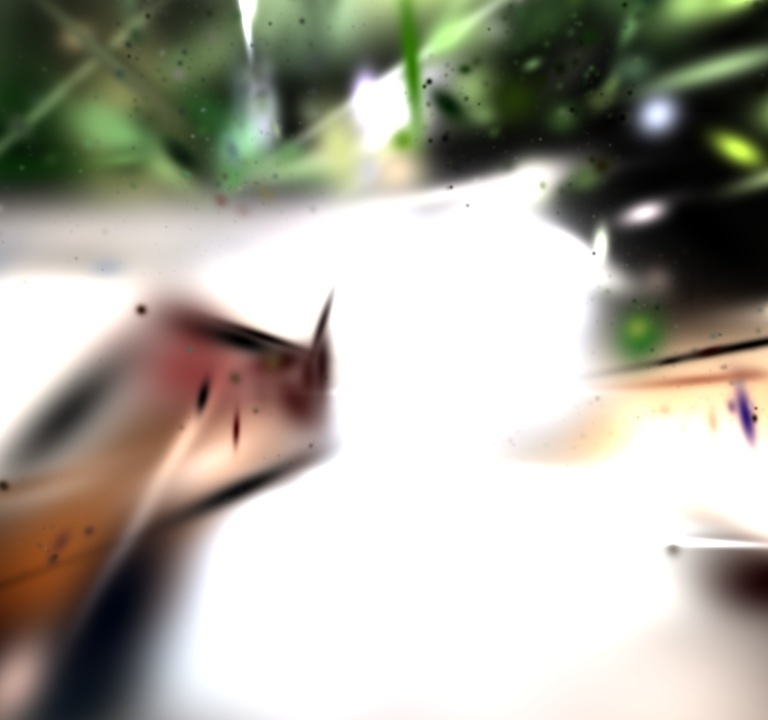}} &
    {\includegraphics[width=0.195\textwidth]{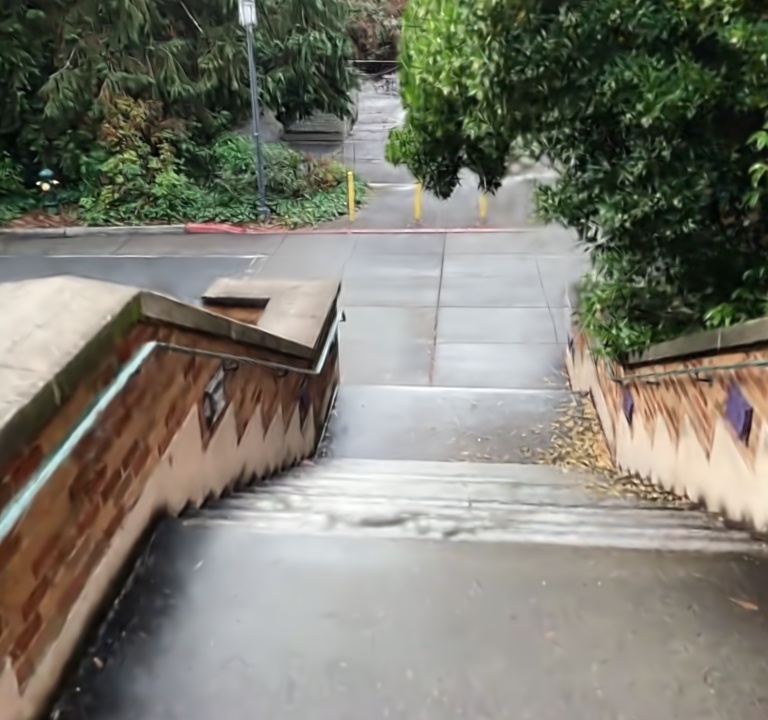}} &
    {\includegraphics[width=0.195\textwidth]{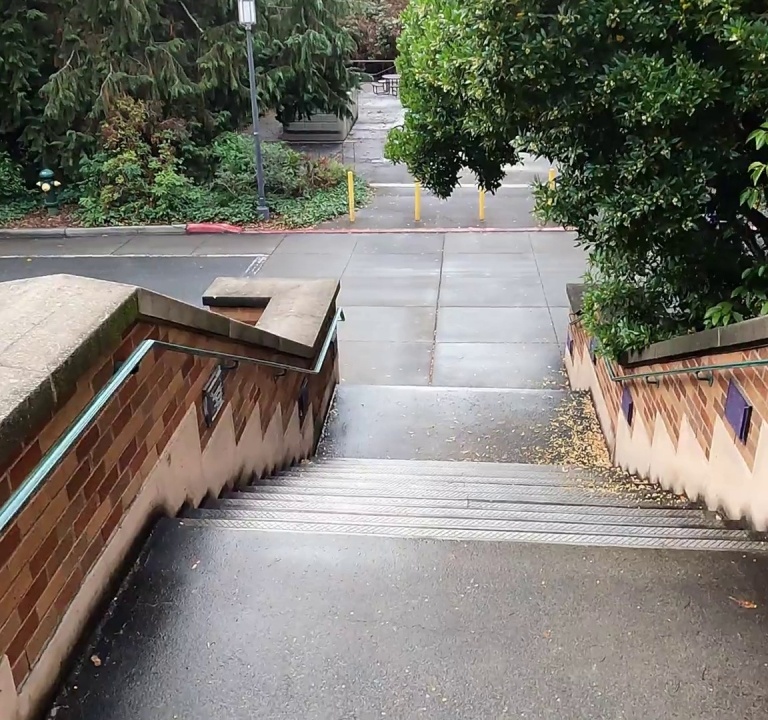}} \\[2pt]
\end{tabular}
\vspace{-9pt}
\caption{
\label{fig:evaluation}
Comparisons of our method with other pose-free solutions (see Sec.~\ref{sec:results}). We see that overall our method achieves higher visual quality, with sharper renderings across all types of scenes.
}
\end{figure*}

\begin{figure*}[t]
\setlength{\tabcolsep}{1pt}
\renewcommand{\arraystretch}{0}
\begin{tabular}{ccccc}
    OTF-NVS & S3PO-GS & Octree-GS & Ours & GT \\[2pt]        
    \includegraphics[width=0.195\textwidth]{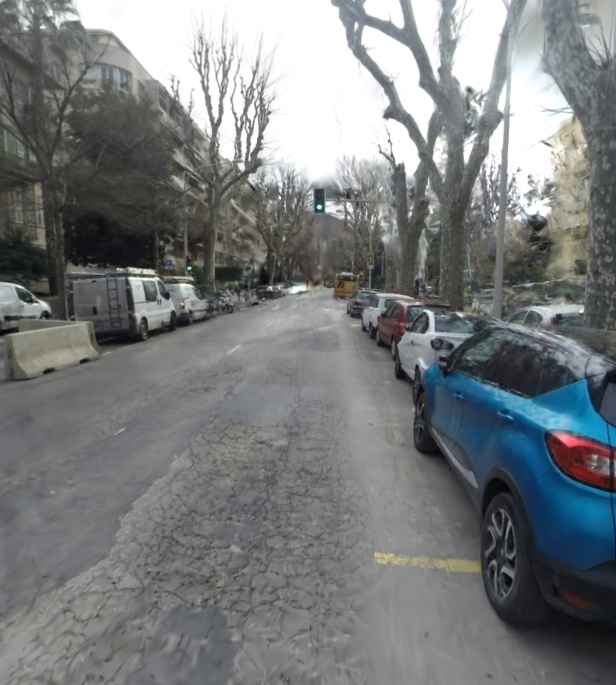} &
    \includegraphics[width=0.195\textwidth]{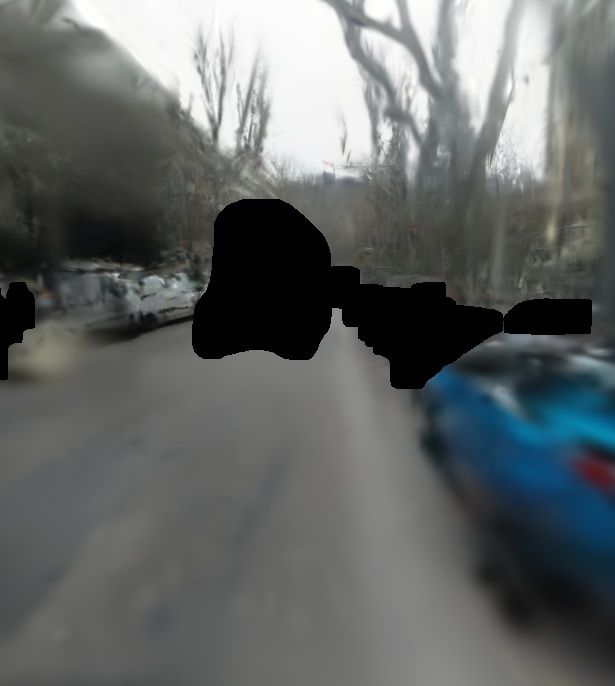} &
    \includegraphics[width=0.195\textwidth]{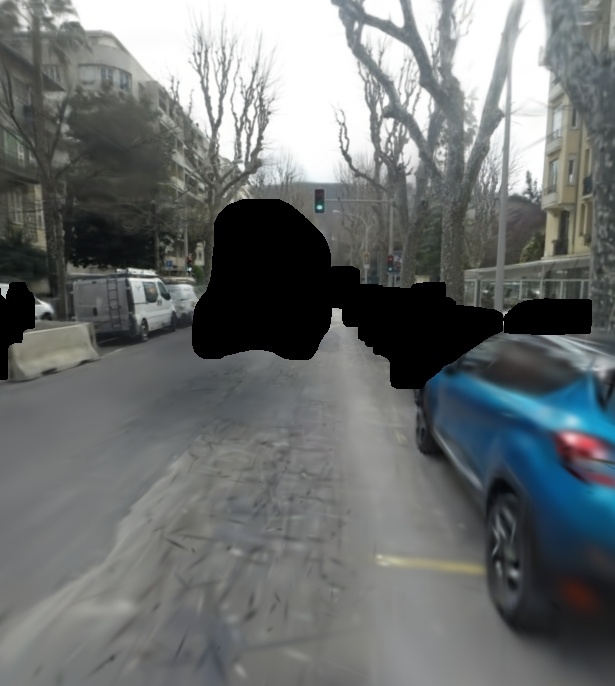} &
    \includegraphics[width=0.195\textwidth]{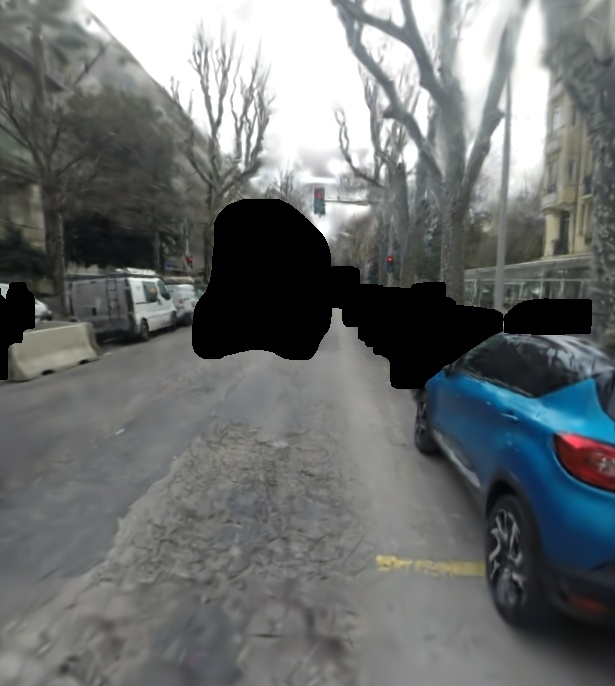} &
    \includegraphics[width=0.195\textwidth]{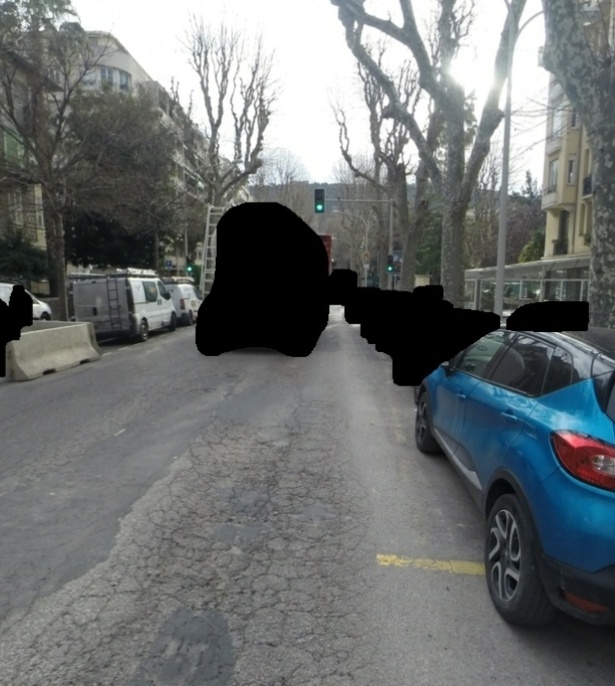} \\[2pt]

    \includegraphics[width=0.195\textwidth]{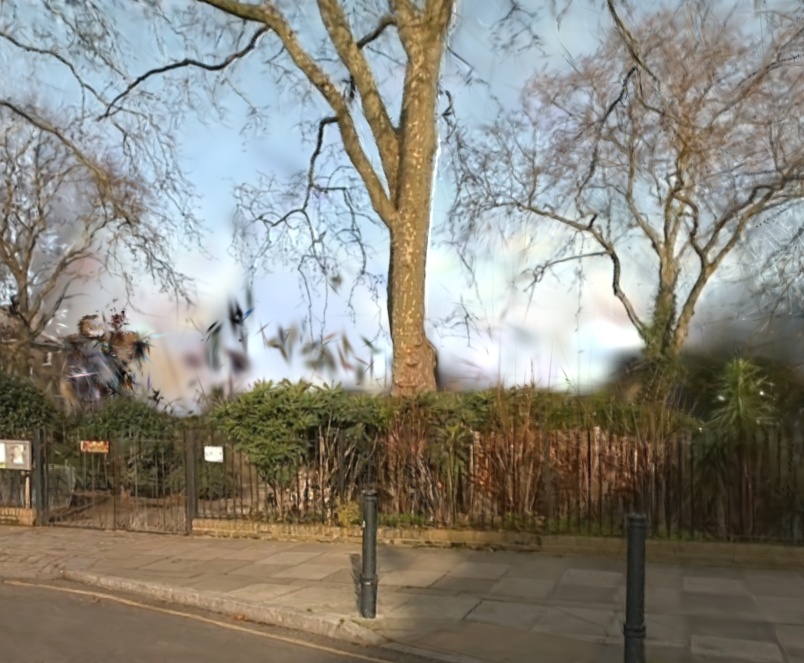} &
    \includegraphics[width=0.195\textwidth]{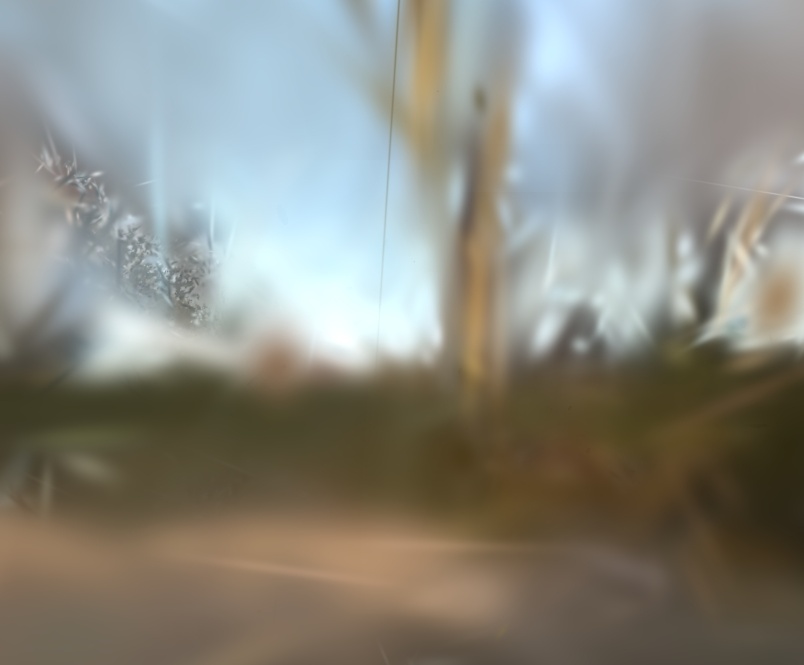} &
    \includegraphics[width=0.195\textwidth]{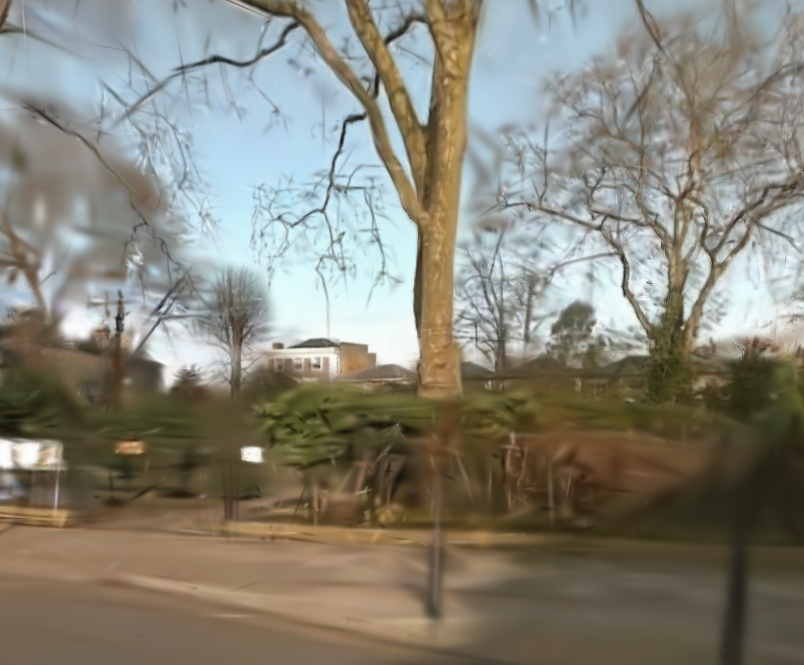} &
    \includegraphics[width=0.195\textwidth]{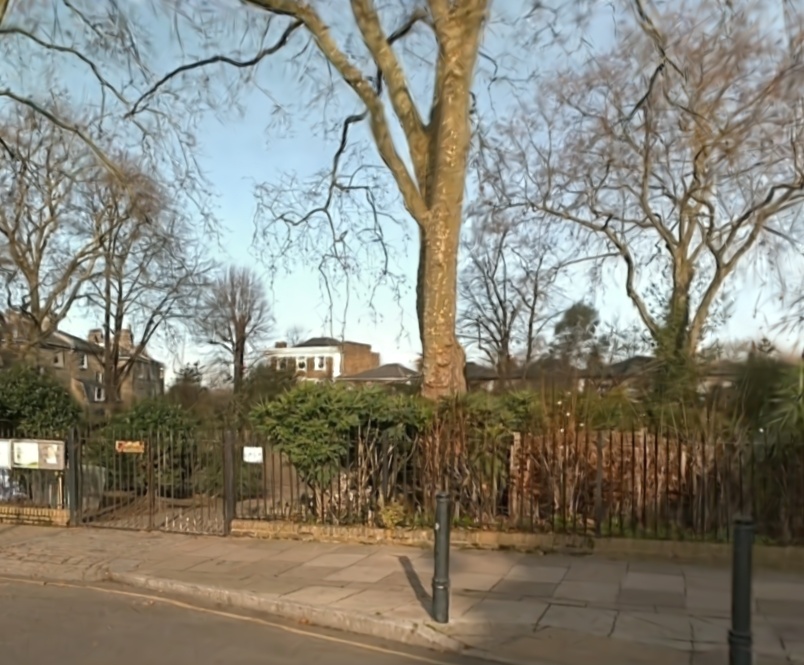} &
    \includegraphics[width=0.195\textwidth]{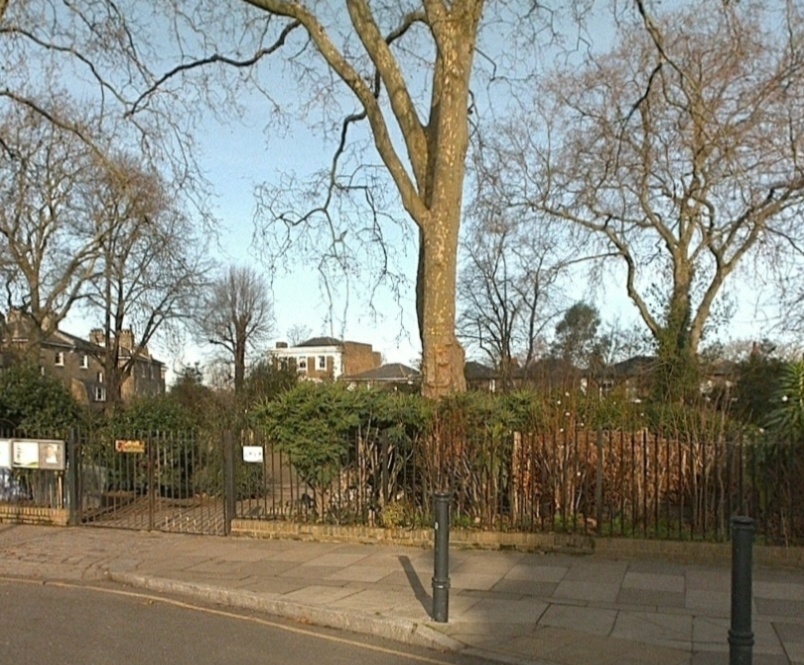} \\[2pt]

    \includegraphics[width=0.195\textwidth]{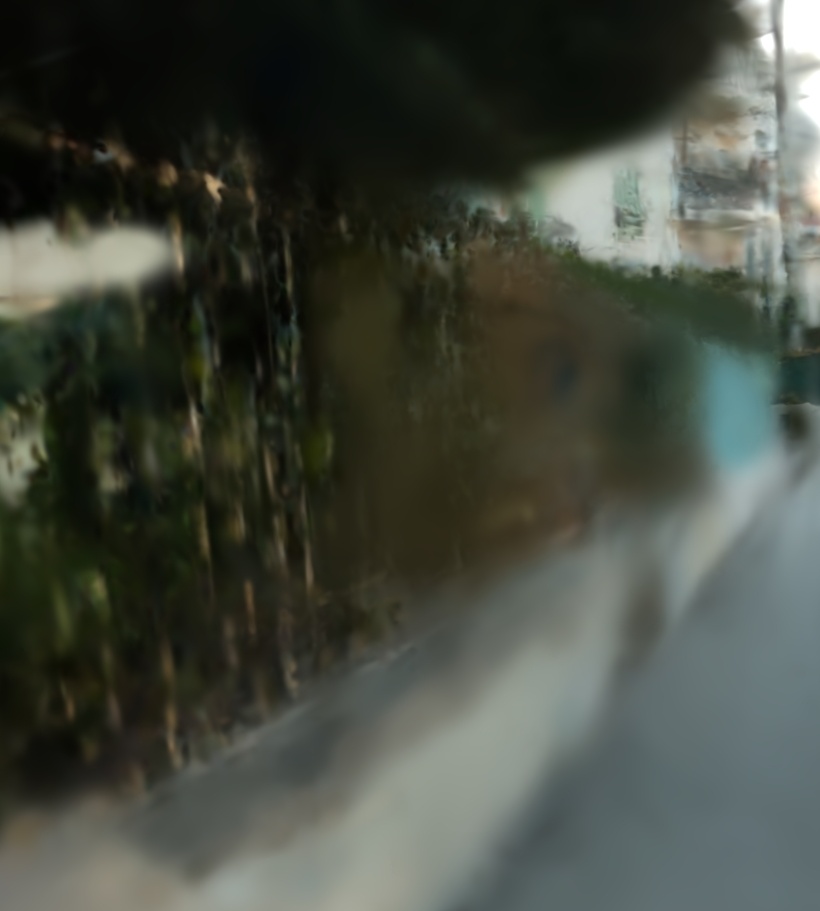} &
    \includegraphics[width=0.195\textwidth]{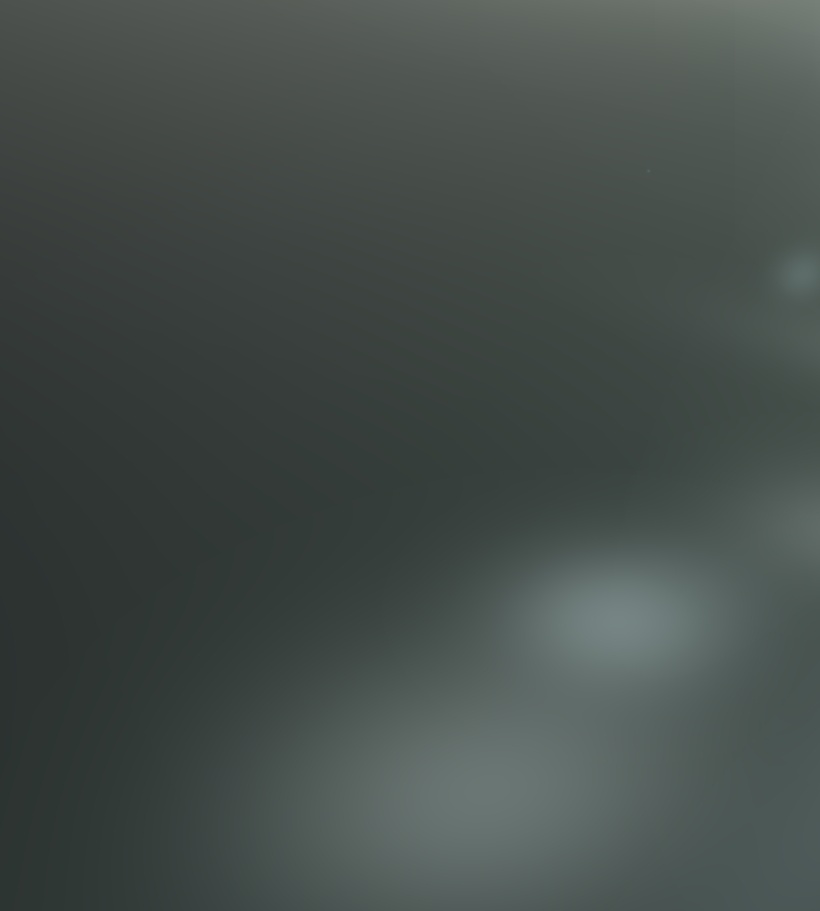} &
    \includegraphics[width=0.195\textwidth]{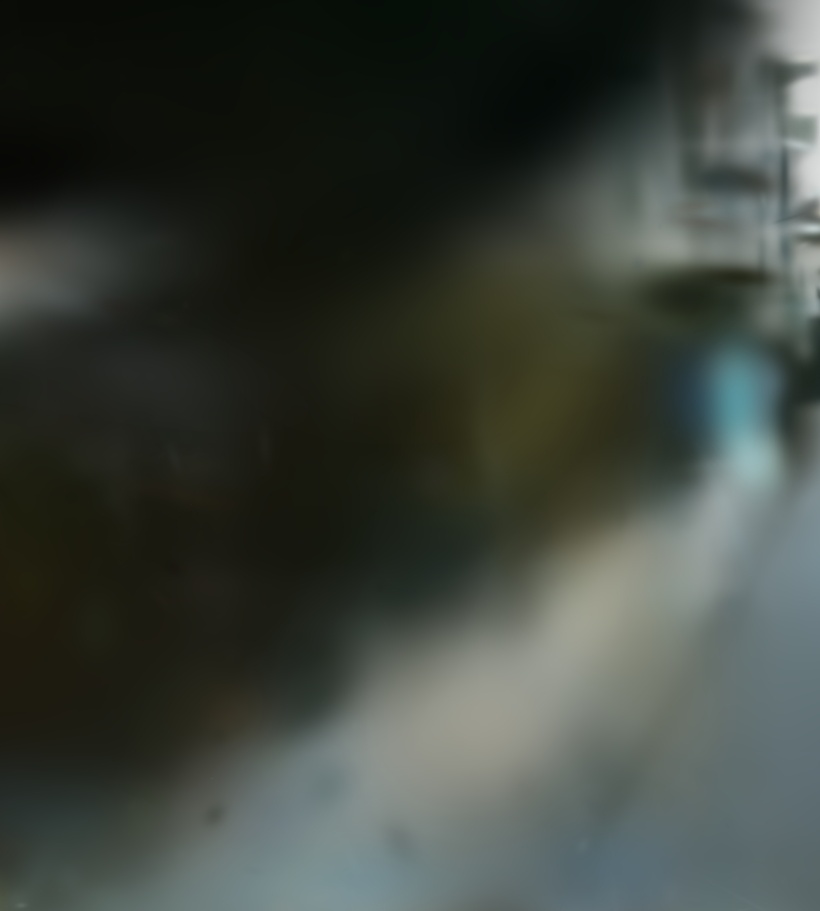} &
    \includegraphics[width=0.195\textwidth]{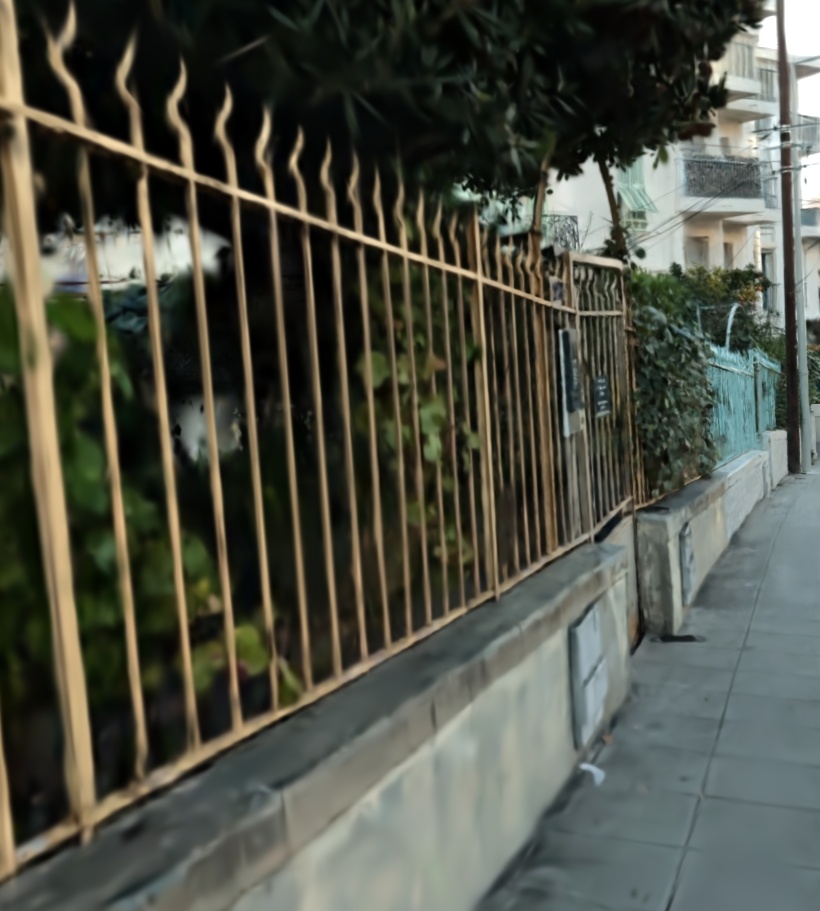} &
    \includegraphics[width=0.195\textwidth]{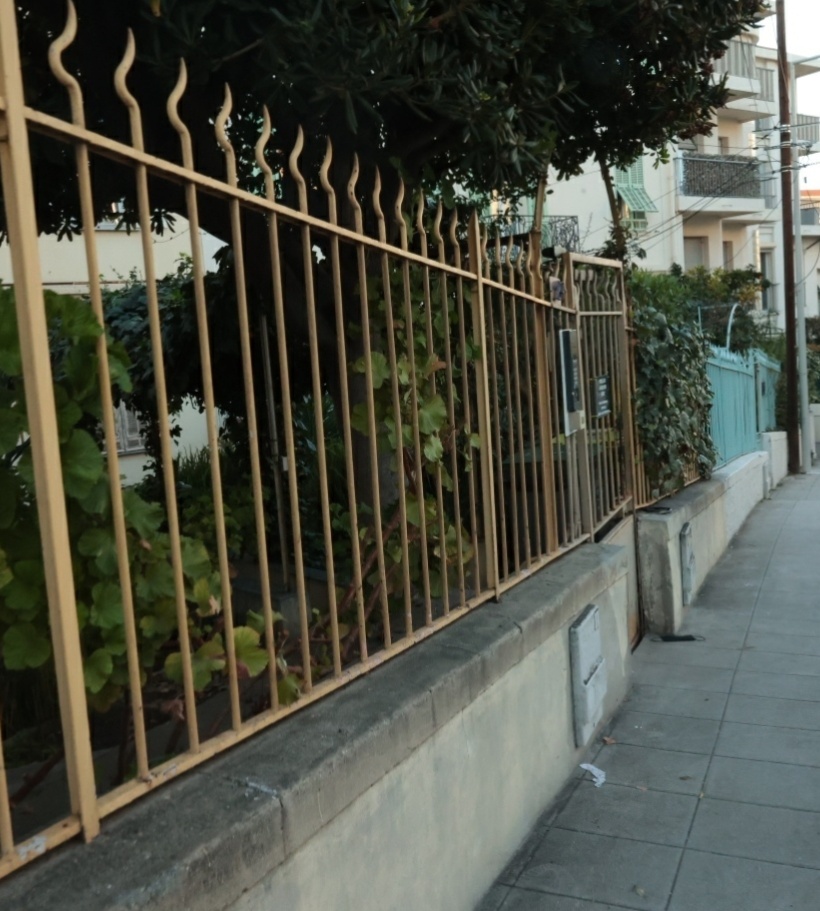} \\[2pt]

    \includegraphics[width=0.195\textwidth]{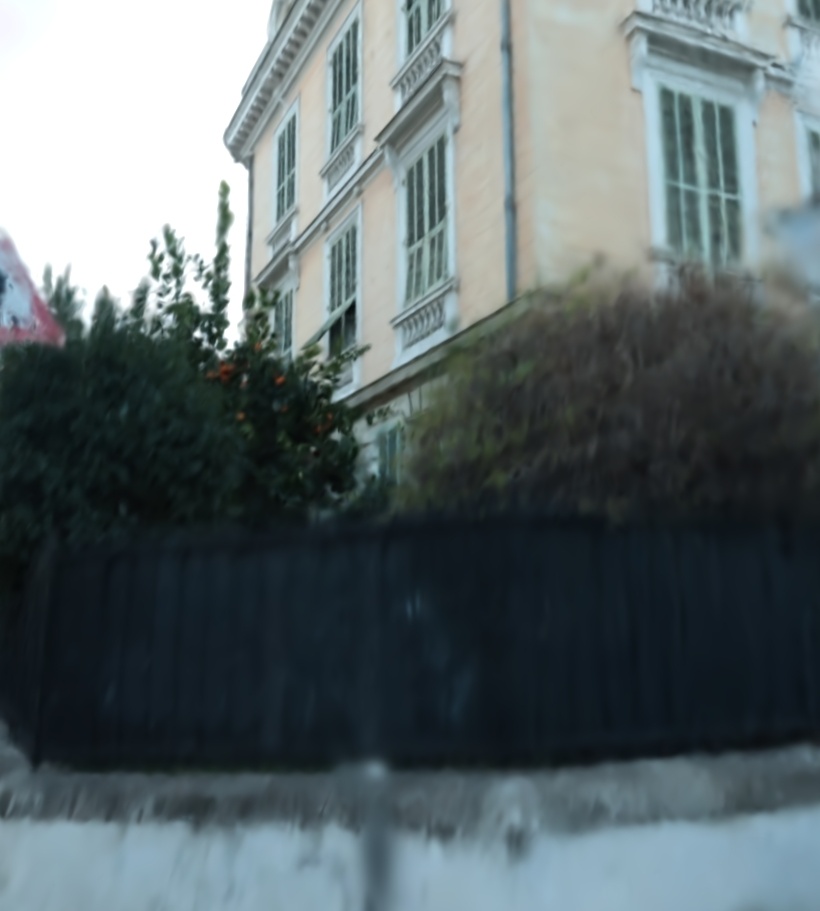} &
    \includegraphics[width=0.195\textwidth]{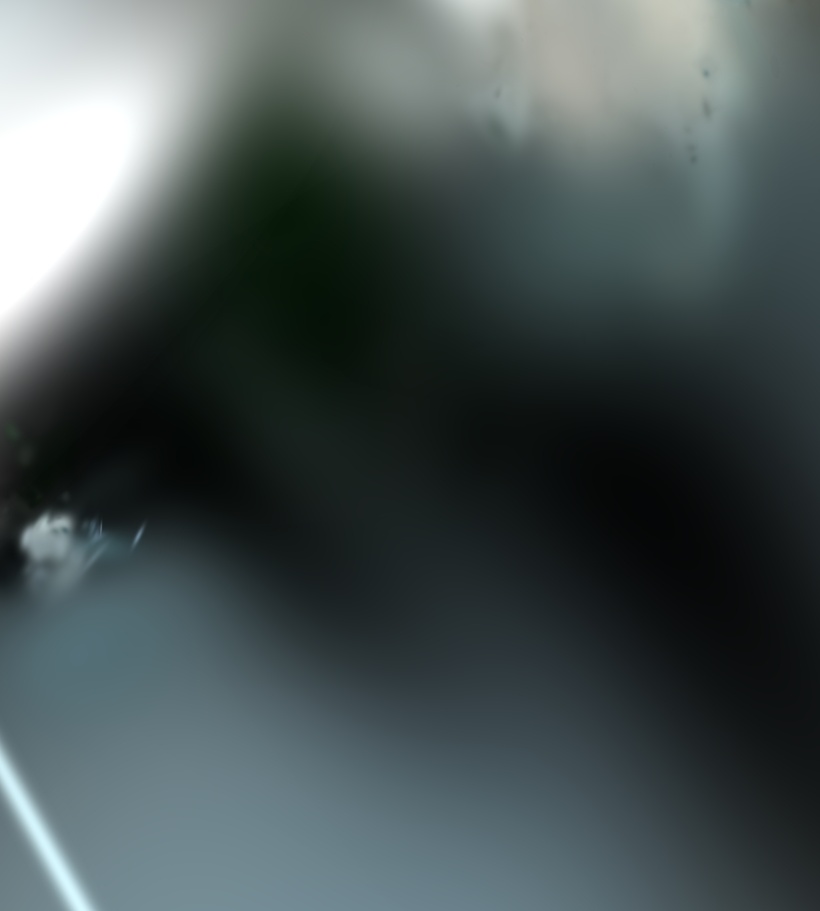} &
    \includegraphics[width=0.195\textwidth]{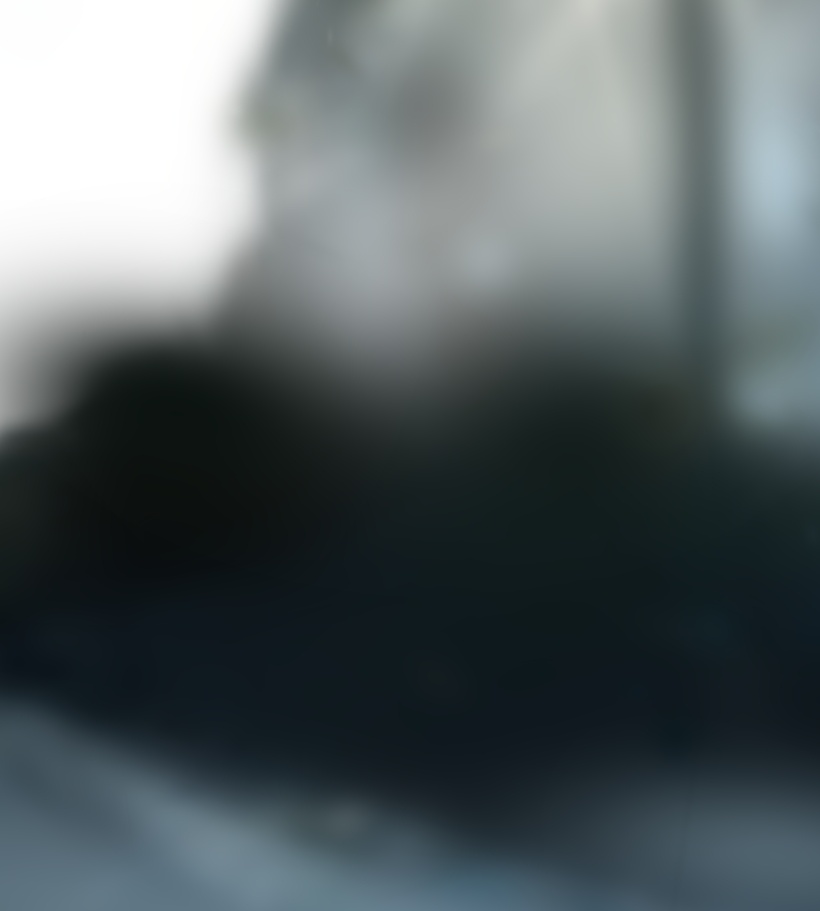} &
    \includegraphics[width=0.195\textwidth]{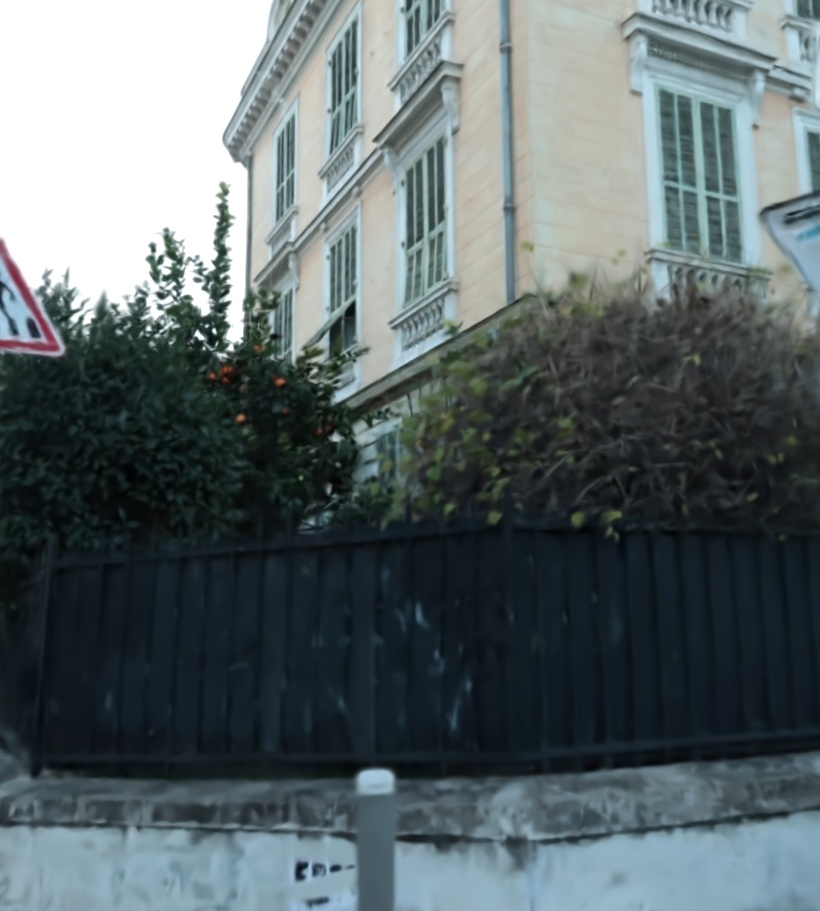} &
    \includegraphics[width=0.195\textwidth]{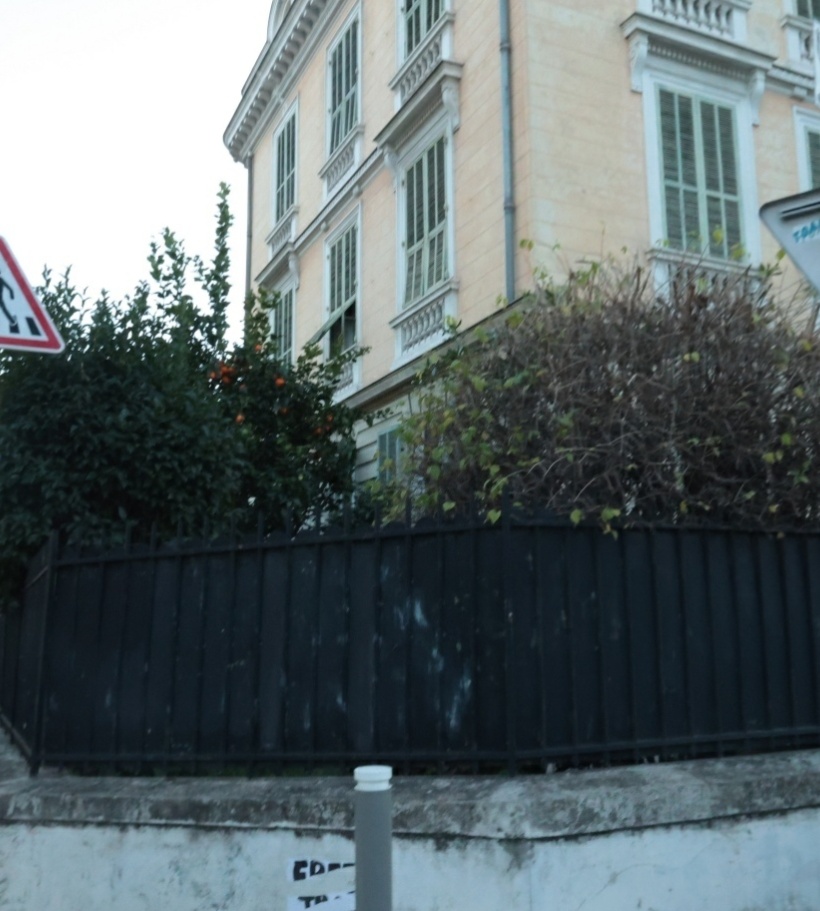} \\[2pt]
\end{tabular}
\vspace{-9pt}
\caption{
\label{fig:evaluation_large}
Comparisons of our method on the large scenes. Note that S3PO-GS and Octree-GS require a significantly larger amount of time for offline pose estimation and 3DGS optimization.
}
\end{figure*}

We present results and comparisons of our method in Fig.~\ref{fig:evaluation} and \ref{fig:evaluation_large}, illustrating that we provide fast reconstruction and maintain good visual quality; additional results can be found in the video. 

\subsection{Evaluation Methodology}

We use the datasets and evaluation protocol as proposed in \citet{meuleman2025onthefly} for Tab.~\ref{tab:slam-eval} and \ref{tab:large-scale}. We compare to the following previous methods: On-the-fly NVS~\cite{meuleman2025onthefly},
Photo-SLAM~\cite{hhuang2024photoslam}, DROID-Splat~\cite{homeyer2024droid} and MonoGS~\cite{gsslam2024}, as well as 
S3PO-GS~\cite{cheng2025outdoor}, LongSplat~\cite{lin2025longsplat} and AnySplat~\cite{jiang2025anysplat}.
AnySplat and DROID-Splat require resized images so we report these results separately in the appendix.
For large scenes we compare to \cite{meuleman2025onthefly}, H3DGS~\cite{hierarchicalgaussians24}, Octree-GS~\cite{octreeGS} and S3PO-GS~\cite{cheng2025outdoor}. We also present evaluation on out-of-order scenes from the standard datasets \textsc{MipNeRF360}~\cite{barron2022mipnerf360}, \textsc{Tanks and Temples}~\cite{tnt} and \textsc{Deep Blending}~\cite{hedman2018deep} in Tab.~\ref{tab:slam-eval-un}; here we just compare to COLMAP+3DGS, using the code from the official 3DGS repository that has TamingGS~\cite{taming3dgs} acceleration for optimization, since all other pose-free methods fail for these scenes.
We run evaluations on a workstation with an Intel Core i9 13900K CPU, 128GB of RAM, and an NVIDIA RTX 4090 GPU, or if we use a different configuration (e.g. when the method requires more GPU memory) we scale the timings to this setup by running our algorithm on Garden on each machine.

\subsection{Comparisons}
\label{sec:comparisons}

\paragraph{Ordered scenes}
Tab.~\ref{tab:slam-eval} summarizes our evaluation results. Among pose-free methods, our approach achieves faster reconstruction than all prior methods except \citet{meuleman2025onthefly}, over which we obtain superior rendering quality.
Our method outperforms all baselines in quality, with the exception of LongSplat on the \textsc{TUM} dataset; however, LongSplat requires 1h41min compared to 1min22sec for our method.
{
LongSplat's local optimization performs well at low resolution (${\sim}0.3$\,Mpx) and high frame overlap (e.g., \textsc{TUM}), but requires ${\sim}75\times$ more compute time and degrades on higher-resolution ($>1$\,Mpx), wider-baseline scenes (\textsc{MipNeRF360}, \textsc{Static\-Hikes}).
GLOMAP+\allowbreak{}Taming3DGS struggles on the larger (${\sim}1000$ frames, tens-of-meters trajectory) and more challenging \textsc{Static\-Hikes} scenes, as Taming3DGS lacks a level-of-detail scheme and adequate densification scheduling for large-scale inputs.
}
Our method also shows better quality than AnySplat, even when AnySplat is augmented with additional optimization (see appendix).
For large scenes, our method achieves superior quality across all evaluations, except for Octree-GS on the \textsc{Small\-City} scene; notably, COLMAP+\allowbreak{}Octree-GS requires 1h20min, compared to 2min25sec for our method. On \textsc{Small\-City}, our quality is comparable to OTF-NVS, though at higher computational cost.

{
We provide fine-tuning results with additional optimisation in the appendix.

20 epochs (+43s) suffice to reach GLOMAP+\allowbreak{}Taming3\-DGS (7k)'s quality.

On \textsc{MipNeRF360}, the average and maximum latency (defined as the time from start of processing to readiness for next frame) are 324 ms and 386 ms.
}

\paragraph{Pose estimation evaluation}
{
We evaluate pose estimation quality and present the results in Tab.~\ref{tab:pose-eval}. 
Spann3R~\cite{wang2025spann3r} is a transformer-based feed-forward 3D reconstruction method.
We use poses provided by \textsc{TUM} and COLMAP as pseudo ground truth for \textsc{MipNeRF360}. 
Our method achieves high accuracy in wide baseline scenes while remaining competitive with dedicated SLAM methods on small baseline benchmarks.
\begin{table}
\small
\setlength{\tabcolsep}{1.5pt}
\caption{\label{tab:pose-eval} Pose estimation results for different methods using absolute and relative error metrics. The {\colorbox{firstcolor}{best}} and {\colorbox{secondcolor}{second best}} are color coded.}
\vspace{-7pt}
{
\begin{tabular}{l|rrrr|rrrr}
\toprule
 & \multicolumn{4}{c|}{\textsc{TUM}} & \multicolumn{4}{c}{\textsc{MipNeRF360}}   \\
 & 
 \footnotesize T.APE$^\downarrow$ & 
 \footnotesize R.APE$^\downarrow$ & 
 \footnotesize T.RPE$^\downarrow$ & 
 \footnotesize R.RPE$^\downarrow$ & 
 \footnotesize T.APE$^\downarrow$ & 
 \footnotesize R.APE$^\downarrow$ & 
 \footnotesize T.RPE$^\downarrow$ & 
 \footnotesize R.RPE$^\downarrow$ \\
\midrule
Spann3R &
89.4 & 0.507 & 33.4 & 0.210 & 32.9 & 0.130 & 42.2 & 0.150 \\
DROID-Splat &
\first{1.0} & \first{0.033} & \first{1.2} & \first{0.016} & {11.7} & {0.052} & {19.1} & {0.074} \\
Photo-SLAM &
{9.0} & \second{0.034} & 8.9 & \second{0.021} & 314.0 & 2.016 & 318.9 & 1.213 \\
MonoGS &
33.5 & 0.197 & 23.3 & 0.063 & 315.5 & 2.373 & 278.3 & 0.983 \\
CF-3DGS &
73.1 & 2.817 & 15.0 & 0.187 & 161.5 & 2.777 & 59.5 & 0.197 \\
On-The-Fly &
40.2 & 0.313 & {5.7} & 0.045 & \second{11.4} & \second{0.035} & \second{16.6} & \second{0.047} \\
S3PO-GS & 14.2 & 0.102 & 3.3 & 0.075 & 126.5 & 0.585 & 157.7 & 0.312 \\
LongSplat & 13.3 & 0.124 & 4.2 & 0.037 & 30.5 & 0.112 & 24.7 & 0.082 \\
Ours & 
\second{2.6} & 0.035 & \second{2.4} & 0.028 & \first{8.5} & \first{0.026} & \first{14.2} & \first{0.041} \\
\bottomrule
\end{tabular}
}
\end{table}

}

\paragraph{Unordered scenes} 
{Tab.~\ref{tab:slam-eval-un} reports results on \textsc{MipNeRF360}, \textsc{Tanks and Temples} and \textsc{Deep Blending}, which contain a mix of ordered and unordered sequences within the same scene (e.g., \textsc{Bicycle} and \textsc{Playroom}) that the sequential comparisons above cannot handle. 
}
We achieve quality that is less than 3DGS at 7k iterations, but is still sufficiently high to be usable, as shown in the video.
Overall, the total computation time required is around 6 times lower than this offline solution, which has higher quality nonetheless.

\edef\savedparindent{\the\parindent}%
\noindent
\begin{minipage}[t]{0.5\linewidth}
\vspace{0pt}
\setlength{\parindent}{\savedparindent}%
\paragraph{Shuffled scenes}
Tab.~\ref{tab:shuffled} shows that randomly re-ordering the input images yields a reasonable quality drop on the ordered \textsc{MipNeRF360} scenes.
\end{minipage}%
\hfill
\begin{minipage}[t]{0.48\linewidth}
\vspace{0pt}
\small
\centering
\setlength{\tabcolsep}{1.5pt}
\captionof{table}{\label{tab:shuffled} Shuffling impact.}
\vspace{-9pt}
\begin{tabular}{l|ccc}
\toprule
 & PSNR$^\uparrow$ & SSIM$^\uparrow$ & LPIPS$^\downarrow$ \\
\midrule
Ordered & 26.29 & 0.839 & 0.241 \\
Shuffled & 25.73 & 0.826 & 0.238 \\
\bottomrule
\end{tabular}
\end{minipage}

\vspace{15pt}

\noindent
\begin{minipage}[t]{0.43\linewidth}
\vspace{0pt}
\subsection{Ablations}
\label{sec:ablations}
We use the ordered scenes of \textsc{MipNeRF360} to perform ablations, since some configurations we test are unsuitable for unordered sequences; the comparative advantages are equivalent in the unordered setting.
We test the following configurations: \textsc{NoVPR}, \parfillskip=0pt
\end{minipage}%
\hfill
\begin{minipage}[t]{0.54\linewidth}
\vspace{0pt}
\small
\centering
\setlength{\tabcolsep}{1.5pt}
\captionof{table}{\label{tab:abla} Ablations.}
\vspace{-9pt}
\begin{tabular}{l|ccc}
\toprule
 & PSNR$^\uparrow$ & SSIM$^\uparrow$ & LPIPS$^\downarrow$ \\
\midrule
\textsc{NoVPR} & 24.78 & 0.747 & 0.291 \\
\textsc{NoMatchVerif} & 25.36 & 0.812 & 0.257 \\
\textsc{NoKFSelection} & 24.56 & 0.733 & 0.303 \\
\textsc{NoRandFixed} & {25.78} & 0.816 & 0.254 \\
\textsc{NoGraphProp} & $\second{25.81}$ & 0.822 & 0.249 \\
\textsc{NoGSUpdate} & 25.45 & 0.814 & 0.267 \\
\textsc{NoLoDepAlgn} & 25.41 & $\second{0.823}$ & $\second{0.249}$ \\
Ours & $\first{26.29}$ & $\first{0.839}$ & $\first{0.241}$ \\
\bottomrule
\end{tabular}

\end{minipage}
\vspace{1.5pt}
\par\noindent
we match to the last five images instead of our out-of-order matching
\textsc{NoMatchVerif}, we directly use the top-K from MixVPR without brute-force matching,
\textsc{NoKFSelection}, we optimize the 20 last keyframes during local BA instead of our selection scheme,
\textsc{NoRandFixed}, we fix the oldest cameras instead of randomize,
\textsc{NoGraphProp}, we use PyPose's PGO~\cite{wang2023pypose} instead of our faster graph propagation (here we focus on speed),
\textsc{NoGSUpdate}, we do not move Gaussians after pose update in loop closure,
\textsc{NoLoDepAlgn}, 
we use global scale and offset for monocular depth alignment instead of computing it locally, per tile. \textsc{Ours} is our full solution.
From Tab.~\ref{tab:abla}, we see that MixVPR and our keyframe selection have the biggest influence on quality overall. All other components contribute to improving the quality of the result, but in a less significant manner.
{
The impact of each component on the pose quality is presented in the appendix. 
}

{
\paragraph{Impact of the hierarchy}
Hierarchy levels are created dynamically throughout the reconstruction, ranging from a single level for small scenes like \textsc{MipNeRF360} to 8 levels for \textsc{CityWalk}. 
Processing \textsc{CityWalk} without hierarchy runs OOM (24GB GPU) after 795 out of 4050 frames (less than 20\%) with 6.7 million Gaussians, while the hierarchy maintains the peak number of trained Gaussians below 3.7 million and allocated memory below 12.8 GB, improving training speed and enabling complete processing.
See Tab. \ref{tab:hierarchy-mem} in the appendix for more details.
}

\section{Limitations}
Our method trades a small quality gap vs. offline approaches for orders-of-magnitude faster processing. 
Pure rotations prevent triangulation, a limitation shared with methods from COLMAP to ORB-SLAM. 
While we show that graph-based propagation is not worse than global PGO (\textsc{NoGraphProp} ablation), it does not match slower global BA (e.g., GLOMAP). 
Failure cases are illustrated in Fig.~\ref{fig:limitations} of the appendix.

\section{Conclusion}

We have presented a method that enables immediate feedback for 3DGS reconstruction from unordered sequences typical of radiance field capture.
To efficiently retrieve keyframes that may lie far back in capture time, we repurpose visual place recognition together with the covisibility graph traditionally used in SLAM for loop closure. We combine this with GPU optimization to perform fast and robust local bundle adjustment over substantially larger windows than in previous work.
For loop closure, we propose a clustering approach suited to unordered input, leveraging the graph to perform optimization-free updates of both camera poses and Gaussian primitives.
Finally, a lightweight progressive hierarchy allows our method to scale to large scenes. Together, these contributions yield the first complete solution that makes immediate feedback possible for high-quality radiance field capture.

\begin{acks}
This work was funded by the European Research Council (ERC) Advanced Grant NERPHYS, number 101141721 \url{https://project.inria.fr/nerphys}.
Views and opinions expressed are however those of the authors only and do not necessarily reflect those of the EU or the European Research Council. 
Neither the EU nor the granting authority can be held responsible for them. 
The authors thank Adobe and NVIDIA for donations. 
Experiments presented in this paper were carried out using the Grid'5000 testbed, supported by a scientific interest group hosted by Inria and including CNRS, RENATER and several Universities and other organizations (\url{https://www.grid5000.fr}).
This work was granted access to the HPC resources of IDRIS under the allocation AD011015561R1 made by GENCI.
\end{acks}

\bibliographystyle{ACM-Reference-Format}
\bibliography{references}


\begin{thebibliography}{69}


\ifx \showCODEN    \undefined \def \showCODEN     #1{\unskip}     \fi
\ifx \showDOI      \undefined \def \showDOI       #1{#1}\fi
\ifx \showISBNx    \undefined \def \showISBNx     #1{\unskip}     \fi
\ifx \showISBNxiii \undefined \def \showISBNxiii  #1{\unskip}     \fi
\ifx \showISSN     \undefined \def \showISSN      #1{\unskip}     \fi
\ifx \showLCCN     \undefined \def \showLCCN      #1{\unskip}     \fi
\ifx \shownote     \undefined \def \shownote      #1{#1}          \fi
\ifx \showarticletitle \undefined \def \showarticletitle #1{#1}   \fi
\ifx \showURL      \undefined \def \showURL       {\relax}        \fi
\providecommand\bibfield[2]{#2}
\providecommand\bibinfo[2]{#2}
\providecommand\natexlab[1]{#1}
\providecommand\showeprint[2][]{arXiv:#2}

\bibitem[Agarwal et~al\mbox{.}(2023)]%
        {ceres}
\bibfield{author}{\bibinfo{person}{Sameer Agarwal}, \bibinfo{person}{Keir
  Mierle}, {and} \bibinfo{person}{The Ceres~Solver Team}.}
  \bibinfo{year}{2023}\natexlab{}.
\newblock \bibinfo{title}{{Ceres Solver}}.
\newblock
\newblock
\urldef\tempurl%
\url{https://github.com/ceres-solver/ceres-solver}
\showURL{%
\tempurl}


\bibitem[Ali-bey et~al\mbox{.}(2023)]%
        {ali2023mixvpr}
\bibfield{author}{\bibinfo{person}{Amar Ali-bey}, \bibinfo{person}{Brahim
  Chaib-draa}, {and} \bibinfo{person}{Philippe Gigu{\`e}re}.}
  \bibinfo{year}{2023}\natexlab{}.
\newblock \showarticletitle{{MixVPR}: Feature Mixing for Visual Place
  Recognition}. In \bibinfo{booktitle}{\emph{Proceedings of the IEEE/CVF Winter
  Conference on Applications of Computer Vision}}. \bibinfo{pages}{2998--3007}.
\newblock


\bibitem[Barron et~al\mbox{.}(2021)]%
        {barron2021mipnerf}
\bibfield{author}{\bibinfo{person}{Jonathan~T. Barron}, \bibinfo{person}{Ben
  Mildenhall}, \bibinfo{person}{Matthew Tancik}, \bibinfo{person}{Peter
  Hedman}, \bibinfo{person}{Ricardo Martin-Brualla}, {and}
  \bibinfo{person}{Pratul~P. Srinivasan}.} \bibinfo{year}{2021}\natexlab{}.
\newblock \showarticletitle{Mip-NeRF: A Multiscale Representation for
  Anti-Aliasing Neural Radiance Fields}.
\newblock \bibinfo{journal}{\emph{ICCV}} (\bibinfo{year}{2021}).
\newblock


\bibitem[Barron et~al\mbox{.}(2022)]%
        {barron2022mipnerf360}
\bibfield{author}{\bibinfo{person}{Jonathan~T. Barron}, \bibinfo{person}{Ben
  Mildenhall}, \bibinfo{person}{Dor Verbin}, \bibinfo{person}{Pratul~P.
  Srinivasan}, {and} \bibinfo{person}{Peter Hedman}.}
  \bibinfo{year}{2022}\natexlab{}.
\newblock \showarticletitle{Mip-NeRF 360: Unbounded Anti-Aliased Neural
  Radiance Fields}.
\newblock \bibinfo{journal}{\emph{CVPR}} (\bibinfo{year}{2022}).
\newblock


\bibitem[Barron et~al\mbox{.}(2023)]%
        {barron2023zipnerf}
\bibfield{author}{\bibinfo{person}{Jonathan~T. Barron}, \bibinfo{person}{Ben
  Mildenhall}, \bibinfo{person}{Dor Verbin}, \bibinfo{person}{Pratul~P.
  Srinivasan}, {and} \bibinfo{person}{Peter Hedman}.}
  \bibinfo{year}{2023}\natexlab{}.
\newblock \showarticletitle{Zip-NeRF: Anti-Aliased Grid-Based Neural Radiance
  Fields}.
\newblock \bibinfo{journal}{\emph{ICCV}} (\bibinfo{year}{2023}).
\newblock


\bibitem[Bavle et~al\mbox{.}(2025)]%
        {sgraphs2}
\bibfield{author}{\bibinfo{person}{Hriday Bavle}, \bibinfo{person}{Jose~Luis
  Sanchez-Lopez}, \bibinfo{person}{Muhammad Shaheer}, \bibinfo{person}{Javier
  Civera}, {and} \bibinfo{person}{Holger Voos}.}
  \bibinfo{year}{2025}\natexlab{}.
\newblock \showarticletitle{S-Graphs 2.0 - A Hierarchical-Semantic Optimization
  and Loop Closure for SLAM}.
\newblock \bibinfo{journal}{\emph{IEEE Robotics and Automation Letters}}
  (\bibinfo{year}{2025}).
\newblock


\bibitem[Chan et~al\mbox{.}(2023)]%
        {chan2023generative}
\bibfield{author}{\bibinfo{person}{Eric~R Chan}, \bibinfo{person}{Koki Nagano},
  \bibinfo{person}{Matthew~A Chan}, \bibinfo{person}{Alexander~W Bergman},
  \bibinfo{person}{Jeong~Joon Park}, \bibinfo{person}{Axel Levy},
  \bibinfo{person}{Miika Aittala}, \bibinfo{person}{Shalini De~Mello},
  \bibinfo{person}{Tero Karras}, {and} \bibinfo{person}{Gordon Wetzstein}.}
  \bibinfo{year}{2023}\natexlab{}.
\newblock \showarticletitle{Generative novel view synthesis with 3d-aware
  diffusion models}. In \bibinfo{booktitle}{\emph{ICCV}}.
\newblock


\bibitem[Chen and Wang(2024)]%
        {chen2024survey3dgs}
\bibfield{author}{\bibinfo{person}{Guikun Chen} {and} \bibinfo{person}{Wenguan
  Wang}.} \bibinfo{year}{2024}\natexlab{}.
\newblock \bibinfo{title}{A Survey on 3D Gaussian Splatting}.
\newblock
\newblock
\showeprint[arxiv]{2401.03890}~[cs.CV]
\urldef\tempurl%
\url{https://arxiv.org/abs/2401.03890}
\showURL{%
\tempurl}


\bibitem[Cheng et~al\mbox{.}(2025b)]%
        {cheng2025outdoor}
\bibfield{author}{\bibinfo{person}{Chong Cheng}, \bibinfo{person}{Sicheng Yu},
  \bibinfo{person}{Zijian Wang}, \bibinfo{person}{Yifan Zhou}, {and}
  \bibinfo{person}{Hao Wang}.} \bibinfo{year}{2025}\natexlab{b}.
\newblock \showarticletitle{Outdoor monocular slam with global scale-consistent
  3d gaussian pointmaps}. In \bibinfo{booktitle}{\emph{ICCV}}.
\newblock


\bibitem[Cheng et~al\mbox{.}(2025a)]%
        {TLS-SLAM}
\bibfield{author}{\bibinfo{person}{Sicong Cheng}, \bibinfo{person}{Songyang
  He}, \bibinfo{person}{Fuqing Duan}, {and} \bibinfo{person}{Ning An}.}
  \bibinfo{year}{2025}\natexlab{a}.
\newblock \showarticletitle{TLS-SLAM: Gaussian Splatting SLAM Tailored for
  Large-Scale Scenes}.
\newblock \bibinfo{journal}{\emph{IEEE Robotics and Automation Letters}}
  (\bibinfo{year}{2025}).
\newblock


\bibitem[Deng et~al\mbox{.}(2025)]%
        {deng2025gigaslam}
\bibfield{author}{\bibinfo{person}{Kai Deng}, \bibinfo{person}{Yigong Zhang},
  \bibinfo{person}{Jian Yang}, {and} \bibinfo{person}{Jin Xie}.}
  \bibinfo{year}{2025}\natexlab{}.
\newblock \showarticletitle{Gigaslam: Large-scale monocular slam with
  hierarchical gaussian splats}. In \bibinfo{booktitle}{\emph{Proceedings of
  the SIGGRAPH Asia 2025 Conference Papers}}. \bibinfo{pages}{1--10}.
\newblock


\bibitem[Deng et~al\mbox{.}(2026)]%
        {deng2026vpgsslamvoxelbasedprogressive3d}
\bibfield{author}{\bibinfo{person}{Tianchen Deng}, \bibinfo{person}{Wenhua Wu},
  \bibinfo{person}{Junjie He}, \bibinfo{person}{Yue Pan},
  \bibinfo{person}{Shenghai Yuan}, \bibinfo{person}{Danwei Wang}, {and}
  \bibinfo{person}{Hesheng Wang}.} \bibinfo{year}{2026}\natexlab{}.
\newblock \bibinfo{title}{VPGS-SLAM: Voxel-based Progressive 3D Gaussian SLAM
  in Large-Scale Scenes}.
\newblock
\newblock
\showeprint[arxiv]{2505.18992}~[cs.CV]


\bibitem[Fan et~al\mbox{.}(2024)]%
        {fan2024instantsplat}
\bibfield{author}{\bibinfo{person}{Zhiwen Fan}, \bibinfo{person}{Wenyan Cong},
  \bibinfo{person}{Kairun Wen}, \bibinfo{person}{Kevin Wang},
  \bibinfo{person}{Jian Zhang}, \bibinfo{person}{Xinghao Ding},
  \bibinfo{person}{Danfei Xu}, \bibinfo{person}{Boris Ivanovic},
  \bibinfo{person}{Marco Pavone}, \bibinfo{person}{Georgios Pavlakos},
  \bibinfo{person}{Zhangyang Wang}, {and} \bibinfo{person}{Yue Wang}.}
  \bibinfo{year}{2024}\natexlab{}.
\newblock \bibinfo{title}{InstantSplat: Unbounded Sparse-view Pose-free
  Gaussian Splatting in 40 Seconds}.
\newblock
\newblock
\showeprint[arxiv]{2403.20309}~[cs.CV]


\bibitem[Fei et~al\mbox{.}(2024)]%
        {fei20243d}
\bibfield{author}{\bibinfo{person}{Ben Fei}, \bibinfo{person}{Jingyi Xu},
  \bibinfo{person}{Rui Zhang}, \bibinfo{person}{Qingyuan Zhou},
  \bibinfo{person}{Weidong Yang}, {and} \bibinfo{person}{Ying He}.}
  \bibinfo{year}{2024}\natexlab{}.
\newblock \showarticletitle{3d gaussian splatting as new era: A survey}.
\newblock \bibinfo{journal}{\emph{IEEE Transactions on Visualization and
  Computer Graphics}} (\bibinfo{year}{2024}).
\newblock


\bibitem[Feng et~al\mbox{.}(2024)]%
        {feng2024CaRtGS}
\bibfield{author}{\bibinfo{person}{Dapeng Feng}, \bibinfo{person}{Zhiqiang
  Chen}, \bibinfo{person}{Yizhen Yin}, \bibinfo{person}{Shipeng Zhong},
  \bibinfo{person}{Yuhua Qi}, {and} \bibinfo{person}{Hongbo Chen}.}
  \bibinfo{year}{2024}\natexlab{}.
\newblock \bibinfo{title}{CaRtGS: Computational Alignment for Real-Time
  Gaussian Splatting SLAM}.
\newblock
\newblock
\showeprint[arxiv]{2410.00486}~[cs.CV]


\bibitem[Fink et~al\mbox{.}(2025)]%
        {fink2025refinement}
\bibfield{author}{\bibinfo{person}{Laura Fink}, \bibinfo{person}{Linus Franke},
  \bibinfo{person}{Bernhard Egger}, \bibinfo{person}{Joachim Keinert}, {and}
  \bibinfo{person}{Marc Stamminger}.} \bibinfo{year}{2025}\natexlab{}.
\newblock \showarticletitle{{Refinement of Monocular Depth Maps via Multi-View
  Differentiable Rendering}}. In \bibinfo{booktitle}{\emph{Vision, Modeling,
  and Visualization}}. \bibinfo{publisher}{The Eurographics Association}.
\newblock


\bibitem[{Fixstars}(2024)]%
        {cudaba}
\bibfield{author}{\bibinfo{person}{{Fixstars}}.}
  \bibinfo{year}{2024}\natexlab{}.
\newblock \bibinfo{title}{{CUDA Bundle Adjustment}}.
\newblock
\newblock
\urldef\tempurl%
\url{https://github.com/fixstars/cuda-bundle-adjustment}
\showURL{%
\tempurl}


\bibitem[Fu et~al\mbox{.}(2024)]%
        {Fu_2024_CVPR}
\bibfield{author}{\bibinfo{person}{Yang Fu}, \bibinfo{person}{Sifei Liu},
  \bibinfo{person}{Amey Kulkarni}, \bibinfo{person}{Jan Kautz},
  \bibinfo{person}{Alexei~A. Efros}, {and} \bibinfo{person}{Xiaolong Wang}.}
  \bibinfo{year}{2024}\natexlab{}.
\newblock \showarticletitle{COLMAP-Free 3D Gaussian Splatting}. In
  \bibinfo{booktitle}{\emph{CVPR}}.
\newblock


\bibitem[Galvez-López and Tardos(2012)]%
        {Lopez2012_BoBW}
\bibfield{author}{\bibinfo{person}{Dorian Galvez-López} {and}
  \bibinfo{person}{Juan~D. Tardos}.} \bibinfo{year}{2012}\natexlab{}.
\newblock \showarticletitle{Bags of Binary Words for Fast Place Recognition in
  Image Sequences}.
\newblock \bibinfo{journal}{\emph{IEEE Transactions on Robotics}}
  (\bibinfo{year}{2012}).
\newblock


\bibitem[Guo et~al\mbox{.}(2025)]%
        {guo2025ontheflylargescale3dreconstruction}
\bibfield{author}{\bibinfo{person}{Yijia Guo}, \bibinfo{person}{Tong Hu},
  \bibinfo{person}{Zhiwei Li}, \bibinfo{person}{Liwen Hu},
  \bibinfo{person}{Keming Qian}, \bibinfo{person}{Xitong Lin},
  \bibinfo{person}{Shengbo Chen}, \bibinfo{person}{Tiejun Huang}, {and}
  \bibinfo{person}{Lei Ma}.} \bibinfo{year}{2025}\natexlab{}.
\newblock \bibinfo{title}{On-the-fly Large-scale 3D Reconstruction from
  Multi-Camera Rigs}.
\newblock
\newblock
\showeprint[arxiv]{2512.08498}~[cs.CV]
\urldef\tempurl%
\url{https://arxiv.org/abs/2512.08498}
\showURL{%
\tempurl}


\bibitem[Hedman et~al\mbox{.}(2018)]%
        {hedman2018deep}
\bibfield{author}{\bibinfo{person}{Peter Hedman}, \bibinfo{person}{Julien
  Philip}, \bibinfo{person}{True Price}, \bibinfo{person}{Jan-Michael Frahm},
  \bibinfo{person}{George Drettakis}, {and} \bibinfo{person}{Gabriel Brostow}.}
  \bibinfo{year}{2018}\natexlab{}.
\newblock \showarticletitle{Deep blending for free-viewpoint image-based
  rendering}.
\newblock \bibinfo{journal}{\emph{ACM Transactions on Graphics}}
  \bibinfo{volume}{37}, \bibinfo{number}{6} (\bibinfo{year}{2018}),
  \bibinfo{pages}{1--15}.
\newblock


\bibitem[Homeyer et~al\mbox{.}(2024)]%
        {homeyer2024droid}
\bibfield{author}{\bibinfo{person}{Christian Homeyer}, \bibinfo{person}{Leon
  Begiristain}, {and} \bibinfo{person}{Christoph Schn{\"o}rr}.}
  \bibinfo{year}{2024}\natexlab{}.
\newblock \showarticletitle{DROID-Splat: Combining end-to-end SLAM with 3D
  Gaussian Splatting}.
\newblock \bibinfo{journal}{\emph{arXiv e-prints}} (\bibinfo{year}{2024}),
  \bibinfo{pages}{arXiv--2411}.
\newblock


\bibitem[Hu et~al\mbox{.}(2025)]%
        {MGSO}
\bibfield{author}{\bibinfo{person}{Yan~Song Hu}, \bibinfo{person}{Nicolas
  Abboud}, \bibinfo{person}{Muhammad~Qasim Ali},
  \bibinfo{person}{Adam~Srebrnjak Yang}, \bibinfo{person}{Imad Elhajj},
  \bibinfo{person}{Daniel Asmar}, \bibinfo{person}{Yuhao Chen}, {and}
  \bibinfo{person}{John~S. Zelek}.} \bibinfo{year}{2025}\natexlab{}.
\newblock \showarticletitle{MGSO: Monocular Real-Time Photometric SLAM with
  Efficient 3D Gaussian Splatting}. In \bibinfo{booktitle}{\emph{IEEE
  International Conference on Robotics and Automation (ICRA)}}.
\newblock


\bibitem[Huang et~al\mbox{.}(2024)]%
        {hhuang2024photoslam}
\bibfield{author}{\bibinfo{person}{Huajian Huang}, \bibinfo{person}{Longwei
  Li}, \bibinfo{person}{Cheng Hui}, {and} \bibinfo{person}{Sai-Kit Yeung}.}
  \bibinfo{year}{2024}\natexlab{}.
\newblock \showarticletitle{Photo-SLAM: Real-time Simultaneous Localization and
  Photorealistic Mapping for Monocular, Stereo, and RGB-D Cameras}. In
  \bibinfo{booktitle}{\emph{CVPR}}.
\newblock


\bibitem[Hughes et~al\mbox{.}(2022)]%
        {hughes2022hydra}
\bibfield{author}{\bibinfo{person}{N. Hughes}, \bibinfo{person}{Y. Chang},
  {and} \bibinfo{person}{L. Carlone}.} \bibinfo{year}{2022}\natexlab{}.
\newblock \showarticletitle{Hydra: A Real-time Spatial Perception System for
  {3D} Scene Graph Construction and Optimization}.
\newblock  (\bibinfo{year}{2022}).
\newblock


\bibitem[Jiang et~al\mbox{.}(2024)]%
        {COGS2024}
\bibfield{author}{\bibinfo{person}{Kaiwen Jiang}, \bibinfo{person}{Yang Fu},
  \bibinfo{person}{Mukund Varma~T}, \bibinfo{person}{Yash Belhe},
  \bibinfo{person}{Xiaolong Wang}, \bibinfo{person}{Hao Su}, {and}
  \bibinfo{person}{Ravi Ramamoorthi}.} \bibinfo{year}{2024}\natexlab{}.
\newblock \showarticletitle{A Construct-Optimize Approach to Sparse View
  Synthesis without Camera Pose}.
\newblock \bibinfo{journal}{\emph{ACM SIGGRAPH Conference Proceedings}}
  (\bibinfo{year}{2024}).
\newblock


\bibitem[Jiang et~al\mbox{.}(2025)]%
        {jiang2025anysplat}
\bibfield{author}{\bibinfo{person}{Lihan Jiang}, \bibinfo{person}{Yucheng Mao},
  \bibinfo{person}{Linning Xu}, \bibinfo{person}{Tao Lu},
  \bibinfo{person}{Kerui Ren}, \bibinfo{person}{Yichen Jin},
  \bibinfo{person}{Xudong Xu}, \bibinfo{person}{Mulin Yu},
  \bibinfo{person}{Jiangmiao Pang}, \bibinfo{person}{Feng Zhao},
  {et~al\mbox{.}}} \bibinfo{year}{2025}\natexlab{}.
\newblock \showarticletitle{Anysplat: Feed-forward 3d gaussian splatting from
  unconstrained views}.
\newblock \bibinfo{journal}{\emph{ACM Transactions on Graphics}}
  (\bibinfo{year}{2025}).
\newblock


\bibitem[Kerbl et~al\mbox{.}(2023)]%
        {kerbl3Dgaussians}
\bibfield{author}{\bibinfo{person}{Bernhard Kerbl}, \bibinfo{person}{Georgios
  Kopanas}, \bibinfo{person}{Thomas Leimk{\"u}hler}, {and}
  \bibinfo{person}{George Drettakis}.} \bibinfo{year}{2023}\natexlab{}.
\newblock \showarticletitle{3D Gaussian Splatting for Real-Time Radiance Field
  Rendering}.
\newblock \bibinfo{journal}{\emph{ACM Transactions on Graphics}}
  (\bibinfo{year}{2023}).
\newblock


\bibitem[Kerbl et~al\mbox{.}(2024)]%
        {hierarchicalgaussians24}
\bibfield{author}{\bibinfo{person}{Bernhard Kerbl}, \bibinfo{person}{Andreas
  Meuleman}, \bibinfo{person}{Georgios Kopanas}, \bibinfo{person}{Michael
  Wimmer}, \bibinfo{person}{Alexandre Lanvin}, {and} \bibinfo{person}{George
  Drettakis}.} \bibinfo{year}{2024}\natexlab{}.
\newblock \showarticletitle{A Hierarchical 3D Gaussian Representation for
  Real-Time Rendering of Very Large Datasets}.
\newblock \bibinfo{journal}{\emph{ACM Transactions on Graphics}}
  (\bibinfo{year}{2024}).
\newblock


\bibitem[Knapitsch et~al\mbox{.}(2017)]%
        {tnt}
\bibfield{author}{\bibinfo{person}{Arno Knapitsch}, \bibinfo{person}{Jaesik
  Park}, \bibinfo{person}{Qian-Yi Zhou}, {and} \bibinfo{person}{Vladlen
  Koltun}.} \bibinfo{year}{2017}\natexlab{}.
\newblock \showarticletitle{Tanks and temples: benchmarking large-scale scene
  reconstruction}.
\newblock \bibinfo{journal}{\emph{ACM Transactions on Graphics}}
  (\bibinfo{year}{2017}).
\newblock


\bibitem[Konolige and Agrawal(2008)]%
        {konolige2008frameslam}
\bibfield{author}{\bibinfo{person}{Kurt Konolige} {and}
  \bibinfo{person}{Motilal Agrawal}.} \bibinfo{year}{2008}\natexlab{}.
\newblock \showarticletitle{FrameSLAM: From bundle adjustment to real-time
  visual mapping}.
\newblock \bibinfo{journal}{\emph{IEEE Transactions on Robotics}}
  (\bibinfo{year}{2008}).
\newblock


\bibitem[K{\"u}mmerle et~al\mbox{.}(2011)]%
        {kuemmerle2011g2o}
\bibfield{author}{\bibinfo{person}{Rainer K{\"u}mmerle},
  \bibinfo{person}{Giorgio Grisetti}, \bibinfo{person}{Hauke Strasdat},
  \bibinfo{person}{Kurt Konolige}, {and} \bibinfo{person}{Wolfram Burgard}.}
  \bibinfo{year}{2011}\natexlab{}.
\newblock \showarticletitle{g2o: A General Framework for Graph Optimization}.
  In \bibinfo{booktitle}{\emph{IEEE International Conference on Robotics and
  Automation (ICRA)}}.
\newblock


\bibitem[Lin et~al\mbox{.}(2025b)]%
        {lin2025longsplat}
\bibfield{author}{\bibinfo{person}{Chin-Yang Lin}, \bibinfo{person}{Cheng Sun},
  \bibinfo{person}{Fu-En Yang}, \bibinfo{person}{Min-Hung Chen},
  \bibinfo{person}{Yen-Yu Lin}, {and} \bibinfo{person}{Yu-Lun Liu}.}
  \bibinfo{year}{2025}\natexlab{b}.
\newblock \showarticletitle{Longsplat: Robust unposed 3d gaussian splatting for
  casual long videos}. In \bibinfo{booktitle}{\emph{ICCV}}.
\newblock


\bibitem[Lin et~al\mbox{.}(2025a)]%
        {depthanything3}
\bibfield{author}{\bibinfo{person}{Haotong Lin}, \bibinfo{person}{Sili Chen},
  \bibinfo{person}{Jun~Hao Liew}, \bibinfo{person}{Donny~Y. Chen},
  \bibinfo{person}{Zhenyu Li}, \bibinfo{person}{Guang Shi},
  \bibinfo{person}{Jiashi Feng}, {and} \bibinfo{person}{Bingyi Kang}.}
  \bibinfo{year}{2025}\natexlab{a}.
\newblock \showarticletitle{Depth Anything 3: Recovering the visual space from
  any views}.
\newblock \bibinfo{journal}{\emph{arXiv preprint arXiv:2511.10647}}
  (\bibinfo{year}{2025}).
\newblock


\bibitem[Lindenberger et~al\mbox{.}(2023)]%
        {lindenberger2023lightglue}
\bibfield{author}{\bibinfo{person}{Philipp Lindenberger},
  \bibinfo{person}{Paul-Edouard Sarlin}, {and} \bibinfo{person}{Marc
  Pollefeys}.} \bibinfo{year}{2023}\natexlab{}.
\newblock \showarticletitle{{LightGlue: Local Feature Matching at Light
  Speed}}. In \bibinfo{booktitle}{\emph{ICCV}}.
\newblock


\bibitem[Liso et~al\mbox{.}(2024)]%
        {Liso_2024_CVPR}
\bibfield{author}{\bibinfo{person}{Lorenzo Liso}, \bibinfo{person}{Erik
  Sandstr\"om}, \bibinfo{person}{Vladimir Yugay}, \bibinfo{person}{Luc
  Van~Gool}, {and} \bibinfo{person}{Martin~R. Oswald}.}
  \bibinfo{year}{2024}\natexlab{}.
\newblock \showarticletitle{Loopy-SLAM: Dense Neural SLAM with Loop Closures}.
  In \bibinfo{booktitle}{\emph{Proceedings of the IEEE/CVF Conference on
  Computer Vision and Pattern Recognition (CVPR)}}.
  \bibinfo{pages}{20363--20373}.
\newblock


\bibitem[Macario~Barros et~al\mbox{.}(2022)]%
        {macario2022comprehensive}
\bibfield{author}{\bibinfo{person}{Andr{\'e}a Macario~Barros},
  \bibinfo{person}{Maugan Michel}, \bibinfo{person}{Yoann Moline},
  \bibinfo{person}{Gwenol{\'e} Corre}, {and} \bibinfo{person}{Fr{\'e}d{\'e}rick
  Carrel}.} \bibinfo{year}{2022}\natexlab{}.
\newblock \showarticletitle{A comprehensive survey of visual slam algorithms}.
\newblock \bibinfo{journal}{\emph{Robotics}} (\bibinfo{year}{2022}).
\newblock


\bibitem[Maggio et~al\mbox{.}(2025)]%
        {maggio2025vggt-slam}
\bibfield{author}{\bibinfo{person}{Dominic Maggio}, \bibinfo{person}{Hyungtae
  Lim}, {and} \bibinfo{person}{Luca Carlone}.} \bibinfo{year}{2025}\natexlab{}.
\newblock \showarticletitle{VGGT-SLAM: Dense RGB SLAM Optimized on the SL (4)
  Manifold}.
\newblock \bibinfo{journal}{\emph{Advances in Neural Information Processing
  Systems}} (\bibinfo{year}{2025}).
\newblock


\bibitem[Mallick et~al\mbox{.}(2024)]%
        {taming3dgs}
\bibfield{author}{\bibinfo{person}{Saswat~Subhajyoti Mallick},
  \bibinfo{person}{Rahul Goel}, \bibinfo{person}{Bernhard Kerbl},
  \bibinfo{person}{Markus Steinberger}, \bibinfo{person}{Francisco~Vicente
  Carrasco}, {and} \bibinfo{person}{Fernando De~La~Torre}.}
  \bibinfo{year}{2024}\natexlab{}.
\newblock \showarticletitle{Taming 3DGS: High-Quality Radiance Fields with
  Limited Resources}. In \bibinfo{booktitle}{\emph{SIGGRAPH Asia 2024
  Conference Papers}}.
\newblock


\bibitem[Matsuki et~al\mbox{.}(2024)]%
        {gsslam2024}
\bibfield{author}{\bibinfo{person}{Hidenobu Matsuki}, \bibinfo{person}{Riku
  Murai}, \bibinfo{person}{Paul H.~J. Kelly}, {and} \bibinfo{person}{Andrew~J.
  Davison}.} \bibinfo{year}{2024}\natexlab{}.
\newblock \showarticletitle{{G}aussian {S}platting {SLAM}}.
\newblock  (\bibinfo{year}{2024}).
\newblock


\bibitem[Meuleman et~al\mbox{.}(2023)]%
        {meuleman2023localrf}
\bibfield{author}{\bibinfo{person}{Andreas Meuleman}, \bibinfo{person}{Yu-Lun
  Liu}, \bibinfo{person}{Chen Gao}, \bibinfo{person}{Jia-Bin Huang},
  \bibinfo{person}{Changil Kim}, \bibinfo{person}{Min~H. Kim}, {and}
  \bibinfo{person}{Johannes Kopf}.} \bibinfo{year}{2023}\natexlab{}.
\newblock \showarticletitle{Progressively Optimized Local Radiance Fields for
  Robust View Synthesis}. In \bibinfo{booktitle}{\emph{CVPR}}.
\newblock


\bibitem[Meuleman et~al\mbox{.}(2025)]%
        {meuleman2025onthefly}
\bibfield{author}{\bibinfo{person}{Andreas Meuleman}, \bibinfo{person}{Ishaan
  Shah}, \bibinfo{person}{Alexandre Lanvin}, \bibinfo{person}{Bernhard Kerbl},
  {and} \bibinfo{person}{George Drettakis}.} \bibinfo{year}{2025}\natexlab{}.
\newblock \showarticletitle{On-the-fly Reconstruction for Large-Scale Novel
  View Synthesis from Unposed Images}.
\newblock \bibinfo{journal}{\emph{ACM Transactions on Graphics}}
  (\bibinfo{year}{2025}).
\newblock


\bibitem[Mildenhall et~al\mbox{.}(2020)]%
        {mildenhall2020nerf}
\bibfield{author}{\bibinfo{person}{Ben Mildenhall}, \bibinfo{person}{Pratul~P.
  Srinivasan}, \bibinfo{person}{Matthew Tancik}, \bibinfo{person}{Jonathan~T.
  Barron}, \bibinfo{person}{Ravi Ramamoorthi}, {and} \bibinfo{person}{Ren Ng}.}
  \bibinfo{year}{2020}\natexlab{}.
\newblock \showarticletitle{NeRF: Representing Scenes as Neural Radiance Fields
  for View Synthesis}. In \bibinfo{booktitle}{\emph{European Conference on
  Computer Vision (ECCV)}}.
\newblock


\bibitem[Mur-Artal et~al\mbox{.}(2015)]%
        {mur2015ORBSLAM}
\bibfield{author}{\bibinfo{person}{Ra\'ul Mur-Artal}, \bibinfo{person}{J.~M.~M.
  Montiel}, {and} \bibinfo{person}{Juan~D. Tard\'os}.}
  \bibinfo{year}{2015}\natexlab{}.
\newblock \showarticletitle{{ORB-SLAM}: a Versatile and Accurate Monocular
  {SLAM} System}.
\newblock \bibinfo{journal}{\emph{IEEE Transactions on Robotics}}
  (\bibinfo{year}{2015}).
\newblock


\bibitem[Murai et~al\mbox{.}(2025)]%
        {murai2024_mast3rslam}
\bibfield{author}{\bibinfo{person}{Riku Murai}, \bibinfo{person}{Eric
  Dexheimer}, {and} \bibinfo{person}{Andrew~J. Davison}.}
  \bibinfo{year}{2025}\natexlab{}.
\newblock \showarticletitle{{MASt3R-SLAM}: Real-Time Dense {SLAM} with {3D}
  Reconstruction Priors}. In \bibinfo{booktitle}{\emph{CVPR}}.
\newblock


\bibitem[{\"O}zye{\c{s}}il et~al\mbox{.}(2017)]%
        {ozyecsil2017survey}
\bibfield{author}{\bibinfo{person}{Onur {\"O}zye{\c{s}}il},
  \bibinfo{person}{Vladislav Voroninski}, \bibinfo{person}{Ronen Basri}, {and}
  \bibinfo{person}{Amit Singer}.} \bibinfo{year}{2017}\natexlab{}.
\newblock \showarticletitle{A survey of structure from motion*.}
\newblock \bibinfo{journal}{\emph{Acta Numerica}} (\bibinfo{year}{2017}).
\newblock


\bibitem[Pan et~al\mbox{.}(2024)]%
        {pan2024glomap}
\bibfield{author}{\bibinfo{person}{Linfei Pan}, \bibinfo{person}{Dániel
  Baráth}, \bibinfo{person}{Marc Pollefeys}, {and}
  \bibinfo{person}{Johannes~Lutz Sch\"{o}nberger}.}
  \bibinfo{year}{2024}\natexlab{}.
\newblock \showarticletitle{Global Structure-from-Motion Revisited}. In
  \bibinfo{booktitle}{\emph{European Conference on Computer Vision (ECCV)}}.
\newblock


\bibitem[Pan et~al\mbox{.}(2025)]%
        {pan2025egg}
\bibfield{author}{\bibinfo{person}{Xiaokun Pan}, \bibinfo{person}{Zhenzhe Li},
  \bibinfo{person}{Zhichao Ye}, \bibinfo{person}{Hongjia Zhai}, {and}
  \bibinfo{person}{Guofeng Zhang}.} \bibinfo{year}{2025}\natexlab{}.
\newblock \showarticletitle{EGG-Fusion: Efficient 3D Reconstruction with
  Geometry-aware Gaussian Surfel on the Fly}. In
  \bibinfo{booktitle}{\emph{Proceedings of the SIGGRAPH Asia 2025 Conference
  Papers}}. \bibinfo{pages}{1--12}.
\newblock


\bibitem[Papantonakis et~al\mbox{.}(2024)]%
        {Reducing3Dgs}
\bibfield{author}{\bibinfo{person}{Panagiotis Papantonakis},
  \bibinfo{person}{Georgios Kopanas}, \bibinfo{person}{Bernhard Kerbl},
  \bibinfo{person}{Alexandre Lanvin}, {and} \bibinfo{person}{George
  Drettakis}.} \bibinfo{year}{2024}\natexlab{}.
\newblock \showarticletitle{{Reducing the Memory Footprint of 3D Gaussian
  Splatting}}. In \bibinfo{booktitle}{\emph{{Proceedings of the ACM on Computer
  Graphics and Interactive Techniques}}}.
\newblock


\bibitem[Perazzi et~al\mbox{.}(2016)]%
        {perazzi2016benchmark}
\bibfield{author}{\bibinfo{person}{Federico Perazzi}, \bibinfo{person}{Jordi
  Pont-Tuset}, \bibinfo{person}{Brian McWilliams}, \bibinfo{person}{Luc
  Van~Gool}, \bibinfo{person}{Markus Gross}, {and} \bibinfo{person}{Alexander
  Sorkine-Hornung}.} \bibinfo{year}{2016}\natexlab{}.
\newblock \showarticletitle{A benchmark dataset and evaluation methodology for
  video object segmentation}. In \bibinfo{booktitle}{\emph{CVPR}}.
\newblock


\bibitem[Persson and Nordberg(2018)]%
        {Persson_2018_ECCVLambdaTwist}
\bibfield{author}{\bibinfo{person}{Mikael Persson} {and} \bibinfo{person}{Klas
  Nordberg}.} \bibinfo{year}{2018}\natexlab{}.
\newblock \showarticletitle{Lambda Twist: An Accurate Fast Robust Perspective
  Three Point (P3P) Solver}. In \bibinfo{booktitle}{\emph{European Conference
  on Computer Vision (ECCV)}}.
\newblock


\bibitem[{Potje} et~al\mbox{.}(2024)]%
        {potje2024xfeat}
\bibfield{author}{\bibinfo{person}{Guilherme {Potje}}, \bibinfo{person}{Felipe
  {Cadar}}, \bibinfo{person}{Andre {Araujo}}, \bibinfo{person}{Renato
  {Martins}}, {and} \bibinfo{person}{Erickson~R. {Nascimento}}.}
  \bibinfo{year}{2024}\natexlab{}.
\newblock \showarticletitle{XFeat: Accelerated Features for Lightweight Image
  Matching}. In \bibinfo{booktitle}{\emph{CVPR}}.
\newblock


\bibitem[Ren et~al\mbox{.}(2022)]%
        {2021megba}
\bibfield{author}{\bibinfo{person}{Jie Ren}, \bibinfo{person}{Wenteng Liang},
  \bibinfo{person}{Ran Yan}, \bibinfo{person}{Luo Mai}, \bibinfo{person}{Shiwen
  Liu}, {and} \bibinfo{person}{Xiao Liu}.} \bibinfo{year}{2022}\natexlab{}.
\newblock \showarticletitle{MegBA: A GPU-Based Distributed Library for
  Large-Scale Bundle Adjustment}. In \bibinfo{booktitle}{\emph{European
  Conference on Computer Vision (ECCV)}}.
\newblock


\bibitem[Ren et~al\mbox{.}(2024)]%
        {octreeGS}
\bibfield{author}{\bibinfo{person}{Kerui Ren}, \bibinfo{person}{Lihan Jiang},
  \bibinfo{person}{Tao Lu}, \bibinfo{person}{Mulin Yu},
  \bibinfo{person}{Linning Xu}, \bibinfo{person}{Zhangkai Ni}, {and}
  \bibinfo{person}{Bo Dai}.} \bibinfo{year}{2024}\natexlab{}.
\newblock \bibinfo{title}{Octree-GS: Towards Consistent Real-time Rendering
  with LOD-Structured 3D Gaussians}.
\newblock
\newblock
\showeprint[arxiv]{2403.17898}~[cs.CV]
\urldef\tempurl%
\url{https://arxiv.org/abs/2403.17898}
\showURL{%
\tempurl}


\bibitem[Sch\"{o}nberger and Frahm(2016)]%
        {schoenberger2016sfm}
\bibfield{author}{\bibinfo{person}{Johannes~Lutz Sch\"{o}nberger} {and}
  \bibinfo{person}{Jan-Michael Frahm}.} \bibinfo{year}{2016}\natexlab{}.
\newblock \showarticletitle{Structure-from-Motion Revisited}. In
  \bibinfo{booktitle}{\emph{CVPR}}.
\newblock


\bibitem[Strasdat et~al\mbox{.}(2010)]%
        {Strasdat2010ScaleDL}
\bibfield{author}{\bibinfo{person}{Hauke~Malte Strasdat},
  \bibinfo{person}{Jos{\'e} M.~M. Montiel}, {and} \bibinfo{person}{Andrew~J.
  Davison}.} \bibinfo{year}{2010}\natexlab{}.
\newblock \showarticletitle{Scale Drift-Aware Large Scale Monocular SLAM}. In
  \bibinfo{booktitle}{\emph{Robotics: Science and Systems}}.
\newblock


\bibitem[Taketomi et~al\mbox{.}(2017)]%
        {taketomi2017visual}
\bibfield{author}{\bibinfo{person}{Takafumi Taketomi}, \bibinfo{person}{Hideaki
  Uchiyama}, {and} \bibinfo{person}{Sei Ikeda}.}
  \bibinfo{year}{2017}\natexlab{}.
\newblock \showarticletitle{Visual SLAM algorithms: A survey from 2010 to
  2016}.
\newblock \bibinfo{journal}{\emph{IPSJ transactions on computer vision and
  applications}} (\bibinfo{year}{2017}).
\newblock


\bibitem[Teed and Deng(2021)]%
        {teed2021droid}
\bibfield{author}{\bibinfo{person}{Zachary Teed} {and} \bibinfo{person}{Jia
  Deng}.} \bibinfo{year}{2021}\natexlab{}.
\newblock \showarticletitle{Droid-slam: Deep visual slam for monocular, stereo,
  and rgb-d cameras}. In \bibinfo{booktitle}{\emph{NIPS}}.
\newblock


\bibitem[Tewari et~al\mbox{.}(2020)]%
        {tewari2020state}
\bibfield{author}{\bibinfo{person}{Ayush Tewari}, \bibinfo{person}{Ohad Fried},
  \bibinfo{person}{Justus Thies}, \bibinfo{person}{Vincent Sitzmann},
  \bibinfo{person}{Stephen Lombardi}, \bibinfo{person}{Kalyan Sunkavalli},
  \bibinfo{person}{Ricardo Martin-Brualla}, \bibinfo{person}{Tomas Simon},
  \bibinfo{person}{Jason Saragih}, \bibinfo{person}{Matthias Nie{\ss}ner},
  {et~al\mbox{.}}} \bibinfo{year}{2020}\natexlab{}.
\newblock \showarticletitle{State of the art on neural rendering}. In
  \bibinfo{booktitle}{\emph{Computer graphics forum}}.
\newblock


\bibitem[Tsintotas et~al\mbox{.}(2022)]%
        {tsintotas2022revisiting}
\bibfield{author}{\bibinfo{person}{Konstantinos~A Tsintotas},
  \bibinfo{person}{Loukas Bampis}, {and} \bibinfo{person}{Antonios
  Gasteratos}.} \bibinfo{year}{2022}\natexlab{}.
\newblock \showarticletitle{The revisiting problem in simultaneous localization
  and mapping: A survey on visual loop closure detection}.
\newblock \bibinfo{journal}{\emph{IEEE Transactions on Intelligent
  Transportation Systems}} (\bibinfo{year}{2022}).
\newblock


\bibitem[Wang et~al\mbox{.}(2023)]%
        {wang2023pypose}
\bibfield{author}{\bibinfo{person}{Chen Wang}, \bibinfo{person}{Dasong Gao},
  \bibinfo{person}{Kuan Xu}, \bibinfo{person}{Junyi Geng},
  \bibinfo{person}{Yaoyu Hu}, \bibinfo{person}{Yuheng Qiu},
  \bibinfo{person}{Bowen Li}, \bibinfo{person}{Fan Yang},
  \bibinfo{person}{Brady Moon}, \bibinfo{person}{Abhinav Pandey},
  \bibinfo{person}{Aryan}, \bibinfo{person}{Jiahe Xu}, \bibinfo{person}{Tianhao
  Wu}, \bibinfo{person}{Haonan He}, \bibinfo{person}{Daning Huang},
  \bibinfo{person}{Zhongqiang Ren}, \bibinfo{person}{Shibo Zhao},
  \bibinfo{person}{Taimeng Fu}, \bibinfo{person}{Pranay Reddy},
  \bibinfo{person}{Xiao Lin}, \bibinfo{person}{Wenshan Wang},
  \bibinfo{person}{Jingnan Shi}, \bibinfo{person}{Rajat Talak},
  \bibinfo{person}{Kun Cao}, \bibinfo{person}{Yi Du}, \bibinfo{person}{Han
  Wang}, \bibinfo{person}{Huai Yu}, \bibinfo{person}{Shanzhao Wang},
  \bibinfo{person}{Siyu Chen}, \bibinfo{person}{Ananth Kashyap},
  \bibinfo{person}{Rohan Bandaru}, \bibinfo{person}{Karthik Dantu},
  \bibinfo{person}{Jiajun Wu}, \bibinfo{person}{Lihua Xie},
  \bibinfo{person}{Luca Carlone}, \bibinfo{person}{Marco Hutter}, {and}
  \bibinfo{person}{Sebastian Scherer}.} \bibinfo{year}{2023}\natexlab{}.
\newblock \showarticletitle{{PyPose}: A Library for Robot Learning with
  Physics-based Optimization}. In \bibinfo{booktitle}{\emph{CVPR}}.
\newblock


\bibitem[Wang and Agapito(2025)]%
        {wang2025spann3r}
\bibfield{author}{\bibinfo{person}{Hengyi Wang} {and} \bibinfo{person}{Lourdes
  Agapito}.} \bibinfo{year}{2025}\natexlab{}.
\newblock \showarticletitle{3d reconstruction with spatial memory}. In
  \bibinfo{booktitle}{\emph{International Conference on 3D Vision (3DV)}}.
\newblock


\bibitem[Whelan et~al\mbox{.}(2015)]%
        {whelan2015elasticfusion}
\bibfield{author}{\bibinfo{person}{Thomas Whelan}, \bibinfo{person}{Stefan
  Leutenegger}, \bibinfo{person}{Renato~F. Salas-Moreno}, \bibinfo{person}{Ben
  Glocker}, {and} \bibinfo{person}{Andrew~J. Davison}.}
  \bibinfo{year}{2015}\natexlab{}.
\newblock \showarticletitle{ElasticFusion: Dense SLAM Without A Pose Graph}. In
  \bibinfo{booktitle}{\emph{Robotics: Science and Systems}}.
\newblock


\bibitem[Whelan et~al\mbox{.}(2016)]%
        {whelan2016elasticfusion}
\bibfield{author}{\bibinfo{person}{Thomas Whelan}, \bibinfo{person}{Renato~F.
  Salas-Moreno}, \bibinfo{person}{Ben Glocker}, \bibinfo{person}{Andrew~J.
  Davison}, {and} \bibinfo{person}{Stefan Leutenegger}.}
  \bibinfo{year}{2016}\natexlab{}.
\newblock \showarticletitle{ElasticFusion: Real-Time Dense SLAM and Light
  Source Estimation}.
\newblock \bibinfo{journal}{\emph{International Journal of Robotics Research}}
  (\bibinfo{year}{2016}).
\newblock


\bibitem[Wu et~al\mbox{.}(2025)]%
        {wu2025monocular}
\bibfield{author}{\bibinfo{person}{Songyin Wu}, \bibinfo{person}{Zhaoyang Lv},
  \bibinfo{person}{Yufeng Zhu}, \bibinfo{person}{Duncan Frost},
  \bibinfo{person}{Zhengqin Li}, \bibinfo{person}{Ling-Qi Yan},
  \bibinfo{person}{Carl Ren}, \bibinfo{person}{Richard Newcombe}, {and}
  \bibinfo{person}{Zhao Dong}.} \bibinfo{year}{2025}\natexlab{}.
\newblock \showarticletitle{Monocular online reconstruction with enhanced
  detail preservation}. In \bibinfo{booktitle}{\emph{Proceedings of the Special
  Interest Group on Computer Graphics and Interactive Techniques Conference
  Conference Papers}}. \bibinfo{pages}{1--11}.
\newblock


\bibitem[Xin et~al\mbox{.}(2025)]%
        {xin2025large}
\bibfield{author}{\bibinfo{person}{Zhe Xin}, \bibinfo{person}{Chenyang Wu},
  \bibinfo{person}{Penghui Huang}, \bibinfo{person}{Yanyong Zhang},
  \bibinfo{person}{Yinian Mao}, {and} \bibinfo{person}{Guoquan Huang}.}
  \bibinfo{year}{2025}\natexlab{}.
\newblock \showarticletitle{Large-Scale Gaussian Splatting SLAM}.
\newblock \bibinfo{journal}{\emph{IEEE International Conference on Robotics and
  Automation (ICRA)}}.
\newblock


\bibitem[Yang et~al\mbox{.}(2025)]%
        {Yang2025V3DG}
\bibfield{author}{\bibinfo{person}{Xijie Yang}, \bibinfo{person}{Linning Xu},
  \bibinfo{person}{Lihan Jiang}, \bibinfo{person}{Dahua Lin}, {and}
  \bibinfo{person}{Bo Dai}.} \bibinfo{year}{2025}\natexlab{}.
\newblock \showarticletitle{Virtualized 3D Gaussians: Flexible Cluster-based
  Level-of-Detail System for Real-Time Rendering of Composed Scenes}. In
  \bibinfo{booktitle}{\emph{SIGGRAPH Conference Proceedings}}.
\newblock


\bibitem[Zhou et~al\mbox{.}(2019)]%
        {Zhou_2019_CVPR}
\bibfield{author}{\bibinfo{person}{Yi Zhou}, \bibinfo{person}{Connelly Barnes},
  \bibinfo{person}{Lu Jingwan}, \bibinfo{person}{Yang Jimei}, {and}
  \bibinfo{person}{Li Hao}.} \bibinfo{year}{2019}\natexlab{}.
\newblock \showarticletitle{On the Continuity of Rotation Representations in
  Neural Networks}. In \bibinfo{booktitle}{\emph{CVPR}}.
\newblock


\bibitem[Zhu et~al\mbox{.}(2025)]%
        {zhu2024_loopsplat}
\bibfield{author}{\bibinfo{person}{Liyuan Zhu}, \bibinfo{person}{Yue Li},
  \bibinfo{person}{Erik Sandström}, \bibinfo{person}{Shengyu Huang},
  \bibinfo{person}{Konrad Schindler}, {and} \bibinfo{person}{Iro Armeni}.}
  \bibinfo{year}{2025}\natexlab{}.
\newblock \bibinfo{title}{LoopSplat: Loop Closure by Registering 3D Gaussian
  Splats}.
\newblock
\newblock


\end{thebibliography}

\appendix
\SetKwInOut{Input}{Input}
\SetKwInOut{Output}{Output}
\SetKwFunction{TwoLevelMatching}{TwoLevelMatching}
\SetKwFunction{UpdateGraph}{UpdateGraph}
\SetKwFunction{SelectKeyframes}{SelectKeyframes}
\SetKwFunction{LoopDetection}{LoopDetection}
\SetKwFunction{LocalBA}{LocalBA}
\SetKwFunction{PlaceGaussians}{PlaceGaussians}
\SetKwFunction{WeldingBA}{WeldingBA}
\SetKwFunction{BackbonePropagation}{BackbonePropagation}
\SetKwFunction{UpdateHierarchy}{UpdateHierarchy}
\SetKwFunction{RetryShelved}{RetryShelved}
\SetKwFunction{OptimizeGaussians}{OptimizeGaussians}
\SetKwFunction{HierarchyTraversal}{HierarchyTraversal}
\SetKwFunction{SelectFramesOpt}{SelectFramesForOpt}
\SetKwFunction{MixVPR}{MixVPR}
\SetKwFunction{XFeatBF}{XFeat.BruteForce}
\SetKwFunction{LightGlue}{LightGlue}
\SetKwFunction{RANSAC}{RANSAC}
\SetKwFunction{DepthAnything}{DepthAnything3}
\SetKwFunction{SpawnGaussians}{SpawnGaussians}
\SetKwFunction{DiffRender}{DiffRender}
\SetKwFunction{RenderLoss}{RenderLoss}
\SetKwFunction{MultinomialSample}{MultinomialSample}
\SetKwFunction{RandomSample}{RandomSample}
\SetKwFunction{ComputeResJac}{ComputeResidualsJacobiansGPU}
\SetKwFunction{Dijkstra}{Dijkstra}
\SetKwFunction{GeodesicDist}{GeodesicDist}
\SetKwFunction{InterpSixD}{Interp6D}
\SetKwFunction{FastKNN}{FastKNN}
\SetKwFunction{KLMerge}{KLMerge}
\SetKwFunction{ScreenSize}{ScreenSize}
\SetKwFunction{FitScaleOffset}{FitScaleOffset}
\SetKwFunction{BilinearUpsample}{BilinearUpsample}
\SetKwFunction{SampleNearby}{SampleNearbyViews}
\SetKwProg{Proc}{Procedure}{}{}
\SetKwProg{Fn}{Function}{}{}

\newcommand{\calK}{\mathcal{K}}
\newcommand{\calM}{\mathcal{M}}
\newcommand{\calV}{\mathcal{V}}
\newcommand{\calS}{\mathcal{S}}
\newcommand{\calG}{\mathcal{G}}
\newcommand{\calC}{\mathcal{C}}
\newcommand{\calH}{\mathcal{H}}
\newcommand{\calW}{\mathcal{W}}
\newcommand{\calL}{\mathcal{L}}

\section{Additional Method Details}

\begin{figure*}[!h]
\def\svgwidth{\linewidth}
\small
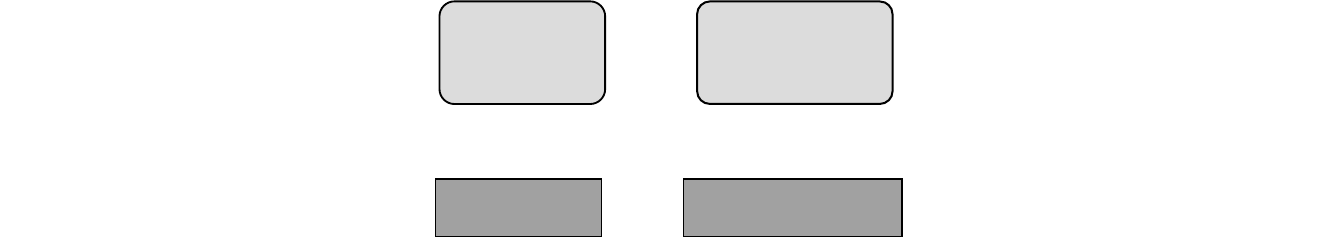
\caption{{Overview of our proposed pipeline. Given an input image $I_q$ sampled from an unordered image collection, we first apply a two-level matching strategy to establish correspondences. Images that cannot be matched are shelved for later processing. For matchable images, camera poses are estimated and refined via local bundle adjustment (BA), with loop closure aided by querying the covisibility graph to identify related keyframes. New Gaussian primitives are then instantiated and integrated into the scene hierarchy, followed by several optimization steps prior to the next input image.}}
\label{fig:pipeline}
\end{figure*}

{
\subsection{Overview}
Algorithm~\ref{alg:main} and Figure~\ref{fig:pipeline} provide an overview of our method.
\begin{algorithm}
{
\caption{Main Pipeline}\label{alg:main}
\Input{Stream of unordered RGB images $\{I_1, I_2, \ldots\}$}
\Output{3DGS reconstruction}
$\calM \gets \emptyset$\tcp*{Registered keyframes}
$G \gets (\emptyset, \emptyset, \emptyset)$\tcp*{Covisibility graph $(\calM,E,W)$}
$\mathcal{G} \gets \emptyset$\tcp*{Active Gaussian primitives}
$\calH \gets \emptyset$\tcp*{Gaussian hierarchy}
$\textit{Shelf} \gets \emptyset$\tcp*{Shelved unmatched frames}
\BlankLine
Bootstrap first 8 well-connected keyframes\tcp*{Sec.~5}
\BlankLine
\ForEach{incoming image $I_q$}{
    $\calV, \calK' \gets$ \TwoLevelMatching{$I_q, \calM$}\tcp*{Sec.~4.1}
    \If{$\calV = \emptyset$}{
        $\textit{Shelf} \gets \textit{Shelf} \cup \{I_q\}$\;
        \textbf{continue}\tcp*{Shelve for later (Sec.~5.3)}
    }
    \UpdateGraph{$G, I_q, \calV$}\tcp*{Sec.~4.2}
    $\textit{loop}, \calC_1, \calC_2 \gets$ \LoopDetection{$I_q, \calK', G$}\tcp*{Sec.~6.1}
    $\calS \gets$ \SelectKeyframes{$\{I_q\}, G, N{=}20$}\tcp*{Sec.~4.3}
    \LocalBA{$\calS$}\tcp*{Sec.~5.1}
    \If{\textit{loop}}{
        \WeldingBA{$\calC_1, \calC_2, G$}\tcp*{Sec.~6.2}
        \BackbonePropagation{$G, \calC_1, \calC_2, \mathcal{G}$}\tcp*{Sec.~6.3}
    }
    \PlaceGaussians{$I_q, \mathcal{G}$}\tcp*{Sec.~5.2}
    $\calM \gets \calM \cup \{I_q\}$\;
    \UpdateHierarchy{$\mathcal{G}, \calH, I_q$}\tcp*{Sec.~7.1}
    \OptimizeGaussians{$\calM, \mathcal{G}$}\tcp*{Sec.~5.2}
    \RetryShelved{$\textit{Shelf}, \calM, G$}\tcp*{Sec.~5.3}
}
}
\end{algorithm}
}

\subsection{Rotation Invariance in Matching}

While robust, learned descriptors, especially those based on CNNs, are typically not rotation invariant and can only handle small rotations. 
To allow our method to handle a wide variety of data, even when frames are captured with strong rotation around the camera axis, we extract MixVPR and XFeat features for four different rotations of the image, with 90° steps. 
For XFeat keypoints, we only extract the keypoints in the upright image and sample the features in all of the four rotated images.
For matching, we first quickly test rotation using MixVPR, before selecting the corresponding rotated XFeat features. 
Using MixVPR to detect orientation before XFeat and LightGlue allows us to maintain matching efficiency. 

\subsection{Details on Local Pose Reconstruction}

\paragraph{Fast Bundle Adjustment} 

We employ a custom version of bundle adjustment (BA) for both local BA and welding window loop closure, which has to be very fast for our incremental reconstruction objective.

Previous fast BA solutions~\cite{meuleman2025onthefly} are 
efficient and accurate only for very small problems ($N=5$), but do not scale:
every landmark is observed by all keyframes in the local window, leading to a quadratic growth of the number of observations with respect to the number of keyframes. 
This occurs since the number of landmarks grows linearly with the number of keyframes. 
As a result such solutions are not sufficiently fast in our context with $N=20$ keyframes.
We address this limitation while maintaining the GPU-compatibility of the solver by allowing each landmark to be observed by a fixed subset of keyframes in the local window, selected based on covisibility. 
This reduces the number of observations significantly, enabling us to scale to larger local windows without sacrificing efficiency.

\paragraph{Shelving}
Shelving is our solution to register disjoint sequences efficiently.
When a new frame cannot be matched to any of the registered set, or if pose initialization fails, it is added to the shelf to process it later.
We continuously compute the best MixVPR match between the shelved frames and the registered set. If the best match is above the cosine similarity of the last registered image, we attempt to add the shelved frame to the registered set.
If frames remain in the shelf after the end of the capture, we process them starting from the best VPR match to the worst.

{We interleave unshelving with new-frame processing, keeping the delay between capture and processing start below the time to handle two keyframes ($<$0.8 s).}

\subsection{Loop Closure Details}
\vspace{.2cm}

\begin{figure}[!h]
\setlength{\tabcolsep}{1pt}
\renewcommand{\arraystretch}{0}
\begin{tabular}{cc}
    \includegraphics[width=0.49\linewidth]{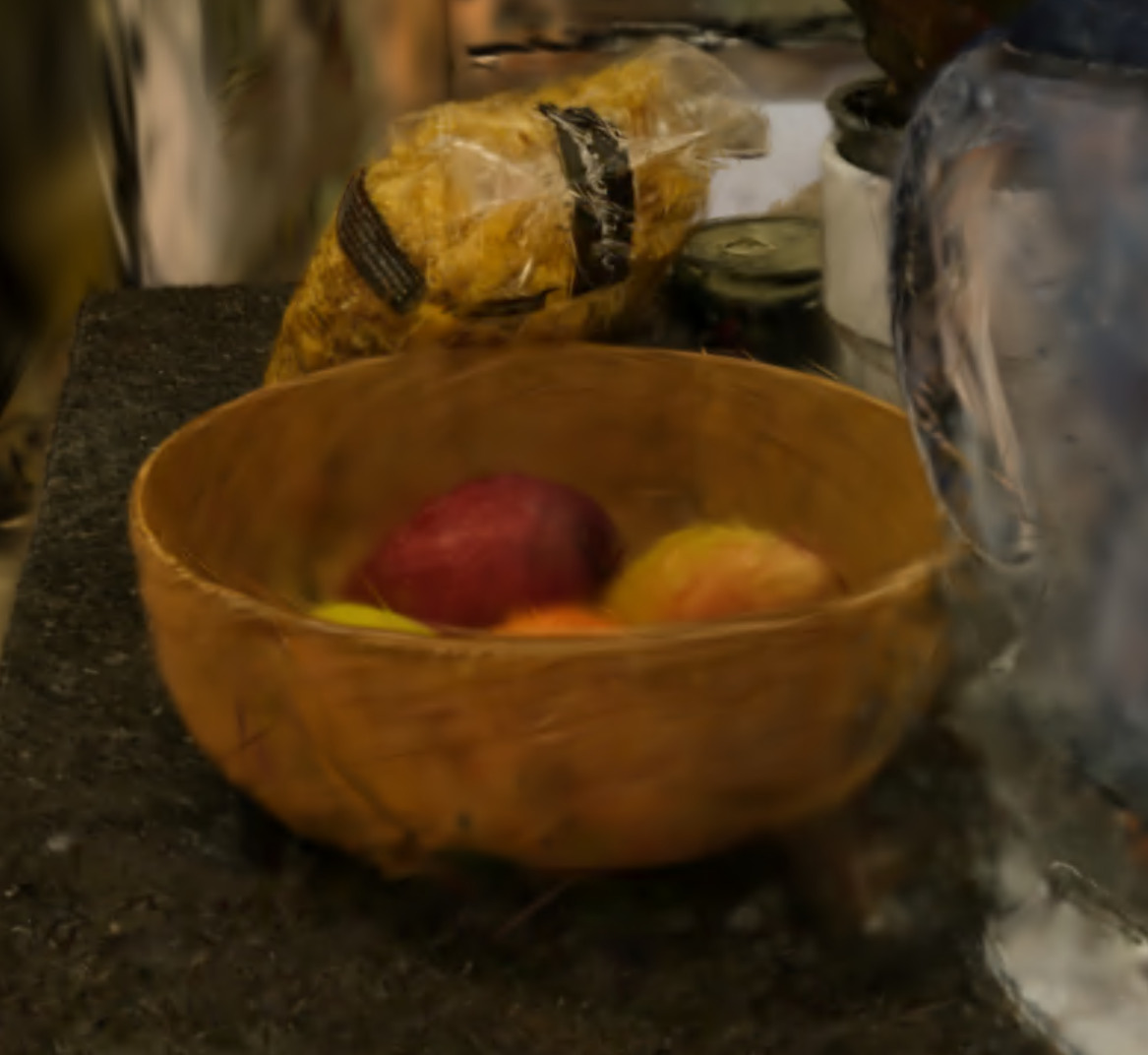} &
    \includegraphics[width=0.49\linewidth]{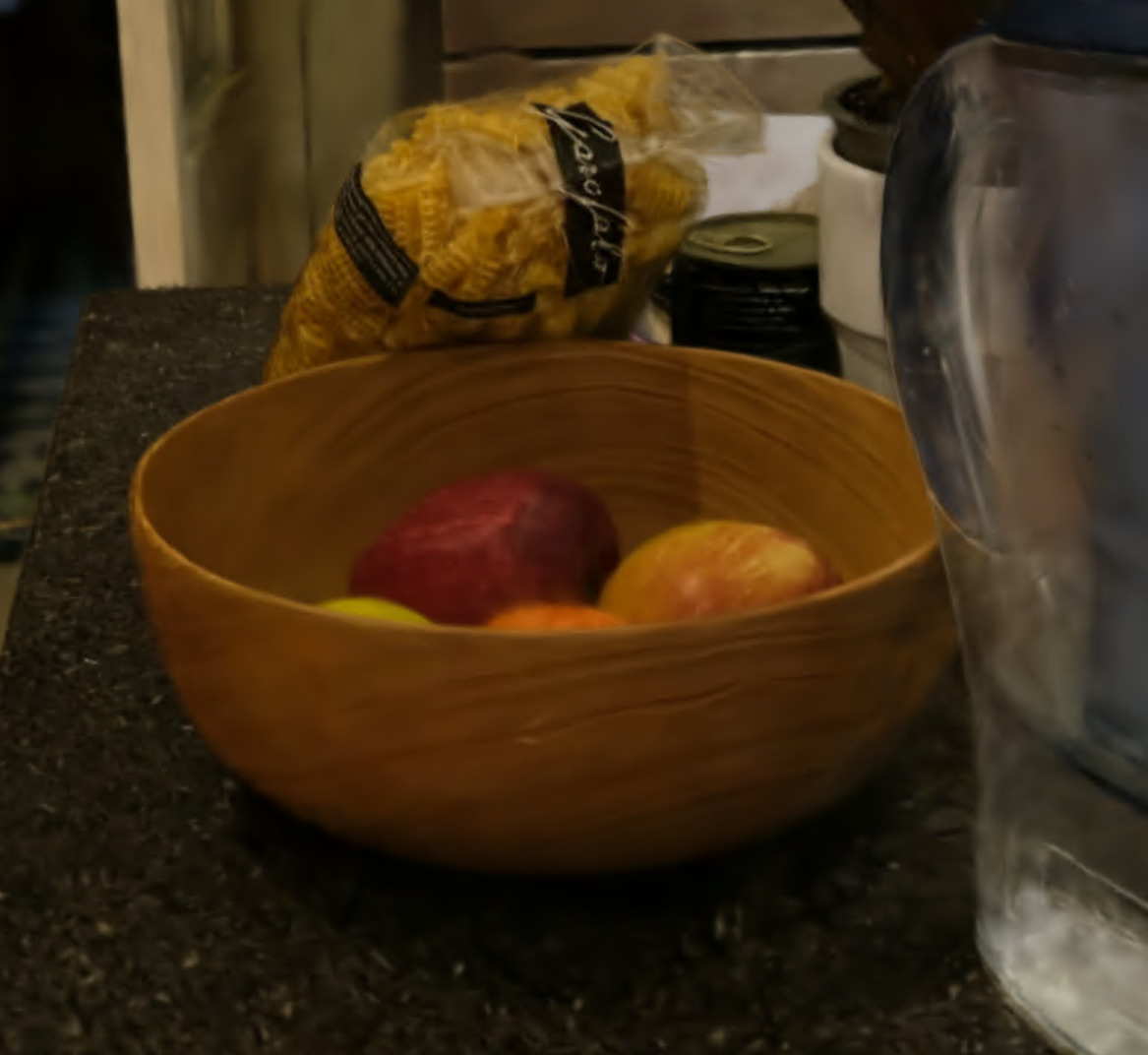} \\[2pt]
    (a) Before loop closure & (b) After loop closure
\end{tabular}
\vspace{-7pt}
\caption{\label{fig:loop_c_example}
    Loop Closure and Gaussian movement: (a) Through accumulated drift, Gaussians are optimized slightly wrong, thus introducing spiky artifacts and superimposition.
    (b) After loop closure with our Gaussian relocation, these artifacts can be removed.
}
\end{figure}

We employ loop closing when we detect disjoint clusters in the visual neighborhood through the covisibility graph.

\paragraph{Clustering}
We define $d_G(\mathcal{C}_1, \mathcal{C}_2) = \min_{I_i \in \mathcal{C}_1, I_j \in \mathcal{C}_2} \text{hops}(I_i, I_j)$ where $\text{hops}(I_i, I_j)$ is the shortest path length in the covisibility graph.
We also look at the size of the smaller cluster. We consider that a loop closure occurs if it has more than $\tau_v$ valid pairs and the clusters are separated by more than $\tau_{lc} = 5$ hops. Concretely, if:
\begin{equation}
d_G(\mathcal{C}_1, \mathcal{C}_2) > \tau_{lc} = 5 \text{ and } \min(|\mathcal{C}_1|, |\mathcal{C}_2|) > \tau_v
\end{equation}
loop closure is triggered.

\paragraph{Welding Window}
Formally, the welding window is defined as:
$\mathcal{W} = \mathcal{S}_{N/2}^{\mathcal{C}_{1}} \cup \mathcal{S}_{N/2}^{\mathcal{C}_{2}}$ where $\mathcal{S}_{t}^{\mathcal{C}}$ is the set of $t$ keyframes selected starting from cluster $\mathcal{C}$. 

\paragraph{Graph-Based Drift Correction}

In each cluster, we first select the set of most connected keyframes within the welding window $\mathcal{W}$: we can expect these \emph{anchor} keyframes to have the most reliable pose after the welding BA:
    \begin{equation}
        I_a^{1} = \operatorname{argmax}_{I \in \mathcal{S}_{N/2}^{\mathcal{C}_{1}}} c(I, \mathcal{W}), ~~~
        I_a^{2} = \operatorname{argmax}_{I \in \mathcal{S}_{N/2}^{\mathcal{C}_{2}}} c(I, \mathcal{W})
    \end{equation}
We then compute the rigid body transformation (SE(3)) between the initial and re-optimized pose for the anchor keyframe in the free cluster (i.e., the most recent keyframes):
    \begin{equation}
    \mathbf{T}_{\Delta} = \mathbf{T}_a^{\text{opt}} \left( \mathbf{T}_a^{\text{init}} \right)^{-1}
    \end{equation}
where $\mathbf{T}_a^{\text{opt}}, \mathbf{T}_a^{\text{init}} \in SE(3)$ denote the optimized and initial poses of $I_a^{1}$.
To obtain the full similarity transformation $\mathbf{S}_{\Delta} \in Sim(3)$, we estimate the scale change $s_{\Delta}$ from landmark depth variations and compose it with the rigid transform:
    $\mathbf{S}_{\Delta} = \text{Sim3}(\mathbf{T}_{\Delta}, s_{\Delta})$.

We now have one anchor in each cluster and the similarity transform between their previous and ``welded'' pose. We need to propagate this transformation to all poses efficiently. To do this we first compute a \emph{backbone path} composed of well-connected keyframes that are likely reliable.

We build a backbone path between the two keyframes along the covisibility graph, using Dijkstra's algorithm to find the path with the smallest cumulative edge weights:
    \begin{equation}
        \mathcal{B} = \text{Dijkstra}\left(I_a^{1}, I_a^{2}, G\right)
    \end{equation}
Note that loop edges have not yet been added to the graph, so the backbone goes through existing connections only. 
We then distribute the computed Sim(3) transformation along this backbone path, proportionally to the distance to the optimized keyframe:
    \begin{equation}
        \alpha(I_b) = \frac{d_G\left(I_b, I_a^{2}\right)}{d_G\left(I_a^{1}, I_a^{2}\right)} \quad \forall I_b \in \mathcal{B}
    \end{equation}
    \begin{equation}
        \mathbf{S}_{\Delta, b} = \text{interp}_{\text{Sim}(3)}(\mathbb{I}, \mathbf{S}_{\Delta}, \alpha(I_b))
    \end{equation}
where $\mathbb{I}$ is identity. We interpolate rotations with 6D representation \cite{Zhou_2019_CVPR} and scale using: $s_{\Delta, b} = s_{\Delta}^{\alpha(I_b)}$.

Finally, we propagate the corrections to the rest of the trajectory by moving each keyframe following the transformation of the closest backbone keyframe in terms of geodesic distance, and applying the same correction. We get the most relevant backbone node for this keyframe as follows:
    \begin{equation}
        I_{b^*} = \operatorname{argmin}_{I_b \in \mathcal{B}} d_G(I_i, I_b) \quad \forall I_i \notin \mathcal{W}
    \end{equation}
and apply the transformation to the pose:
    \begin{equation}
        \mathbf{T}_i^{\text{new}} = \mathbf{S}_{\Delta, b^*} \circ \mathbf{T}_i^{\text{old}}
    \end{equation}

After the loop closure has been treated, we add the loop matches to the covisibility graph so that they are used in subsequent local bundle adjustment.

\paragraph{Example}
Fig.~\ref{fig:loop_c_example} shows an example on the \textsc{Counter} scene from the MipNeRF360 dataset. The accumulated drift is rectified via loop closure, and all Gaussians are updated. This allows the pipeline to optimize clean Gaussian models. 

\begin{table*}[!ht]
\setlength{\tabcolsep}{3pt}
	\caption{\label{tab:slam-eval-s3po} 
	We show results for novel view synthesis on pose-free methods. We also include GLOMAP followed by Taming 3DGS (at 7k) (G+T3DGS) as an offline reference.
	The {\colorbox{firstcolor}{best}} and {\colorbox{secondcolor}{second best}} are color coded for pose-free methods.}
\vspace{-9pt}
\small
\begin{tabular}{l|cccc|cccc|cccc}
\toprule
 &
\multicolumn{4}{c|}{\textsc{TUM}} &
\multicolumn{4}{c|}{\textsc{MipNeRF360}} &
\multicolumn{4}{c}{\textsc{StaticHikes}}  \\
	{} & PSNR$^\uparrow$ & SSIM$^\uparrow$ & LPIPS$^\downarrow$ & Time$^\downarrow$ &
	PSNR$^\uparrow$ & SSIM$^\uparrow$ & LPIPS$^\downarrow$ & Time$^\downarrow$ &
	PSNR$^\uparrow$ & SSIM$^\uparrow$ & LPIPS$^\downarrow$ & Time$^\downarrow$ \\
\midrule
Photo-SLAM & 
{19.30} & {0.700} & 0.382 & {0:02:12} & {16.54} & {0.505} & {0.603} & {0:02:11} & 14.13 & {0.316} & 0.660 & \cellcolor{secondcolor}{0:02:01} \\
MonoGS & 
16.60 & 0.682 & {0.381} & 0:16:18 & 14.46 & 0.436 & 0.663 & 0:04:05 & {15.46} & 0.301 & {0.659} & 0:09:19 \\
On-The-Fly NVS & 
{23.02} & {0.821} & {0.250} & \cellcolor{firstcolor}{0:00:50} & \cellcolor{secondcolor}{24.31} & \cellcolor{secondcolor}{0.775} & \cellcolor{secondcolor}{0.300} & \cellcolor{firstcolor}{0:01:02} & \cellcolor{secondcolor}{20.40} & \cellcolor{secondcolor}{0.589} & \cellcolor{secondcolor}{0.365} & \cellcolor{firstcolor}{0:01:30} \\
S3PO-GS & 
{10.79} & {0.729} & {0.341} & {1:26:49} & {18.75} & {0.518} & {0.589} & {0:18:52} & {14.87} & {0.316} & {0.640} & {0:50:43} \\
S3PO-GS w\texttt{\textbackslash} refinement & 
{21.88} & {0.777} & {0.309} & {1:29:04} & {19.82} & {0.553} & {0.576} & {0:23:29} & {18.09} & {0.368} & {0.611} & {0:54:44} \\
LongSplat  & 
\cellcolor{firstcolor}{26.51} & \cellcolor{firstcolor}{0.875} & \cellcolor{firstcolor}{0.176} & {1:41:08} & {23.63} & {0.657} & {0.415} & {2:25:26} & {20.05} & {0.461} & {0.481} & {22:40:56} \\
Ours &\cellcolor{secondcolor}{24.77} & \cellcolor{secondcolor}{0.853} & \cellcolor{secondcolor}{0.205} & \cellcolor{secondcolor}{0:01:22} & \cellcolor{firstcolor}{26.29} & \cellcolor{firstcolor}{0.839} & \cellcolor{firstcolor}{0.241} & \cellcolor{secondcolor}{0:01:17} & \cellcolor{firstcolor}{22.12} & \cellcolor{firstcolor}{0.696} & \cellcolor{firstcolor}{0.289} & {0:02:14} \\
\hline
\hline
G+T3DGS (7k) & 
25.29 & 0.868 & 0.191 & 0:03:33 & 27.52 & 0.866 & 0.226 & 0:08:50 & 20.23 & 0.537 & 0.465 & 1:02:34 \\
\bottomrule
\end{tabular}
\end{table*}

\begin{table*}[!ht]
\caption{\label{tab:large-scale-s3po} Results for large scale scenes. For other methods time includes total time for COLMAP using the H3DGS and Octree-GS process and the actual 3DGS optimization of each method. The {\colorbox{firstcolor}{best}} and {\colorbox{secondcolor}{second best}} are color coded.}
\vspace{-9pt}
\small
\setlength{\tabcolsep}{3pt}
\begin{tabular}{l|cccc|cccc|cccc}
\toprule
 &
\multicolumn{4}{c|}{\textsc{SmallCity*}} &
\multicolumn{4}{c|}{\textsc{Wayve*}} &
\multicolumn{4}{c}{\textsc{CityWalk}}  \\
  & PSNR$^\uparrow$ & SSIM$^\uparrow$ & LPIPS$^\downarrow$ & Time$^\downarrow$ &
	PSNR$^\uparrow$ & SSIM$^\uparrow$ & LPIPS$^\downarrow$ & Time$^\downarrow$ &
	PSNR$^\uparrow$ & SSIM$^\uparrow$ & LPIPS$^\downarrow$ & Time$^\downarrow$ \\
\midrule
{H3DGS} & 
21.17 & 0.679 & \second{}0.285 & 2:55:28 & 20.80 & 0.737 & \first{}0.227 & 7:29:45 & 11.78 & 0.557 & 0.560 & 22:09:20 \\
Octree-GS &  \first{}{24.18} & \first{}{0.831} & \first{}{0.273} & {1:20:59} &\second{} {22.58} & \second{}{0.765} & {0.313} & {2:21:48} & {17.33} & {0.638} & {0.520} & {3:11:40} \\
\hline
S3PO-GS & {18.24} & {0.610} & {0.510} & {0:53:43} & {13.32} & {0.602} & {0.459} & {1:20:37} & {5.95} & {0.102} & {0.645} & {5:57:13} \\ 
S3PO-GS w\texttt{\textbackslash} refinement & {19.71} & {0.652} & {0.462} & {0:57:31} & {17.81} & {0.644} & {0.423} & {1:25:26} & {10.72} & {0.439} & {0.579} & {6:00:28} \\
{On-The-Fly NVS} & \second{}23.59 & 0.789 & 0.323 &\first{}0:01:45 & 20.29 & 0.739 & 0.303 & \second{}0:04:29 & \second{}21.71 & \second{}0.712 & \second{}0.395 & \first{}0:25:03 \\
Ours& 23.56&\second{}0.800&0.302 & \second{}0:02:25 & \first{}23.04&\first{}0.822&\second{}	0.234&\first{}0:04:12 & \first{}22.03& \first{}0.772	& \first{} 0.336	& \second{}0:29:38 \\
\bottomrule
\end{tabular}
\end{table*}

\begin{table*}[!hb]
\caption{\label{tab:slam-eval-low-res} {Novel view quality results for methods that require low-resolution input.}}
\vspace{-9pt}
{
\begin{tabular}{l|cccc|cccc|cccc}
\toprule
 &
\multicolumn{4}{c|}{\textsc{TUM}} &
\multicolumn{4}{c|}{\textsc{MipNeRF360}} &
\multicolumn{4}{c}{\textsc{StaticHikes}}  \\
	{} & PSNR$^\uparrow$ & SSIM$^\uparrow$ & LPIPS$^\downarrow$ & Time$^\downarrow$ &
	PSNR$^\uparrow$ & SSIM$^\uparrow$ & LPIPS$^\downarrow$ & Time$^\downarrow$ &
	PSNR$^\uparrow$ & SSIM$^\uparrow$ & LPIPS$^\downarrow$ & Time$^\downarrow$ \\
\midrule
DROID-Splat &
{19.49} & {0.721} & {0.325} & {0:06:35} & \second {25.87} & {0.776} & {0.258} & {0:11:18} & {19.65} & {0.470} & {0.506} & {0:09:26} \\
CF-3DGS &
15.05 & 0.578 & 0.405 & 1:10:27 & 13.52 & 0.295 & 0.621 & 1:08:14 & 15.21 & 0.301 & 0.560 & 7:52:54 \\
On-The-Fly NVS  &
\second{22.45} & \second{0.815} & \second{0.225} & \first{0:01:11} & {25.80} & \second{0.834} & \second{0.182} & \first{0:00:55} & \second{21.93} & \second{0.673} & \second{0.270} & \first{0:01:32} \\
Ours & \first{24.51} & \first{0.861} & \first{0.174} & \second{0:01:36} & \first{26.99} & \first{0.850} & \first{0.169} & \second{0:00:59} & \first{23.78} & \first{0.760} & \first{0.242} & \second{0:01:49} \\
\bottomrule
\end{tabular}
}
\end{table*}

\begin{figure*}[!h]
\setlength{\tabcolsep}{1pt}
\renewcommand{\arraystretch}{0}
\begin{tabular}{cccccc}
    OTF-NVS & S3PO-GS & MonoGS & Photo-SLAM & Ours & GT \\[2pt]
    {\includegraphics[width=0.16\textwidth]{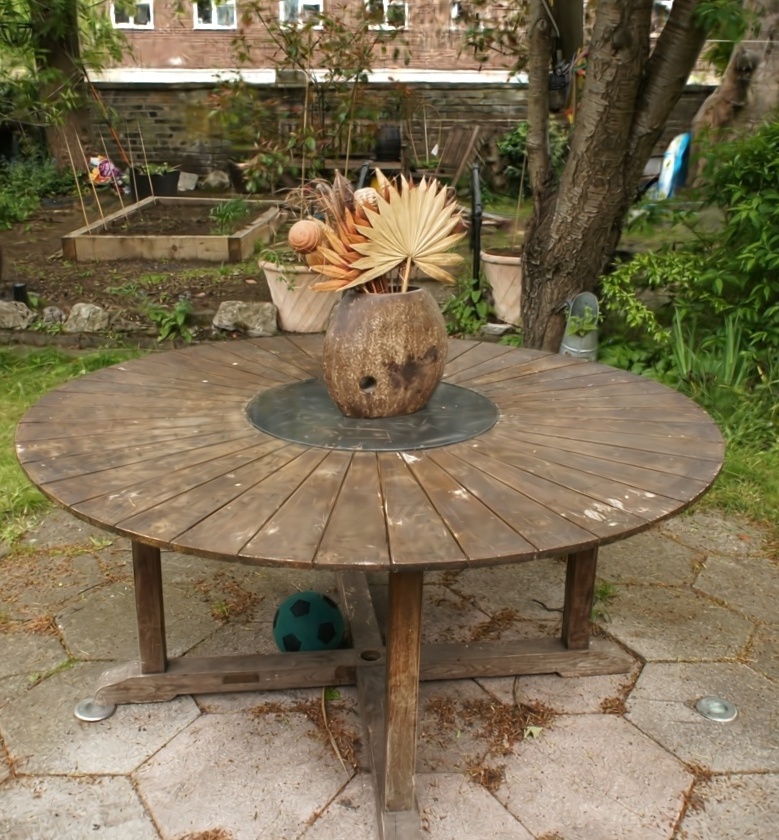}} &
    {\includegraphics[width=0.16\textwidth]{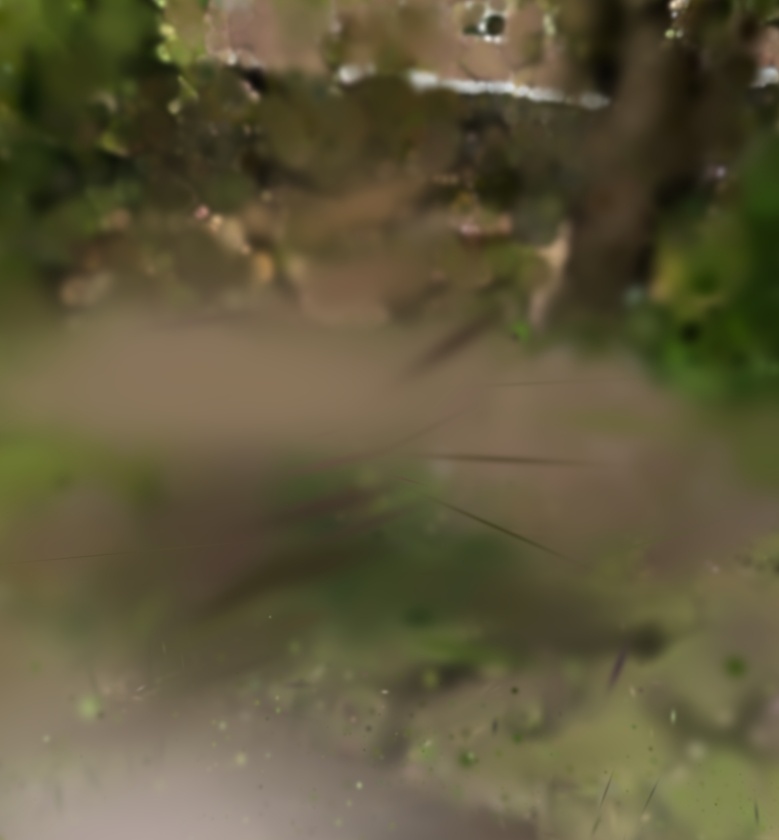}} &
    {\includegraphics[width=0.16\textwidth]{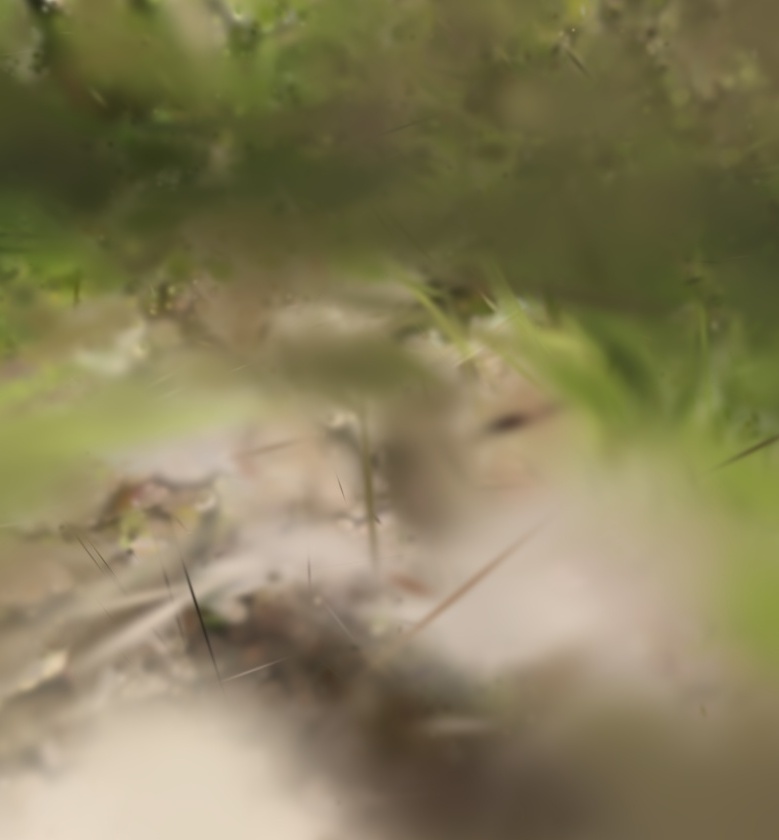}} &
    {\includegraphics[width=0.16\textwidth]{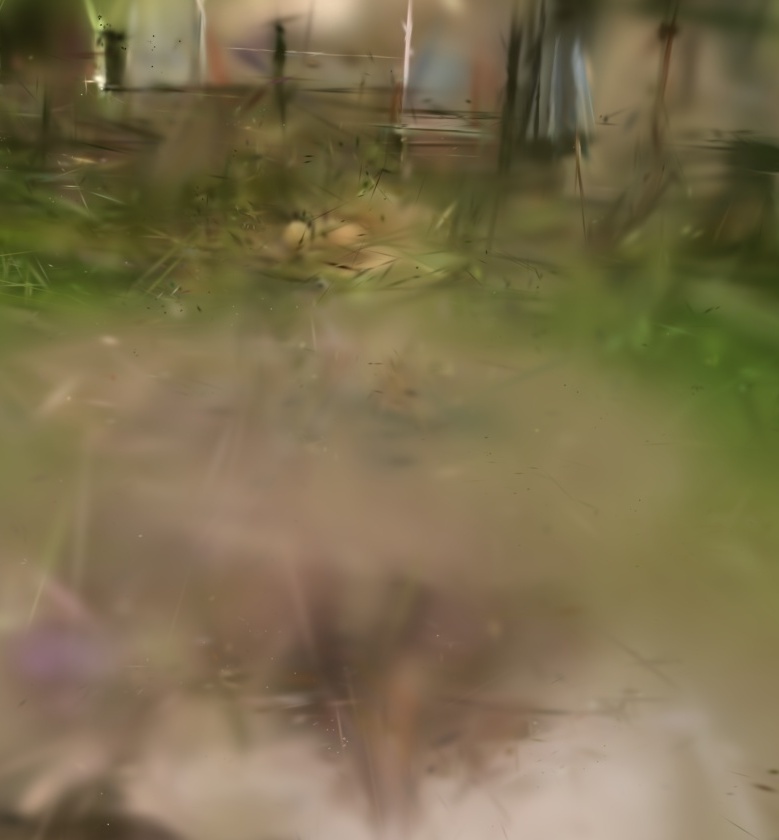}} &
    {\includegraphics[width=0.16\textwidth]{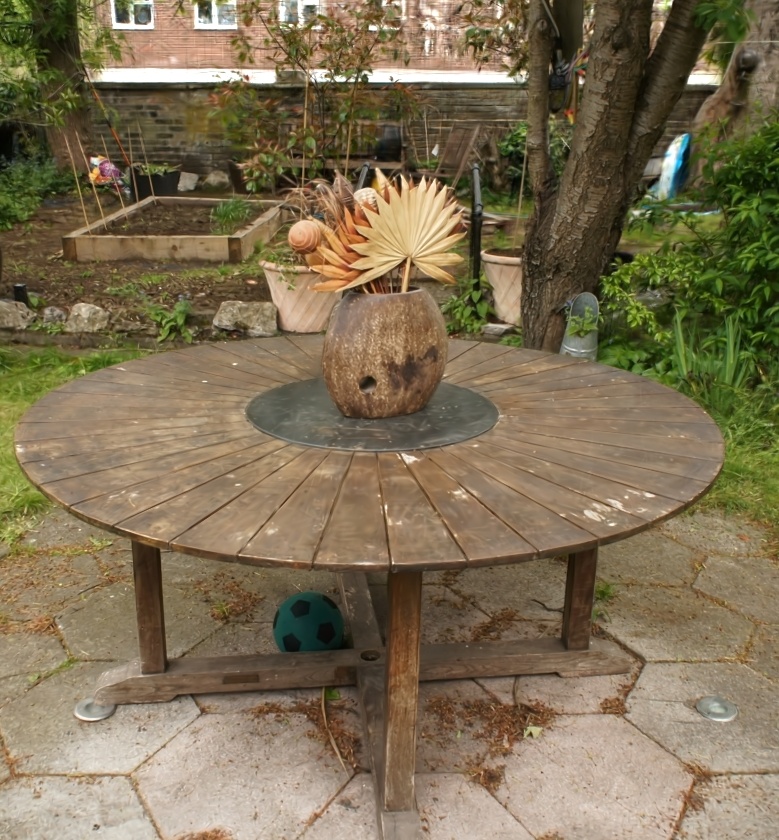}} &
    {\includegraphics[width=0.16\textwidth]{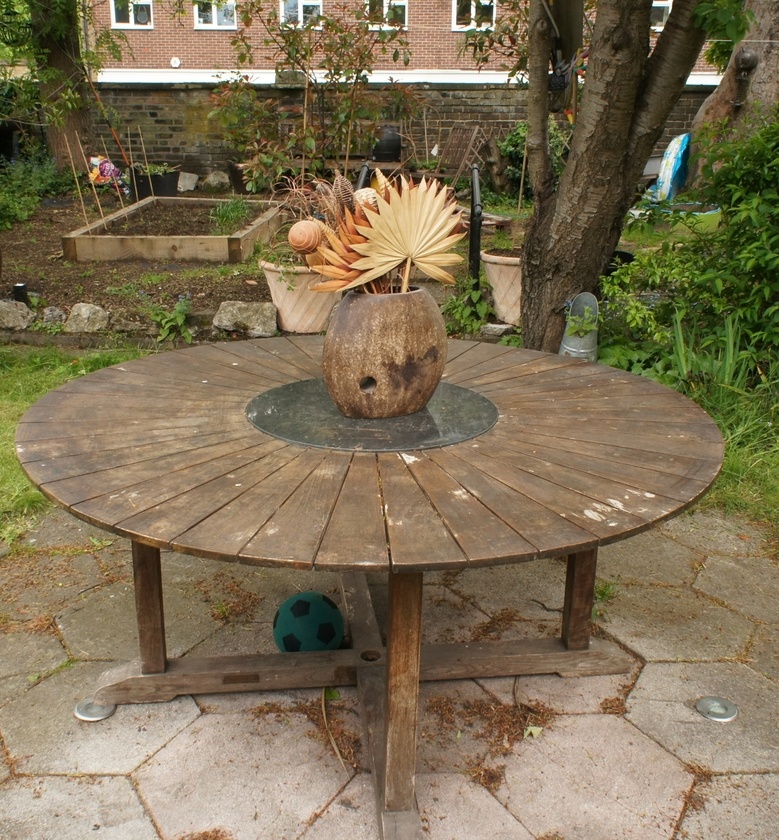}} \\[2pt]

    {\includegraphics[width=0.16\textwidth]{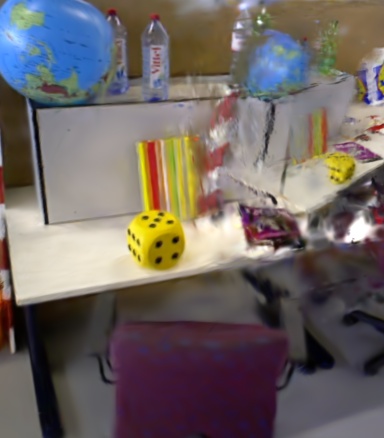}} &
    {\includegraphics[width=0.16\textwidth]{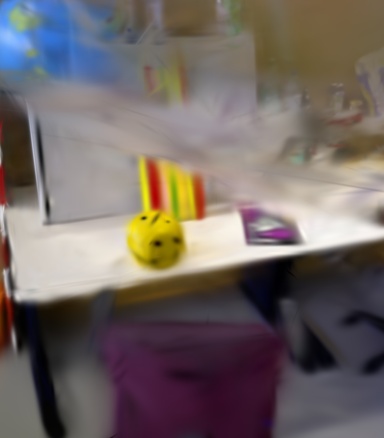}} &
    {\includegraphics[width=0.16\textwidth]{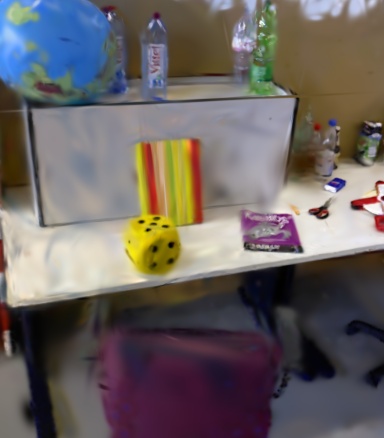}} &
    {\includegraphics[width=0.16\textwidth]{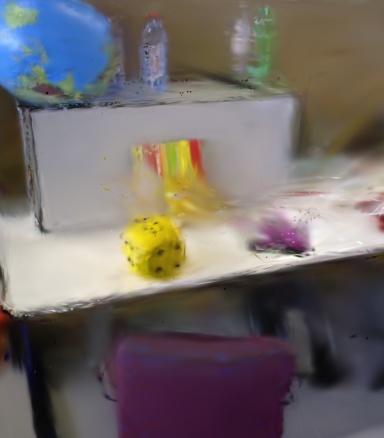}} &
    {\includegraphics[width=0.16\textwidth]{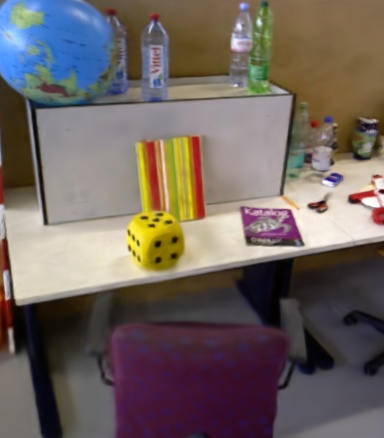}} &
    {\includegraphics[width=0.16\textwidth]{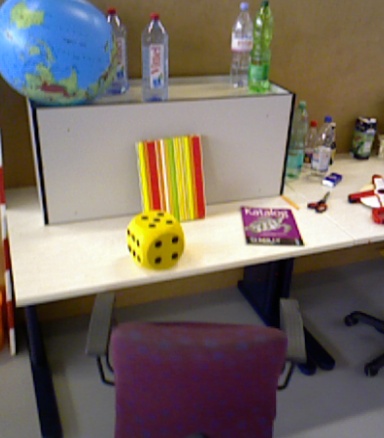}} \\[2pt]

    {\includegraphics[width=0.16\textwidth]{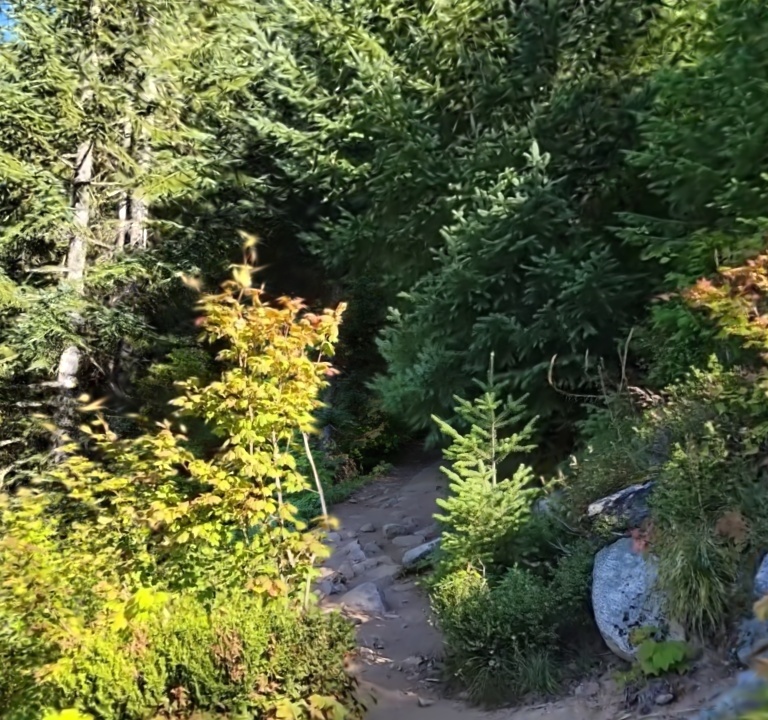}} &
    {\includegraphics[width=0.16\textwidth]{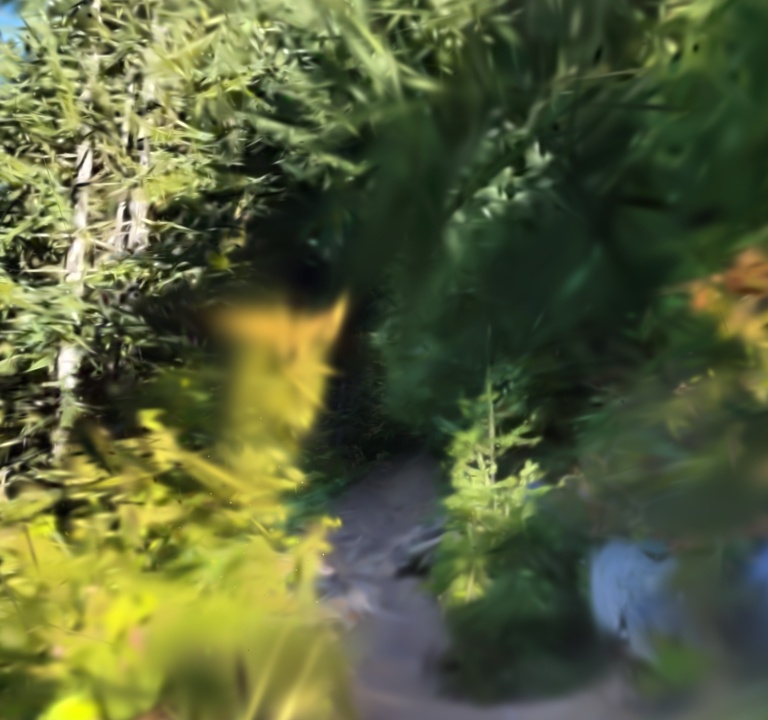}} &
    {\includegraphics[width=0.16\textwidth]{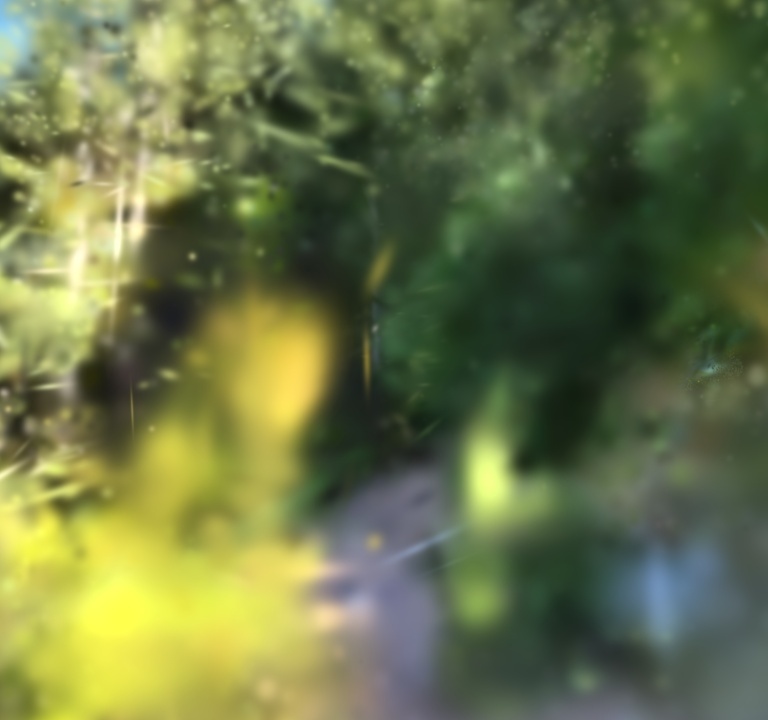}} &
    {\includegraphics[width=0.16\textwidth]{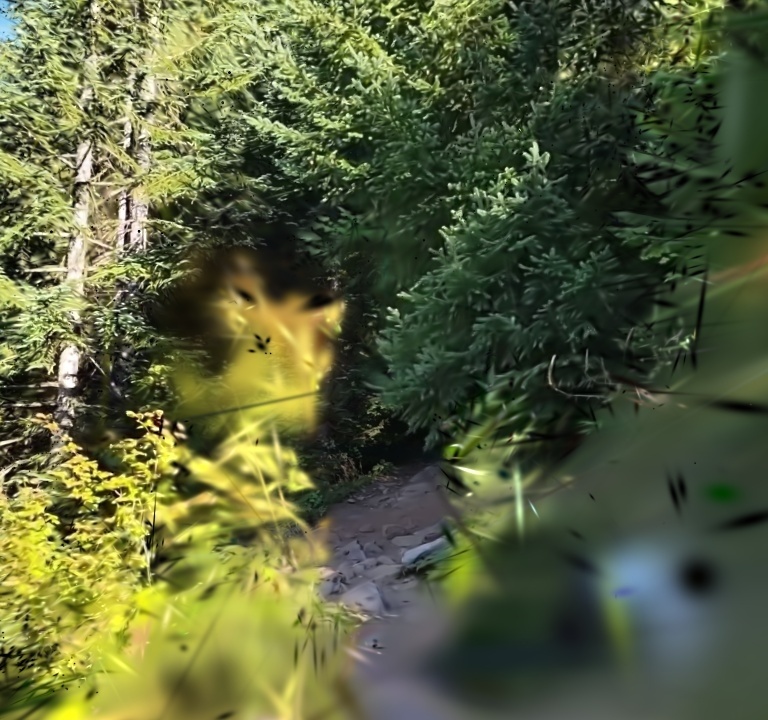}} &
    {\includegraphics[width=0.16\textwidth]{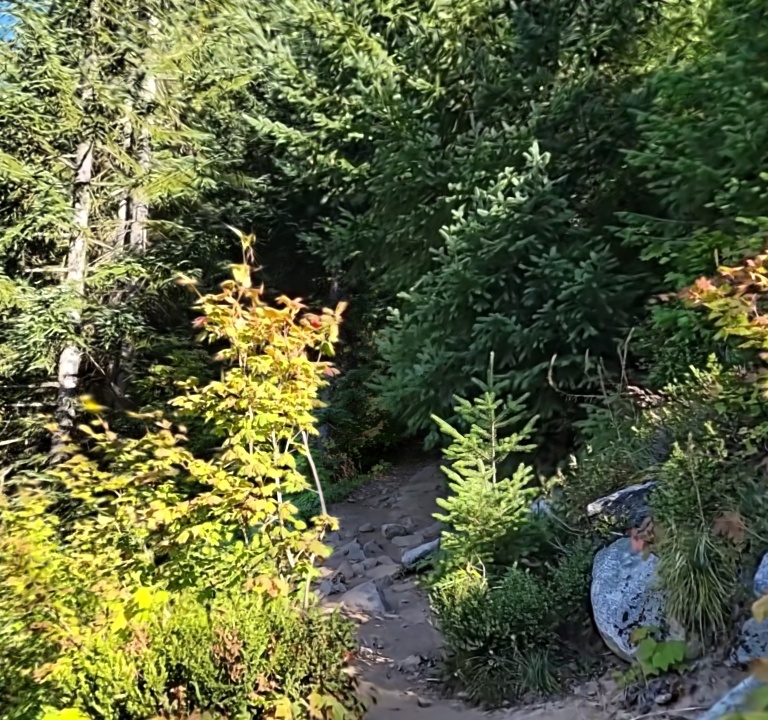}} &
    {\includegraphics[width=0.16\textwidth]{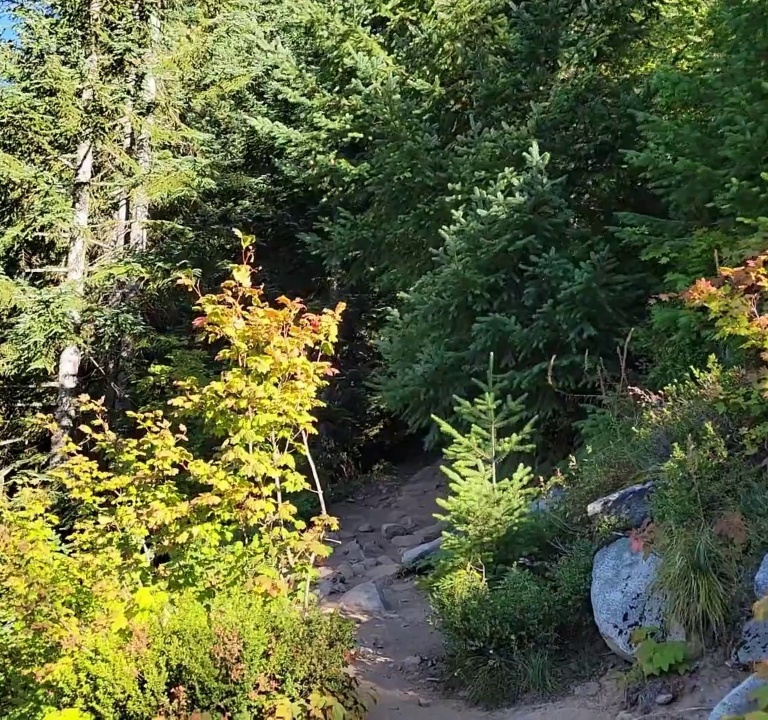}} \\[2pt]
\end{tabular}
\vspace{-9pt}
\caption{
\label{fig:results_page_supp}
{Additional comparisons.}
}
\end{figure*}

\section{Additional Evaluations}
\subsection{Novel View Synthesis}
In addition to the numbers in the main paper, we present additional qualitative results (Fig.~\ref{fig:results_page_supp}) and metrics comparing S3PO-GS's fast version without refinement in Tab.~\ref{tab:slam-eval-s3po}. 
As seen, while it provides good quality on the MipNeRF dataset with about 20\% less training time, neither reaches our quality. Similar results can be seen on the large-scale scenes in Tab.~\ref{tab:large-scale-s3po}.

\paragraph{Low resolution} 
DROID-Splat and CF-3DGS go OOM with full resolution images. Tab.~\ref{tab:slam-eval-low-res} shows results with images resized to 446x336 for \textsc{TUM} and 640 width for \textsc{MipNeRF360} and \textsc{StaticHikes}. 

\paragraph{AnySplat}
AnySplat~\cite{jiang2025anysplat} provides a large vision model backbone and directly predicts Gaussians from images. This limits image sizes to $448\times448$ pixels, and is quite expensive with regard to VRAM, limiting applicability: the method runs out of memory on an H100 on StaticHikes.
\begin{table}[!h]
\small
\setlength{\tabcolsep}{1.5pt}
\caption{\label{tab:anysplat} Comparison against AnySplat that requires rescaled and cropped images to 448$\times$448 pixels.}
\vspace{-9pt}
\begin{tabular}{l|cccc}
\toprule
 & PSNR$^\uparrow$ & SSIM$^\uparrow$ & LPIPS$^\downarrow$ & Time$^\downarrow$  \\
\midrule
AnySplat & 18.05	&0.43 & 0.39 & 0:00:25\\
AnySplat +1k & 21.53	&0.60& 0.30 & 0:02:10\\
Ours & 24.98 &0.81 & 0.21 & 0:00:40\\
\bottomrule
\end{tabular}
\end{table}

In Tab.~\ref{tab:anysplat}, we compare our approach in this low-resolution setup on the MipNeRF360 dataset. Both the feed-forward setup and the 1K iteration post-optimization method described by them fail to reach our quality.

{
\vspace{1cm}
\paragraph{Fine-tuning}
Table~\ref{tab:fine-tuning} shows the results of fine-tuning after the initial reconstruction on \textsc{MipNeRF360}.
\begin{table}[!h]
\setlength{\tabcolsep}{3pt}
	\caption{\label{tab:fine-tuning} 
	Results with additional fine-tuning epochs after the initial reconstruction. 
    }
\vspace{-9pt}
{
\small
\begin{tabular}{l|cccc}
\toprule
{} & PSNR$^\uparrow$ & SSIM$^\uparrow$ & LPIPS$^\downarrow$ & Time$^\downarrow$ \\
\midrule
Ours & {26.29} & {0.839} & {0.241} & {0:01:17} \\
Ours + 10 & 27.30 & 0.855 & 0.215 & 0:01:38 \\
Ours + 20 & 27.69 & 0.862 & 0.206 & 0:02:00 \\
Ours + 30 & 27.86 & 0.864 & 0.202 & 0:02:22 \\
\hline
\hline
G+T3DGS (7k) & 27.52 & 0.866 & 0.226 & 0:08:50 \\
\bottomrule
\end{tabular}
}
\end{table}

}

{
\subsection{Additional Ablations}
\paragraph{Hyperparameter sensitivity}
Tab.~\ref{tab:hyperparam} shows the sensitivity of our method to $\tau_{lc}$ and $\tau_v$ on \textsc{MipNeRF360}.
\begin{table}[!h]
\setlength{\tabcolsep}{3pt}
	\caption{\label{tab:hyperparam} Impact of the hyperparameters.
    }
\vspace{-9pt}
{
\small
\begin{tabular}{l|ccc}
\toprule
{} & PSNR$^\uparrow$ & SSIM$^\uparrow$ & LPIPS$^\downarrow$ \\
\midrule
$\tau_{lc} = 3$ & 26.22 & 0.831 & 0.239 \\
$\tau_{lc} = 7$ & 26.24 & 0.835 & 0.236 \\
$\tau_v = 1$ & 26.11 & 0.831 & 0.240 \\
$\tau_v = 3$ & 26.20 & 0.832 & 0.238 \\
Default & 26.29 & 0.839 & 0.241 \\
\bottomrule
\end{tabular}
}
\end{table}

}

{
\paragraph{Ablations on unordered scenes} 
We provide additional ablations on \textsc{MipNeRF360} including unordered scenes (Tab.~\ref{tab:unordered_ablations}).
\begin{table}
\setlength{\tabcolsep}{3pt}
	\caption{\label{tab:unordered_ablations} Ablations including unordered scenes.
    }
\vspace{-7pt}
{
\small
\begin{tabular}{l|ccc}
\toprule
{} & PSNR$^\uparrow$ & SSIM$^\uparrow$ & LPIPS$^\downarrow$ \\
\midrule
\textsc{NoMatchVerif} & 24.99 & 0.750 & 0.285 \\
\textsc{NoKFSelection} & 23.18 & 0.662 & 0.354 \\
\textsc{NoRandFixed} & 24.93 & 0.747 & 0.291 \\
\textsc{NoGraphProp} & 25.27 & 0.757 & 0.279 \\
\textsc{NoGSUpdate} & 25.00 & 0.750 & 0.291 \\
\textsc{NoLoDepAlgn} & 24.99 & 0.758 & 0.282 \\
Ours & 25.60 & 0.772 & 0.270 \\
\bottomrule
\end{tabular}
}
\end{table}

}

{
\paragraph{Pose quality ablation}
The impact of remaining components on the pose estimation quality is shown in Tab.~\ref{tab:pose_ablations}. Note that \textsc{NoGSUpdate} and \textsc{NoLoDepAlgn} have no impact on pose estimation and are therefore excluded from the table.
\begin{table}[!h]
\setlength{\tabcolsep}{3pt}
	\caption{\label{tab:pose_ablations} Impact of each component on pose estimation.
    }
\vspace{-7pt}
{
\begin{tabular}{l|cccc}
\toprule
 & T.APE$^\downarrow$ & R.APE$^\downarrow$ & T.RPE$^\downarrow$ & R.RPE$^\downarrow$ \\
\midrule
\textsc{NoVPR} & 10.62 & 0.03 & 15.33 & 0.04 \\
\textsc{NoMatchVerif} & 8.70 & 0.03 & 14.28 & 0.04 \\
\textsc{NoKFSelection} & 10.33 & 0.03 & 15.85 & 0.04 \\
\textsc{NoRandFixed} & 8.51 & 0.02 & 14.07 & 0.04 \\
\textsc{NoGraphProp} & 8.63 & 0.03 & 14.28 & 0.04 \\
\textsc{Ours} & 8.49 & 0.03 & 14.21 & 0.04 \\
\bottomrule
\end{tabular}
}
\end{table}

}

\begin{figure}[!h]
\setlength{\tabcolsep}{1pt}
\renewcommand{\arraystretch}{0}
\begin{tabular}{cccc}
    \includegraphics[width=0.242\linewidth]{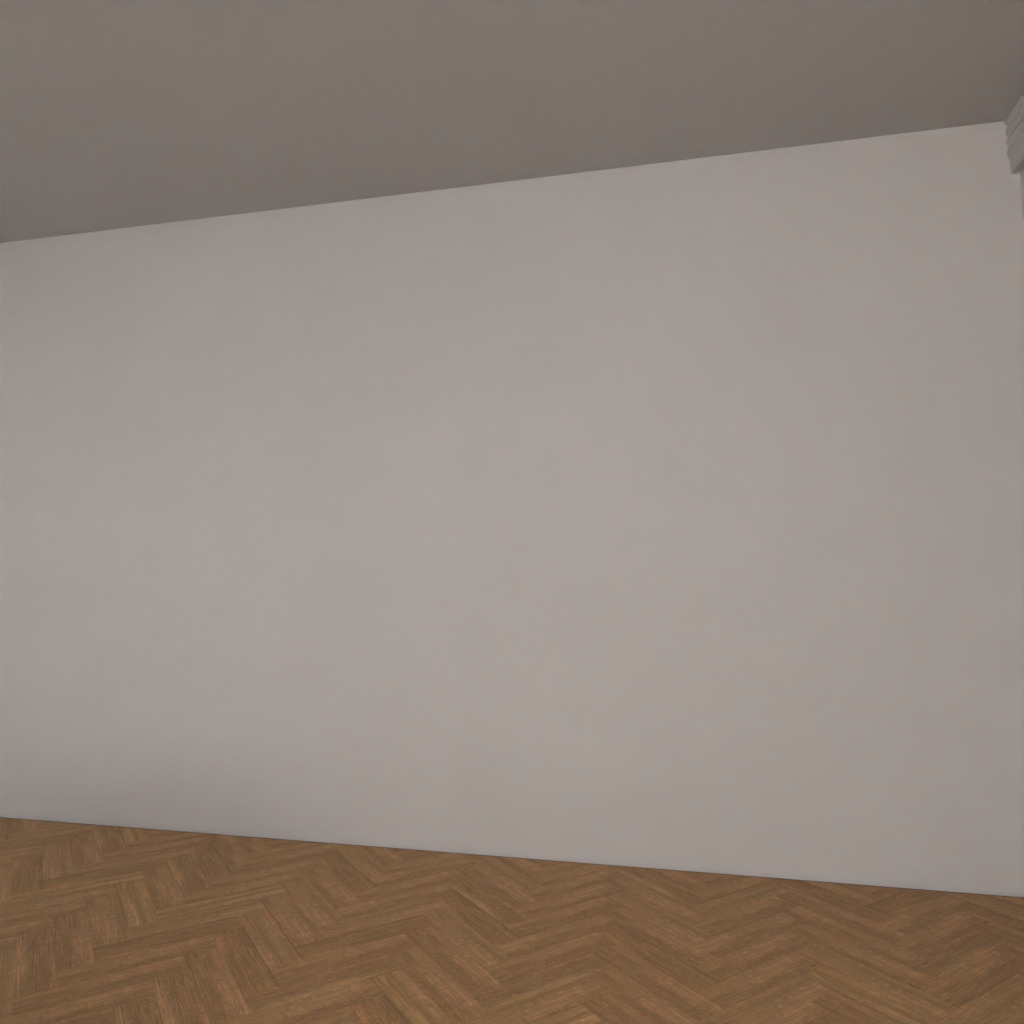} &
    \includegraphics[width=0.242\linewidth]{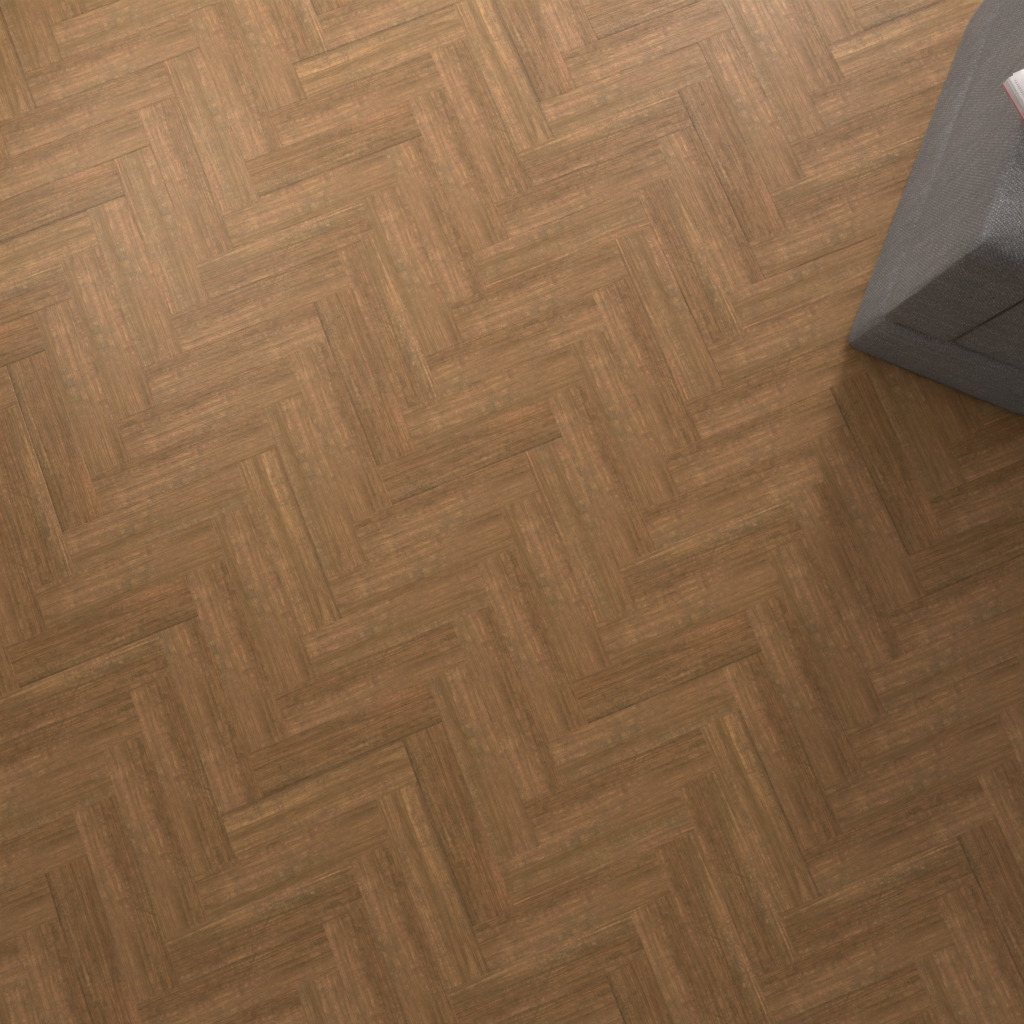} &
    \includegraphics[width=0.242\linewidth]{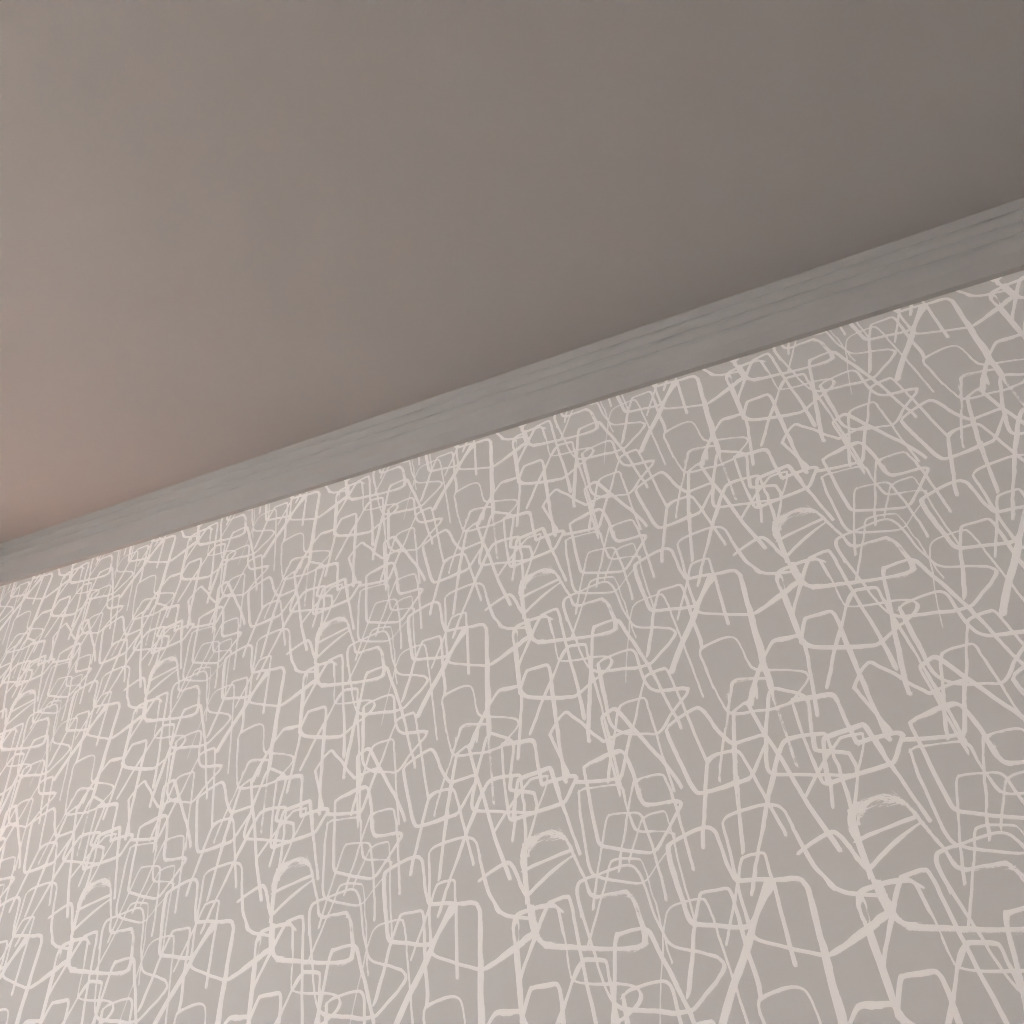} &
    \includegraphics[width=0.242\linewidth]{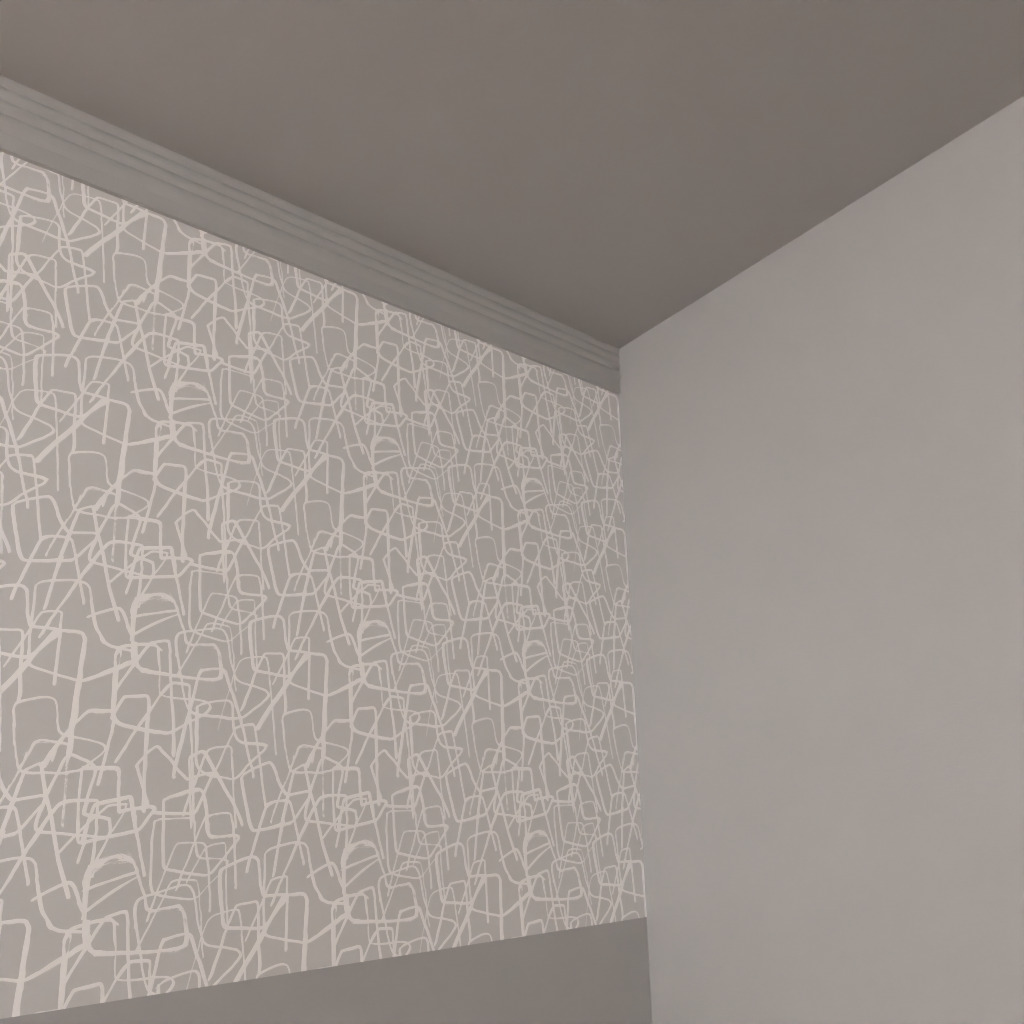}
\end{tabular}
\vspace{-9pt}
\caption{\label{fig:repeat}
Scene with textureless areas and artificial repetitions}
\end{figure}

\paragraph{VPR robustness}
In all scenes tested, none failed due to VPR thanks to geometric verification (XFeat+LightGlue+RANSAC), which eliminates false positives. Top-5 by MixVPR and by geometric inliers have overlap in 100\% of frames on MipNeRF360, and 92\% on a synthetic scene with artificial repetitions and low-texture areas (Fig.~\ref{fig:repeat}).
\vspace{1cm}

\begin{figure*}[!b]
\centering
\setlength{\tabcolsep}{1pt}
\renewcommand{\arraystretch}{0}
\newsavebox\limA
\sbox\limA{%
    \begin{minipage}{0.395\textwidth}%
        \begin{tabular}{cccc}
            \includegraphics[width=0.24\linewidth]{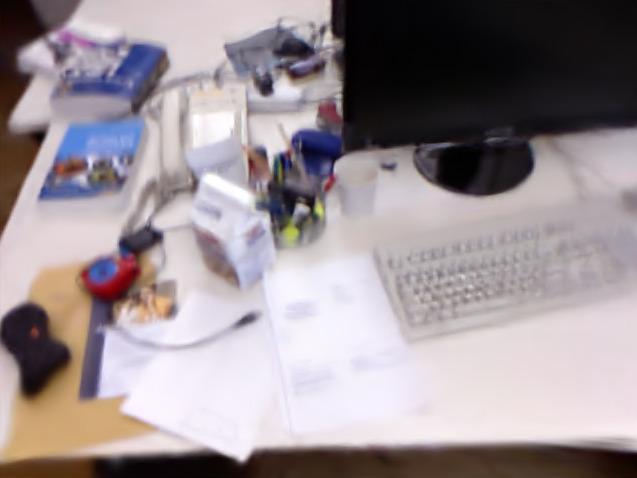} &
            \includegraphics[width=0.24\linewidth]{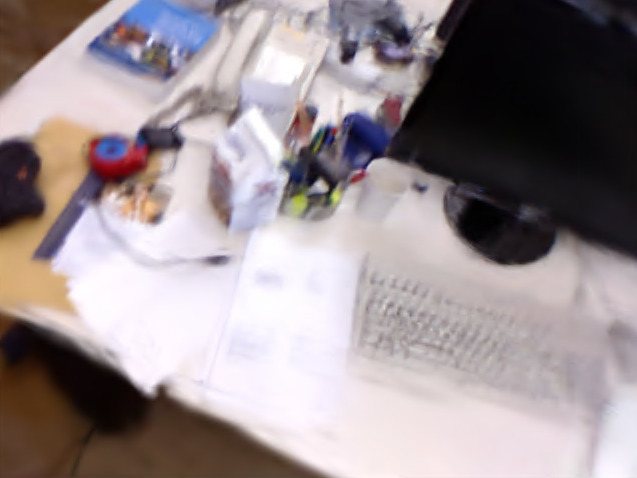} &
            \includegraphics[width=0.24\linewidth]{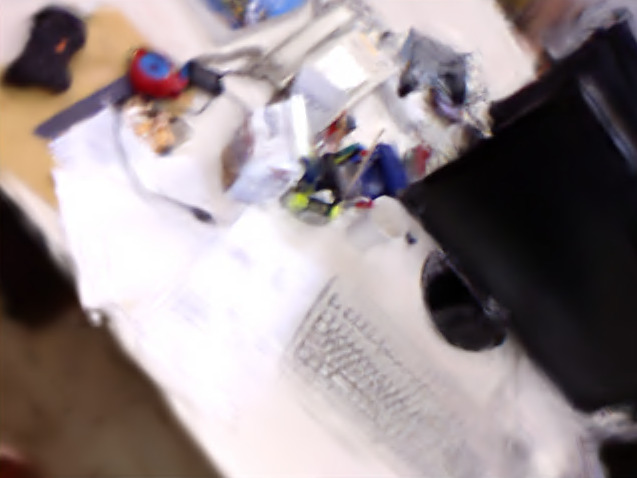} &
            \includegraphics[width=0.24\linewidth]{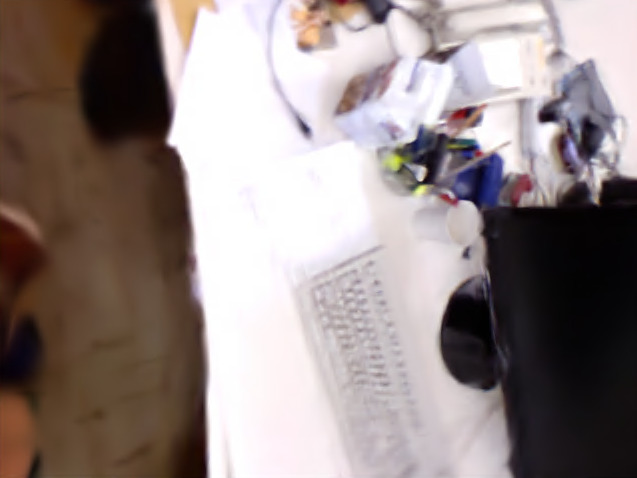} \\[2pt]
            \includegraphics[width=0.24\linewidth]{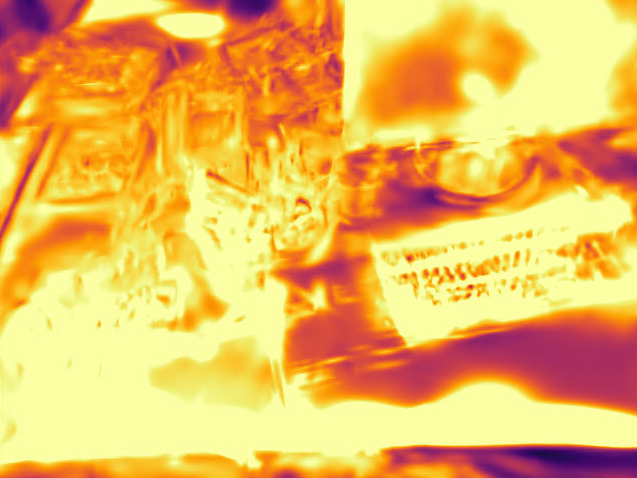} &
            \includegraphics[width=0.24\linewidth]{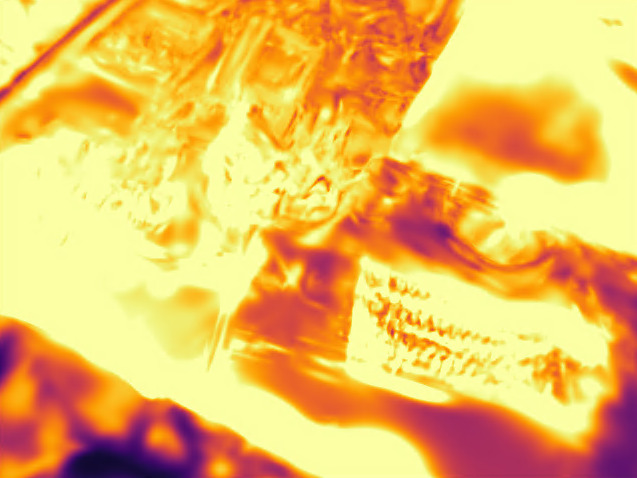} &
            \includegraphics[width=0.24\linewidth]{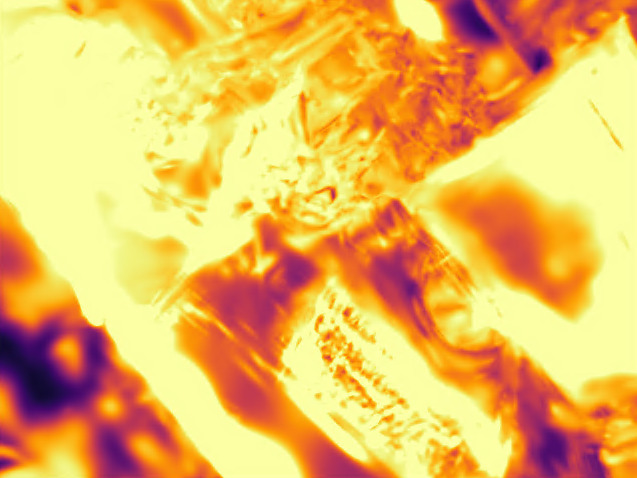} &
            \includegraphics[width=0.24\linewidth]{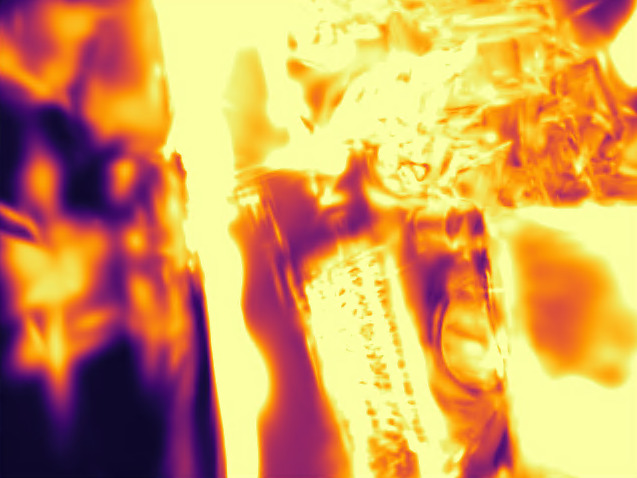}
        \end{tabular}%
    \end{minipage}%
}
\newdimen\limH
\limH=\ht\limA \advance\limH by \dp\limA
\resizebox{1\textwidth}{!}{%
\begin{tabular}{ccc}
    \begin{minipage}[t][\limH][t]{0.395\textwidth}\vspace*{0pt}%
        \usebox\limA
    \end{minipage} &
    \begin{minipage}[t][\limH][t]{0.195\textwidth}\vspace*{0pt}%
        \includegraphics[width=\linewidth]{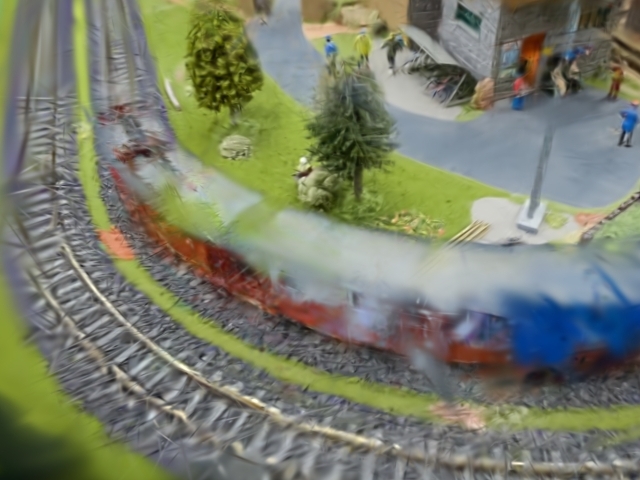}%
    \end{minipage} &
    \begin{minipage}[t]{0.38\textwidth}\vspace*{0pt}%
        \begin{tabular}{cc}
            \includegraphics[width=0.49\linewidth]{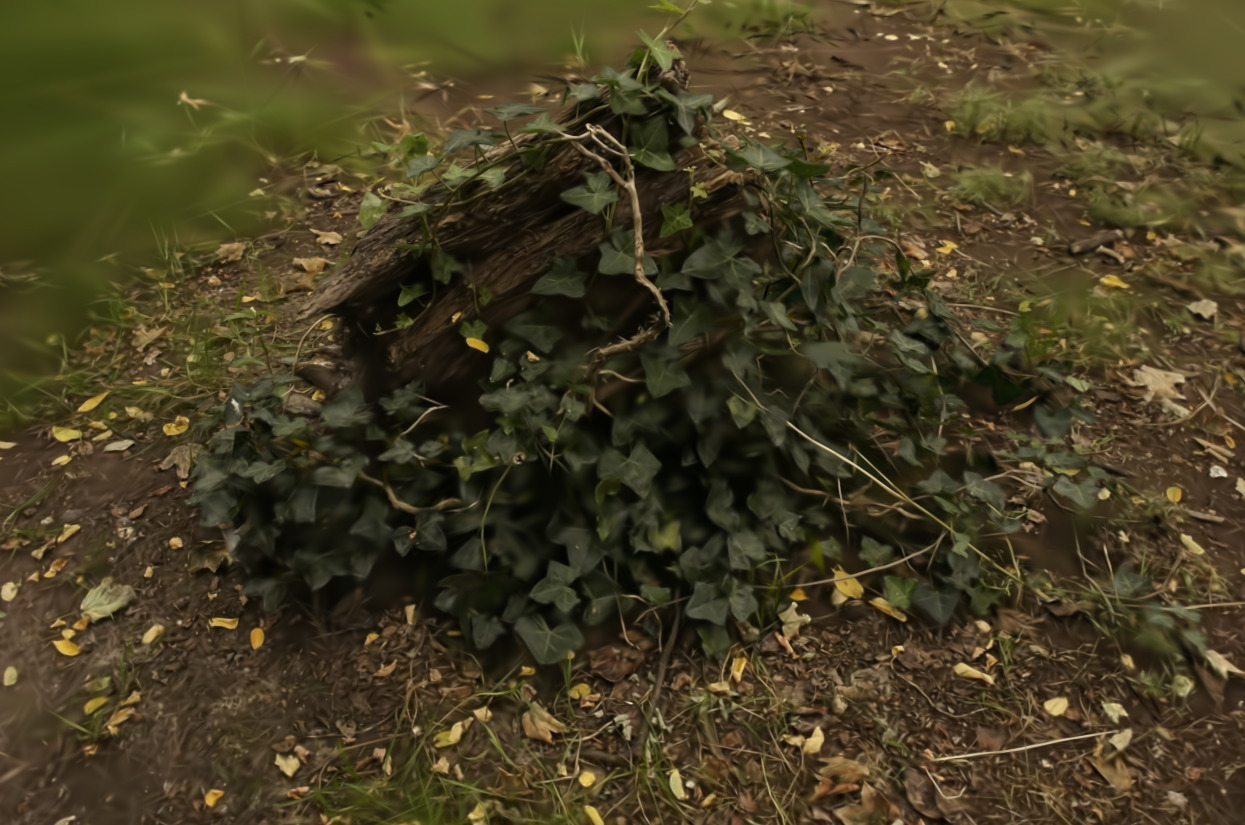} &
            \includegraphics[width=0.49\linewidth]{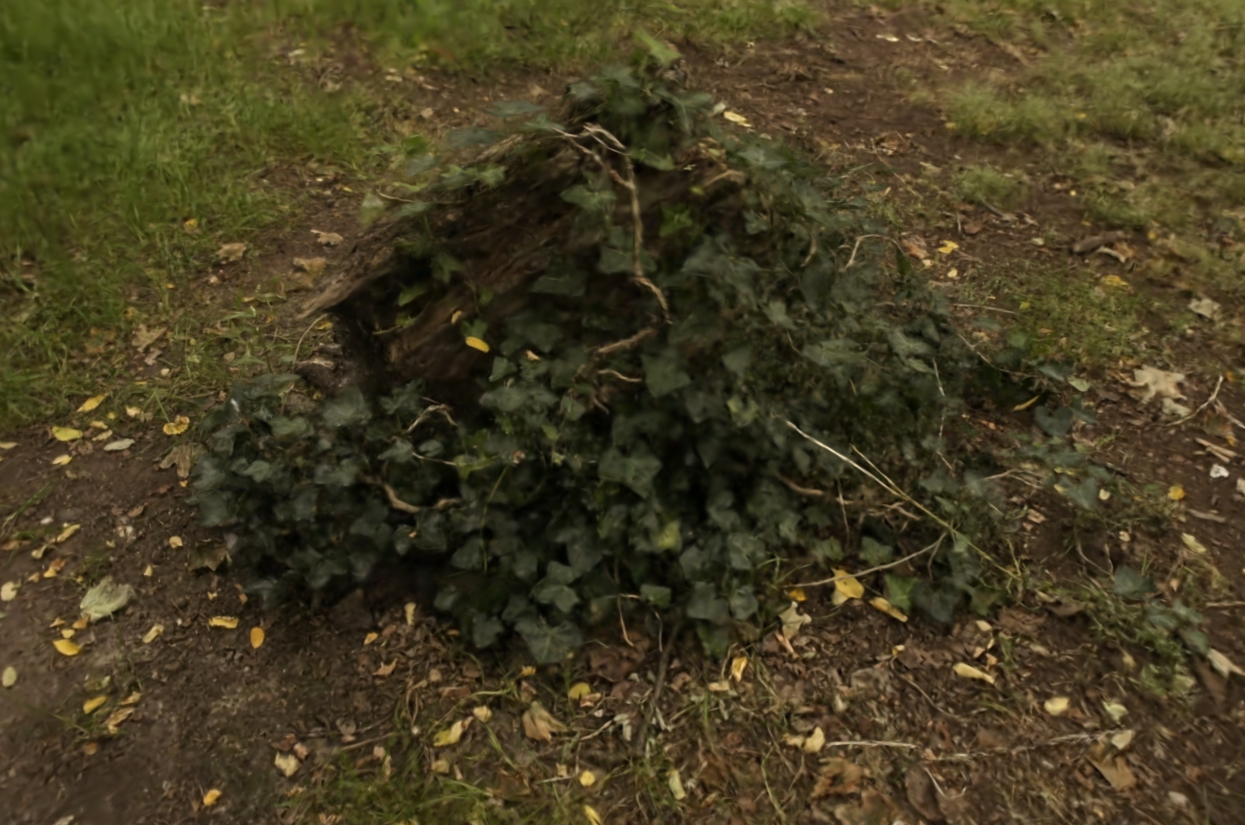} \\[2pt]
            GLOMAP+T3DGS (7k) & Ours \\[3pt]
            \hline
            \\[2pt]
        \end{tabular}%
    \end{minipage}
    \\[2pt]
    (a) Low parallax & (b) Dynamic elements & (c) Global BA \\[2pt]
\end{tabular}
}
\vspace{-9pt}
\caption{
\label{fig:limitations}
{Limitations. (a) Without parallax (pure rotations), accurate triangulation is impossible, leading to poor geometry, visible in the depth renders (bottom row).
(b) We do not handle dynamic elements, which can lead to artifacts. These two limitations are shared with most SfM, SLAM, and static radiance field methods.
(c) Our approach produces a sharper background than GLOMAP+T3DGS by avoiding densification, but our pose estimation is less accurate than slower global BA methods such as GLOMAP, yielding a lower-quality foreground.}
}
\end{figure*}

\paragraph{BA runtime}
We compare our bundle adjustment solution with state-of-the-art BA solvers in g2o~\cite{kuemmerle2011g2o}, CudaBA~\cite{cudaba} and the MiniBA of \citet{meuleman2025onthefly} on a synthesized problem. As seen in Tab.~\ref{tab:ba_speed}, our solution convincingly outperforms all other solutions in speed.

\begin{table}[!h]
\caption{\label{tab:ba_speed} Runtime comparison for 10 iterations of different bundle adjustment optimizers for a problem with 20 poses and 10,000 landmarks. Note that g2o~\cite{kuemmerle2011g2o} and CudaBA~\cite{cudaba} are not designed to optimize intrinsics.}
\vspace{-9pt}
\small
\begin{tabular}{l|cccc}
\toprule
 & g2o & CudaBA & On-The-Fly & Ours \\
Runtime (ms) & 461.4 & 78.1 & 79.7 & 15.9 \\
\bottomrule
\end{tabular}
\end{table}

{
\paragraph{Hierarchy and memory scaling}
Tab.~\ref{tab:hierarchy-mem} summarizes peak number of trained Gaussians and peak GPU memory with and without our progressive hierarchy on \textsc{StaticHikes} (medium-sized) and \textsc{CityWalk} (large-scale). We sample these metrics each time new Gaussians are added, reporting GPU memory used by PyTorch-allocated tensors. On medium-sized scenes the hierarchy brings a 38\% reduction in peak number of Gaussians, while on \textsc{CityWalk} processing without it is infeasible.

\begin{table}[!h]
\setlength{\tabcolsep}{4pt}
\caption{\label{tab:hierarchy-mem} 
{
Peak number of trained Gaussians and peak GPU memory with and without our progressive hierarchy. 
\textsc{StaticHikes} is representative of medium-sized scenes; \textsc{CityWalk} represents large-scale captures, where processing without the hierarchy is infeasible (OOM on a 24\,GB GPU before 20\% of the capture is processed).
}
}
\vspace{-9pt}
\small
{
\begin{tabular}{l|cc|cc}
\toprule
 & \multicolumn{2}{c|}{\textsc{StaticHikes}} & \multicolumn{2}{c}{\textsc{CityWalk}} \\
 & \#GS & Mem & \#GS & Mem \\
\midrule
w/o hierarchy & 2.1\,M & 8.8\,GB & >6.7\,M$^{\dagger}$ & OOM ($>$24\,GB)$^{\dagger}$ \\
w/ hierarchy  & 1.3\,M & 6.2\,GB & 3.7\,M & 12.8\,GB \\
\bottomrule
\end{tabular}
}
\vspace{2pt}
\\
{\footnotesize $^{\dagger}$Reached after 795/4050 frames before OOM.}
\end{table}

\subsection{Limitations}
Fig.~\ref{fig:limitations} shows failure cases of our method. 
We use the fr1/rpy rotation-only sequence from \textsc{TUM}, originally captured for RGB-D SLAM evaluation where depth removes the need for triangulation, and the \textsc{Train} scene from \textsc{Davis}~\cite{perazzi2016benchmark}.
}

\end{document}